\documentclass[11pt]{article}

\usepackage{microtype}
\usepackage{graphicx}
\usepackage{subcaption}
\usepackage{booktabs}
\usepackage{microsoft-tech-report}
\usepackage{hyperref}
\usepackage{amsmath}
\usepackage{amssymb}
\usepackage{mathtools}
\usepackage{amsthm}
\usepackage{bm}
\usepackage{multirow}
\usepackage{enumitem}
\usepackage[capitalize,noabbrev]{cleveref}
\usepackage{xspace}
\usepackage{titletoc}
\usepackage{placeins}
\usepackage{makecell}
\usepackage{tcolorbox}
\usepackage[ruled,linesnumbered,noend,noline]{algorithm2e}
\DontPrintSemicolon
\usepackage{listings}
\usepackage{marvosym}
\usepackage{adjustbox}
\usepackage{CJKutf8}
\tcbuselibrary{theorems,skins,breakable,listings}

\definecolor{promptbg}{HTML}{F8F8F8}
\definecolor{promptborder}{HTML}{1F4E79}
\newtcolorbox{promptbox}[1][]{
  colback=promptbg, colframe=promptborder, fonttitle=\bfseries\sffamily,
  breakable, enhanced, arc=3pt, boxrule=0.8pt,
  left=10pt, right=10pt, top=8pt, bottom=8pt,
  title={#1}
}

\newif\ifappendixtoc
\appendixtocfalse

\definecolor{color_blue}{HTML}{E7EFFA}
\definecolor{color_green}{HTML}{E6F8E0}
\definecolor{color_gray}{HTML}{ECECEC}
\definecolor{theoremblue}{HTML}{EBF5FB}
\definecolor{theoremborder}{HTML}{2980B9}
\definecolor{propgreen}{HTML}{EAFAF1}
\definecolor{propborder}{HTML}{27AE60}
\definecolor{defyellow}{HTML}{FEF9E7}
\definecolor{defborder}{HTML}{F39C12}
\definecolor{remarkgray}{HTML}{F2F3F4}
\definecolor{remarkborder}{HTML}{7F8C8D}

\hypersetup{
  colorlinks=true,
  linkcolor=msftblue,
  citecolor=msftblue,
  urlcolor=msftblue,
  pdftitle={Mage-Flow: Efficient Native-Resolution Image Generation and Editing}
}

\newtcbtheorem[auto counter]{theorem}{Theorem}{
  colback=theoremblue, colframe=theoremborder, fonttitle=\bfseries,
  breakable, enhanced, sharp corners, boxrule=0.6pt, left=6pt, right=6pt, top=6pt, bottom=6pt,
  crefname={Theorem}{Theorems}
}{thm}
\newtcbtheorem[use counter from=theorem]{proposition}{Proposition}{
  colback=propgreen, colframe=propborder, fonttitle=\bfseries,
  breakable, enhanced, sharp corners, boxrule=0.6pt, left=6pt, right=6pt, top=6pt, bottom=6pt,
  crefname={Proposition}{Propositions}
}{prop}
\newtcbtheorem[use counter from=theorem]{remark}{Remark}{
  colback=remarkgray, colframe=remarkborder, fonttitle=\bfseries,
  breakable, enhanced, sharp corners, boxrule=0.6pt, left=6pt, right=6pt, top=6pt, bottom=6pt,
  crefname={Remark}{Remarks}
}{rem}
\crefname{tcb@cnt@theorem}{Theorem}{Theorems}
\crefname{tcb@cnt@proposition}{Proposition}{Propositions}
\crefname{tcb@cnt@remark}{Remark}{Remarks}
\crefname{algocf}{Algorithm}{Algorithms}

\newcommand{\magevae}{\textit{Mage-VAE}\xspace}
\newcommand{\mageflow}{\textit{Mage-Flow}\xspace}
\newcommand{\magebase}{\textit{Mage-Flow-Base}\xspace}
\newcommand{\mageturbo}{\textit{Mage-Flow-Turbo}\xspace}
\newcommand{\mageedit}{\textit{Mage-Flow-Edit}\xspace}
\newcommand{\mageeditbase}{\textit{Mage-Flow-Edit-Base}\xspace}
\newcommand{\mageeditturbo}{\textit{Mage-Flow-Edit-Turbo}\xspace}
\newcommand{\mageflowsci}{\textit{Mage-Flow-SciForma}\xspace}

\techreportshorttitle{Mage-Flow: Light-weight Multimodal Generation Models with Efficient Visual Tokenization}

\makeatletter
\let\orig@includegraphics\includegraphics
\renewcommand{\includegraphics}[2][]{%
  \IfFileExists{#2}%
    {\orig@includegraphics[#1]{#2}}%
    {\fbox{\parbox[c][2.6cm][c]{0.6\linewidth}{\centering\ttfamily\scriptsize%
       [missing figure]\\#2}}}%
}
\makeatother

\begin{document}
\thispagestyle{empty}

% ============================================================
%  Header: Microsoft logo + date
% ============================================================
\noindent
\begin{minipage}[c]{0.5\linewidth}
\raggedright
\raisebox{-0.5\height}{\msftbrandmark}
\end{minipage}
\begin{minipage}[c]{0.49\linewidth}
\raggedleft
{\msftdatefont\small\color{msftgray}July, 2026}
\end{minipage}\par
\vspace{0.35em}
\noindent{\color{msftline}\rule{\linewidth}{0.8pt}\par}

% ============================================================
%  Title + Authors
% ============================================================
\vspace{1.0em}
\begin{center}
{{\msfttitlefont\fontsize{19}{25}\selectfont\color{msftdark}
Mage-Flow: An Efficient Native-Resolution Foundation Model for Image Generation and Editing \par}}
\vspace{1.25em}

{\normalsize\rmfamily\color{msftdark}
\hyperref[sec:contributor]{Microsoft Mage Team}\par
}
\end{center}

% ============================================================
%  Abstract box
% ============================================================
\vspace{0.45em}
\begin{msfttitlebox}
\setlength{\parindent}{0cm}
\setlength{\parskip}{0cm}

\begin{abstract}
Large-scale visual generators are increasingly capable but costly to train, fine-tune, and deploy. We introduce \textbf{\mageflow}, a compact 4B-scale generative stack for efficient text-to-image generation and instruction-based image editing. The stack is built from two co-designed components: \textbf{\magevae}, a lightweight high-fidelity latent tokenizer, and a \textbf{Native-Resolution Multimodal Diffusion Transformer} trained with rectified flow matching.
\magevae uses one-step diffusion-style encoding and decoding with anchor-latent regularization, preserving the reconstruction quality of strong public VAEs while reducing tokenization cost by more than an order of magnitude. Together with native-resolution packing and stack-level CUDA kernel fusion, the stack supports flexible-resolution training and improves end-to-end training throughput by about $2.5\times$. Built on this foundation, we develop a complete model family with Base, RL-aligned, and Turbo variants for both generation and editing. Diffusion-NFT improves prompt following, text rendering, aesthetic quality, and editing fidelity, while few-step distillation with adversarial perceptual guidance produces 4-step Turbo models for low-latency inference. Despite its compact scale, \mageflow and \mageedit achieves competitive performance across standard generation and editing benchmarks. More importantly, the Turbo variants make high-resolution generation and editing practical for interactive use: at $1024^2$ resolution on a single NVIDIA A100 GPU, \mageturbo generates an image in $0.59$s, and \mageeditturbo edits an image in $1.02$s, while maintaining a small memory footprint. These results show that careful tokenizer--backbone--system co-design can deliver strong high-resolution generation and editing within an efficient 4B model family.
\end{abstract}

\vspace{0.14cm}
{\setlength{\parskip}{0.06cm}\small
{\msftmetalabel{Project Page}\href{https://microsoft.github.io/Mage}{https://microsoft.github.io/Mage}\par}
{\msftmetalabel{Code}\href{https://github.com/microsoft/Mage}{https://github.com/microsoft/Mage}\par}
{\msftmetalabel{Model}\href{https://huggingface.co/collections/microsoft/mage}{https://huggingface.co/collections/microsoft/mage}\par}
{\msftmetalabel{Date}July, 2026\par}
}
\end{msfttitlebox}
\suppressfloats[t]

% ============================================================
%  Main content
% ============================================================

\section{Introduction}
\label{sec:introduction}

\begin{figure*}[p]
\centering
\includegraphics[width=\textwidth,height=0.9\textheight,keepaspectratio]{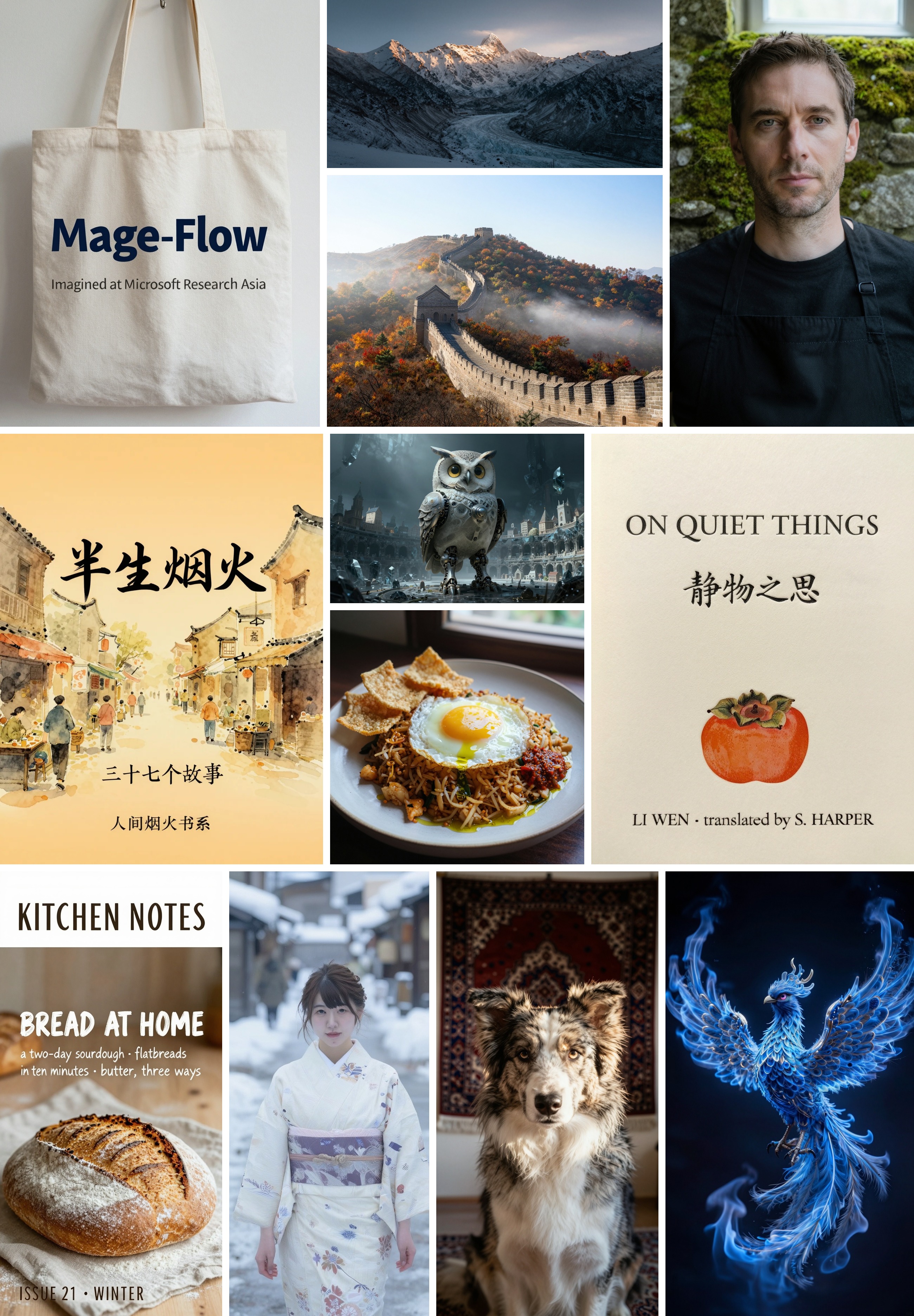}
\caption{\textbf{Qualitative showcase of \mageflow.}
Native-resolution samples spanning editorial design, photorealism, food photography, bilingual text
rendering, portraits, and stylized art. \mageflow supports flexible generation with height and width
from $512$ to $2048$, including extreme $4{:}1$ aspect ratios such as $512\times2048$ and
$2048\times512$. The examples demonstrate coherent layouts, fine visual details, and legible
English and Chinese text across diverse resolutions and aspect ratios.}
\label{fig:showcase}
\end{figure*}

\begin{figure*}[p]
\centering
\includegraphics[width=\textwidth,height=0.88\textheight,keepaspectratio]{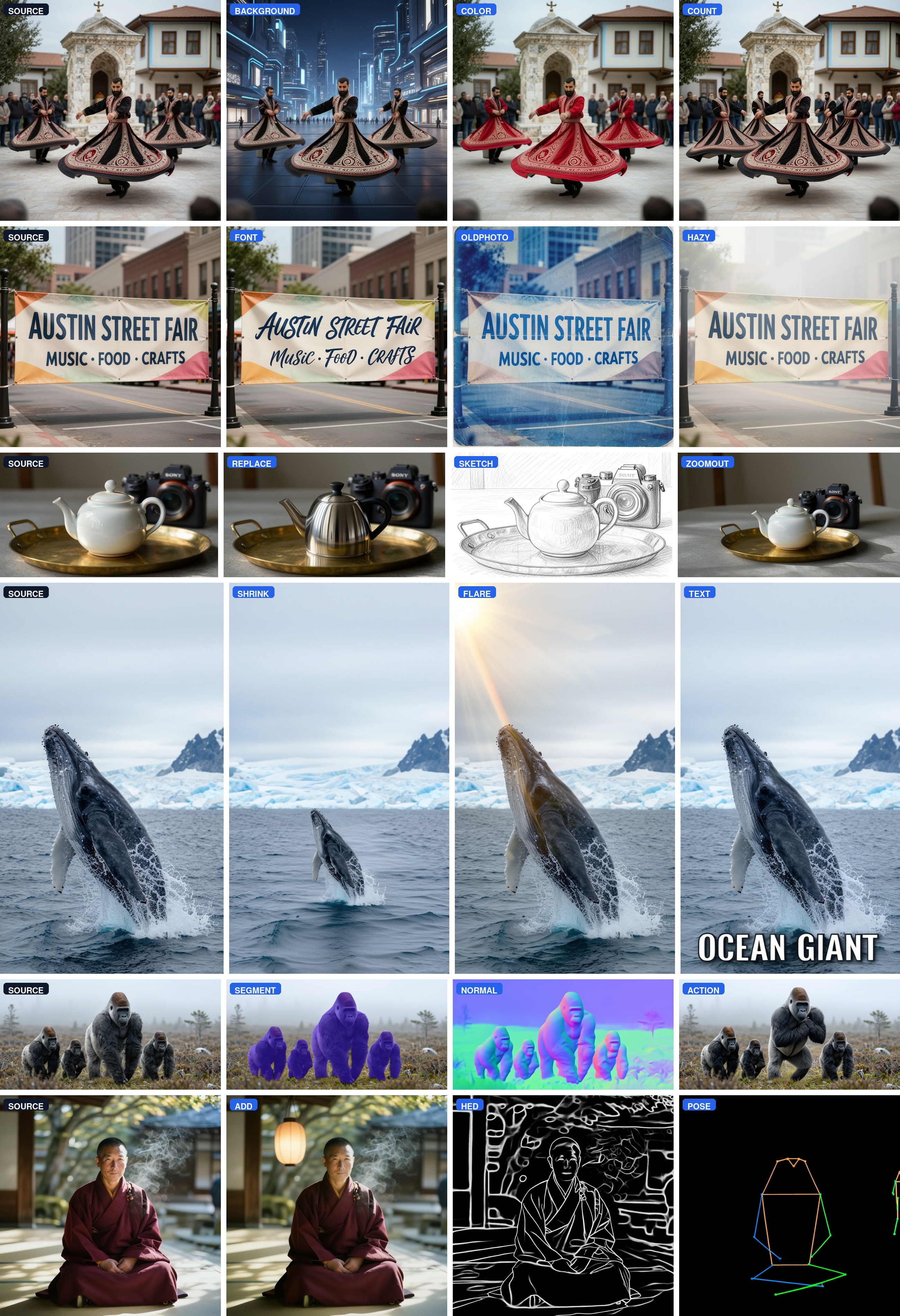}
\caption{\textbf{Qualitative showcase of \mageedit{} (I).}
Source-centric examples spanning background, color, count, font, old-photo and haze effects, subject
addition/replacement, sketch rendering, zoom-out, object shrinking, lens-flare synthesis, text
addition, segmentation, surface normals, action change, HED edges, and pose extraction. All showcased
edit types are supported bidirectionally.}
\label{fig:edit_gallery_showcase_1}
\end{figure*}

\begin{figure*}[p]
\centering
\includegraphics[width=\textwidth,height=0.88\textheight,keepaspectratio]{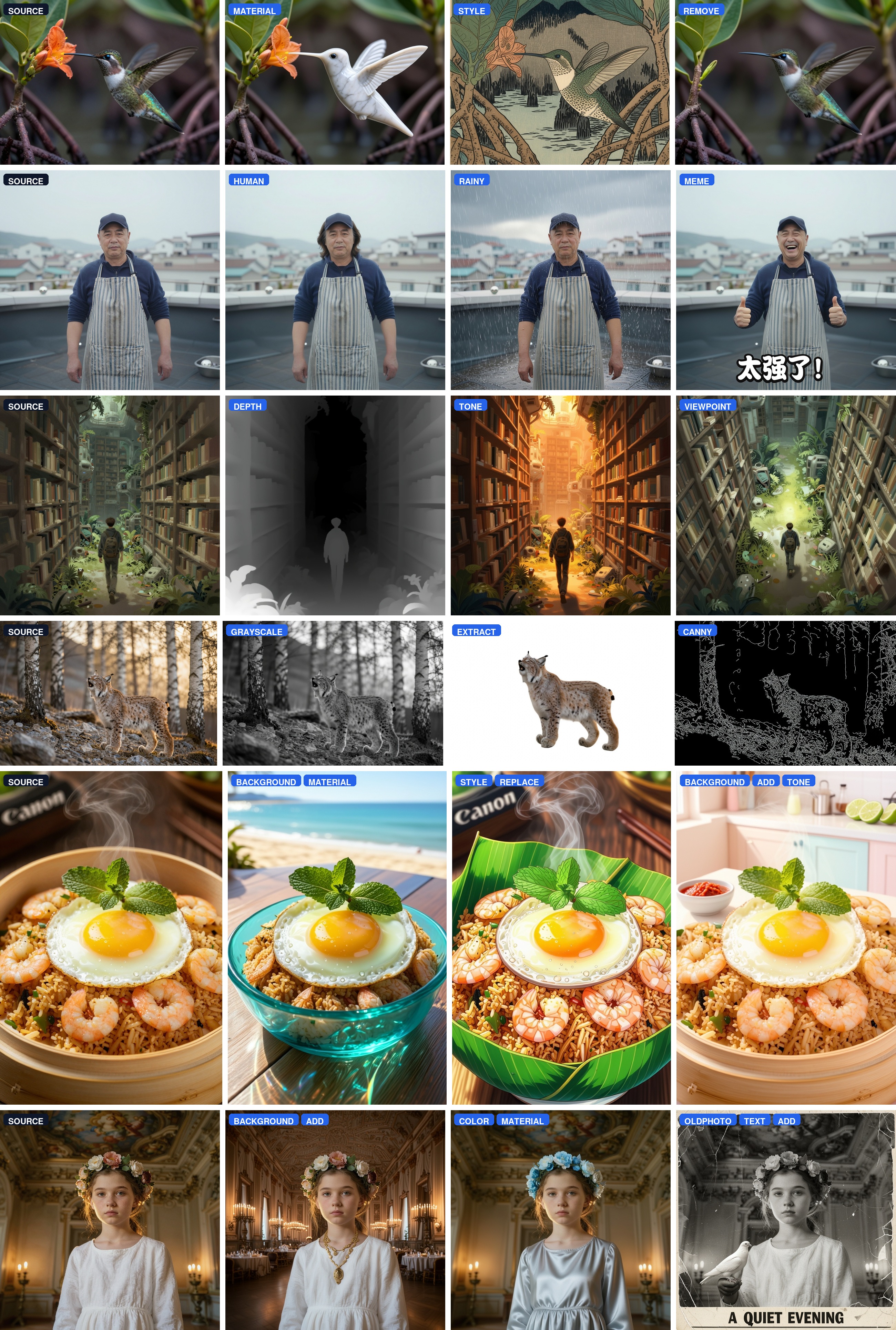}
\caption{\textbf{Qualitative showcase of \mageedit{} (II).}
Additional examples cover material and style changes, subject removal, human retouching, rainy weather,
meme creation, depth, tone, viewpoint, grayscale conversion, extraction, Canny edges, and multi-type
editing. All showcased edit types are supported bidirectionally.}
\label{fig:edit_gallery_showcase_2}
\end{figure*}

Recent advances in visual generative modeling have substantially improved both text-to-image generation and instruction-based image editing. Contemporary systems are expected to synthesize high-fidelity images, follow compositional prompts, render multilingual text, preserve spatial layout, and support localized or iterative editing. These capabilities enable a broad range of creation workflows, including poster and document design, product visualization, UI prototyping, scientific diagrams, and interactive image editing.

However, strong visual generation remains difficult to access and study under practical compute budgets. Closed-source systems such as GPT-Image~\cite{gptimage1}, Nano Banana~\cite{nanopro}, and Seedream~\cite{seedream2025seedream} provide strong performance but limit transparency and reproducibility. Meanwhile, competitive open-source models increasingly rely on large backbones: recent generators include Z-Image~\cite{zimage} at 6B, Qwen-Image~\cite{qwenimage} at 20B, FLUX.2~\cite{flux-2-2025} at 32B, and Hunyuan-Image-3.0~\cite{hunyuanimage3} at 80B, while strong editing models such as Qwen-Image-Edit~\cite{qwenimage} and
FireRed-Image-Edit~\cite{firered2025} also reach 20B scale. While these models achieve impressive absolute quality, their scale increases the cost of inference, fine-tuning, controlled ablations, and domain adaptation. As a result, there remains a gap between releasing open weights and making visual generation systems truly practical to study, modify, and deploy under realistic compute budgets.

In this report, we present the \textbf{\mageflow generative stack}, a compact and research-friendly foundation for efficient image generation and editing. The stack consists of \textbf{\magevae}, a high-fidelity lightweight latent tokenizer derived from our prior CoD-Lite design~\cite{codlite2026}, and a 4B \textbf{Native-Resolution Multimodal Diffusion Transformer} (NR-MMDiT) trained with rectified flow matching in the \magevae latent space. This shared stack, together with native-resolution packing and fused-kernel training nfrastructure, supports two model instantiations: \textbf{\mageflow} for text-to-image generation and \textbf{\mageedit} for instruction-based image editing. Rather than scaling to tens of billions of parameters, our goal is to provide a compact and extensible 4B-scale baseline for visual generation, controllable editing, post-training alignment, and vertical application research.

The \mageflow stack is designed around system-level efficiency rather than generator scaling alone. In latent generative pipelines, optimizing the diffusion backbone alone is insufficient: the tokenizer is invoked during training, inference, and repeated editing, and its cost grows rapidly with image resolution. We therefore first redesign the tokenizer component of the stack. Most modern public VAEs are optimized for pixel-level reconstruction but retain architectures whose encoder and decoder costs scale unfavorably at high resolution. In few-step high-resolution generation and editing, this cost can approach the diffusion backbone itself in latency and memory, making the tokenizer a practical pipeline bottleneck. 
To address this issue, we train \textbf{\magevae} from scratch under three design principles. First, building on our prior CoD-Lite exploration~\cite{codlite2026}, we treat the tokenizer as a learned image codec: the decoder is a fully convolutional pixel-diffusion model pre-trained with a compression-oriented objective and then distilled to a single step, avoiding the global attention blocks and multi-step computation that make conventional VAE decoders expensive at high resolution. Second, motivated by the structural symmetry of auto-encoding, we construct the encoder as the architectural dual of the decoder: a one-step diffusion model that generates latents conditioned on pixels, making encoding as lightweight as decoding. Third, we replace the standard Gaussian-prior KL with an anchor-latent KL that regularizes the posterior toward the latent distribution of a strong public VAE, yielding a generation-ready latent space for downstream diffusion training. As a result, \magevae attains reconstruction fidelity on par with FLUX.2-VAE while requiring approximately $12\times$ and $22\times$ fewer encoding and decoding MACs per pixel, respectively. This efficient tokenizer is the first layer of the \mageflow stack, enabling high-resolution generation and repeated editing with substantially lower tokenization overhead.

On top of this latent space, the stack uses a shared \textbf{Native-Resolution Multimodal Diffusion Transformer} as the generative backbone. The 4B NR-MMDiT is trained with rectified flow matching and is used by both text-to-image generation and instruction-based editing. Instead of conventional bucket-based training, where each optimization step is restricted to one predefined resolution and aspect-ratio bucket, we adopt a native-resolution packing scheme~\cite{wang2025nit}. It packs variable-length image sequences with arbitrary resolutions and aspect ratios, together with variable-length text sequences, into a single batch using FlashAttention's variable-length kernels~\cite{dao2023flashattention2,zadouri2026flashattention} and per-sample 2D rotary embeddings. This removes the single-bucket restriction, exposes each update to heterogeneous native image sizes, and allows one checkpoint to generalize naturally to flexible output sizes. The same packing mechanism also improves inference efficiency by evaluating the conditional and unconditional classifier-free guidance branches in one packed forward pass.

Beyond architecture, efficient native-resolution training requires stack-level systems optimization. We therefore fuse the dominant memory-bound operator chains in the repeated blocks of \magevae, the Qwen3-VL \cite{qwen3vl} text encoder, and NR-MMDiT into custom CUDA kernels. These fused kernels reduce activation memory traffic and kernel-launch overhead, increasing MFU from approximately $14\%$ to $29\%$ and achieving about a $2.5\times$ end-to-end training speedup (\tableautorefname~\ref{tab:efficiency}). Together, the lightweight tokenizer, native-resolution backbone, packed CFG inference, and fused-kernel training infrastructure make efficient 4B-scale native-resolution generation and editing practical under a fixed compute budget.

\begin{figure*}[t]
\centering
\begin{minipage}[c]{\textwidth}
  \centering
  \includegraphics[width=0.5\linewidth]{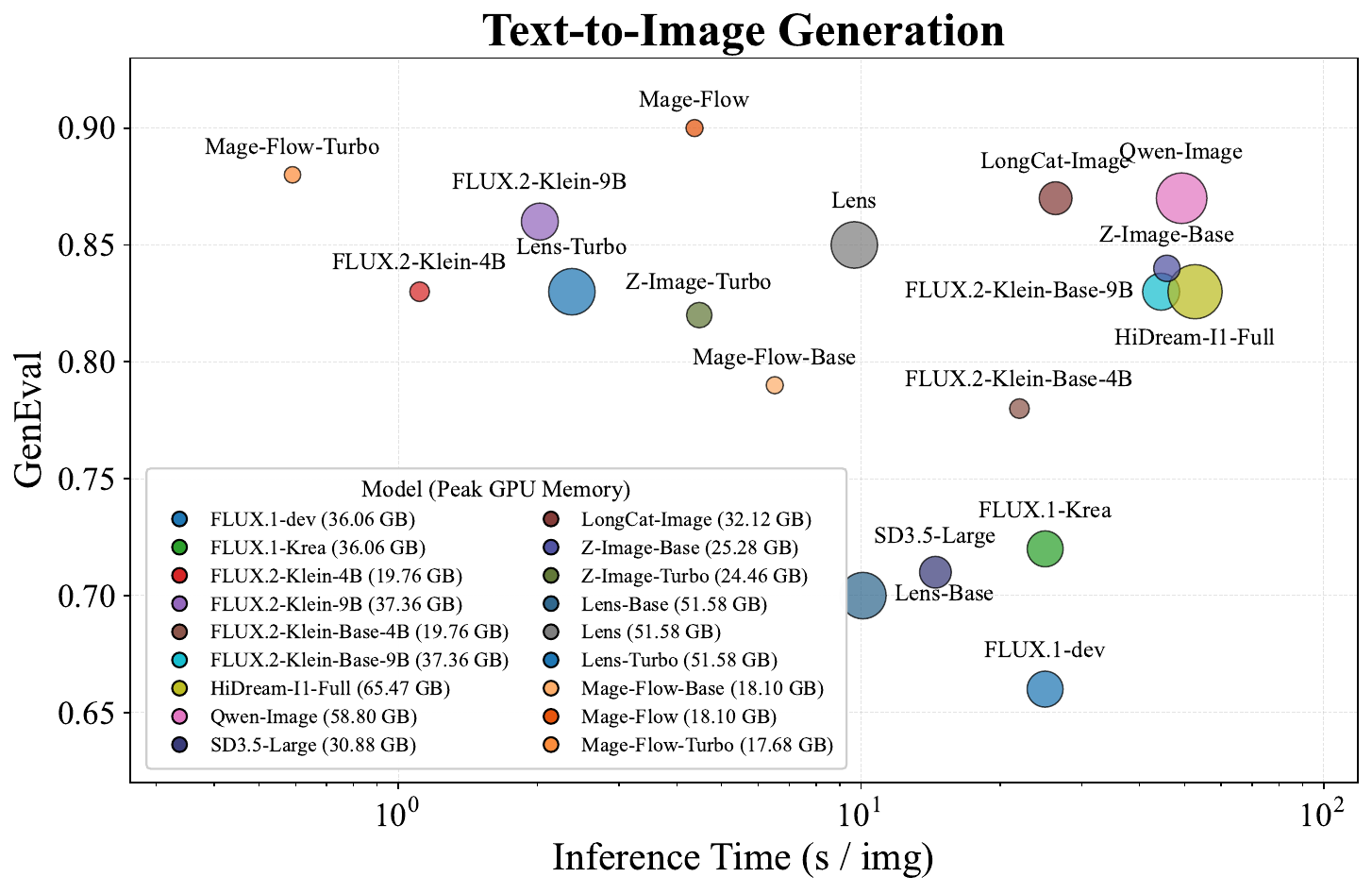}\hfill
  \includegraphics[width=0.5\linewidth]{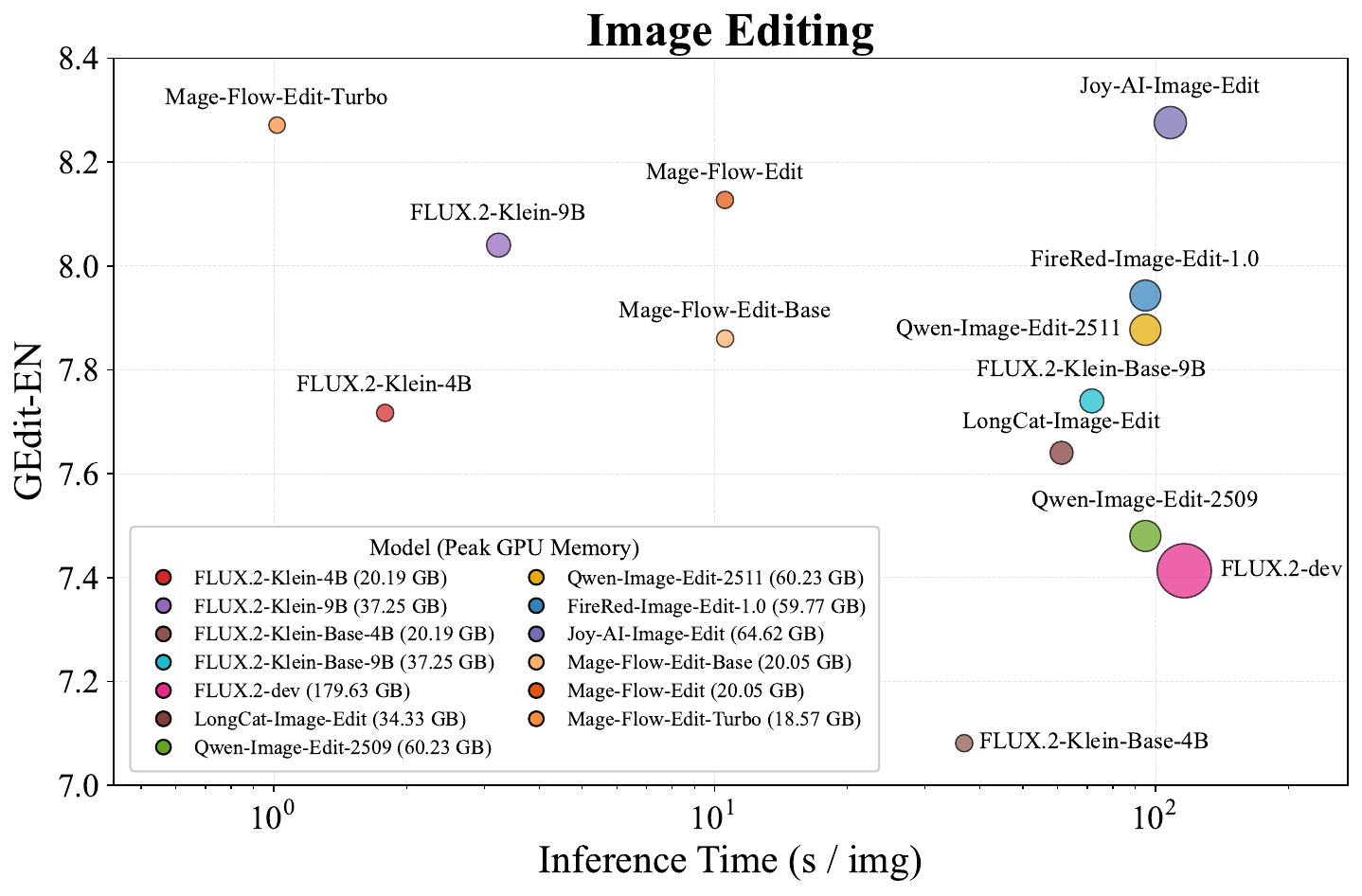}\hfill
  \captionof{figure}{\textbf{Quality--speed--memory comparison.}
  We compare representative text-to-image generation and image-editing models on NVIDIA A100 GPUs. The left panel reports GenEval performance, and the right panel reports GEdit-Bench-EN performance. The x-axis shows end-to-end inference time per image, the y-axis shows benchmark performance, and the marker area is proportional to peak GPU memory. All models are evaluated on a single A100 GPU except FLUX.2-dev, which is evaluated with two A100 GPUs due to memory requirements; for FLUX.2-dev, the reported time is the two-GPU inference time and the reported memory is the sum of peak memory across both GPUs. \mageflow and \mageedit form a favorable quality--efficiency frontier with strong scores, low latency, and small memory footprint.}
  \label{fig:t2i_editing_speed}
\end{minipage}
\end{figure*}

We instantiate this shared stack into a complete \textbf{generation-and-editing model family}. For text-to-image generation, we first train \textbf{\magebase} as a 4B native-resolution foundation model, align it into \textbf{\mageflow} with Diffusion-NFT~\cite{zheng2025diffusionnft} to improve prompt following, aesthetics, bilingual text rendering, and preference alignment, and distill it into the 4-step \textbf{\mageturbo} using Decoupled-DMD guidance~\cite{liu2025decoupleddmd} with adversarial perceptual guidance~\cite{ge2025senseflow}. For instruction-based editing, we reuse the same \magevae latent space and NR-MMDiT backbone, but change the conditioning format to include editing instructions and source-image latents. This yields \textbf{\mageeditbase}, \textbf{\mageedit}, and \textbf{\mageeditturbo}. Throughout editing training, we mix editing data with generation data, which improves source-conditioned editability while preserving the open-ended generative prior. As shown in Fig.~\ref{fig:t2i_editing_speed}, the resulting 4B-scale family achieves a strong quality--efficiency trade-off against representative open-source systems. At $1024^2$ resolution, on a single NVIDIA A100 GPU, \mageflow generates images in $4.37$s, while the 4-step \mageturbo reduces latency to $0.59$s. For editing, the 30-step \mageedit runs at $10.55$s, and the 4-step \mageeditturbo accelerates inference to $1.02$s. Across generation and editing, peak GPU memory remains around $18$-$20$ GB, the lowest among the compared models. Combined with competitive or superior benchmark performance against much larger open-source systems such as Qwen-Image~\cite{qwenimage}, Z-Image~\cite{zimage}, FLUX.2~\cite{flux-2-2025}, and FireRed-Image-Edit~\cite{firered2025}, these results show that the \mageflow stack provides a compact, fast, and memory-efficient foundation for local desktop deployment and downstream research. Overall, our main contributions are summarized as follows:
\begin{itemize}
    \item We propose the \textbf{\mageflow generative stack}, a compact 4B-scale foundation for efficient text-to-image generation and instruction-based image editing. The stack combines a lightweight tokenizer, a native-resolution diffusion backbone, and a fused-kernel training infrastructure, providing an extensible open baseline for visual generation research.

    \item We design \textbf{\magevae}, a lightweight high-fidelity latent tokenizer based on one-step diffusion-style encoding and decoding with anchor-latent KL regularization. It preserves the reconstruction quality of strong public VAEs while substantially reducing high-resolution encoding and decoding cost.

    \item We introduce a \textbf{Native-Resolution MMDiT} backbone trained with rectified flow matching. With native-resolution packing, variable-length text/image batching, packed CFG inference, and stack-level CUDA kernel fusion, the backbone enables efficient flexible-resolution training and inference under a fixed compute budget.

    \item We build a complete \textbf{generation-and-editing model family} from the shared stack, including Base, RL-aligned, and Turbo variants for both \mageflow and \mageedit. Across generation and editing settings, our models achieve a strong quality--speed--memory trade-off: they match or surpass much larger open-source systems while delivering faster inference and lower peak GPU memory among the compared models.
\end{itemize}

\section{Related Work}
\label{sec:related}
\subsection{Image Generation Foundation Models}
\label{sec:related_foundation}

\paragraph{Text-to-image generators.}
Text-to-image generation has evolved from pixel-space diffusion and U-Net-based latent diffusion toward large diffusion transformers and rectified-flow models. Stable Diffusion~\citep{ldm} popularized generation in a compressed VAE latent space, greatly reducing the cost of pixel-space diffusion, while SDXL~\citep{podell2024sdxl} scaled the U-Net design with stronger text conditioning and multi-aspect-ratio training. Diffusion Transformers~\citep{peebles2023dit} further shifted the field toward transformer backbones, with SD3~\cite{esser2024sd3} adopting MMDiT-style architectures trained with rectified flow, and SANA~\citep{xie2025sana} improving high-resolution efficiency through linear-attention diffusion transformers. Recent open-source systems continue to scale this paradigm, including Z-Image~\citep{zimage}, Qwen-Image~\citep{qwenimage}, FLUX.2~\citep{flux-2-2025}, HunyuanImage~3.0~\citep{hunyuanimage3}, and LongCat-Image~\citep{LongCat-Image}, while closed-source systems such as Nano Banana~\citep{nanopro} and Seedream~\citep{seedream2025seedream} show strong generation quality but limited transparency and reproducibility. Many of these models operate at 6B--80B scales, creating practical barriers for adaptation and systematic research. In contrast, \mageflow fixes the generator at a compact 4B scale and co-designs it with a lightweight VAE, aiming to provide an efficient, extensible, and research-friendly foundation for high-quality visual generation.

\paragraph{Instruction-based image editors.}
Diffusion-based image editing has progressed from inversion- and guidance-based methods to end-to-end instruction-tuned editors. Early methods such as SDEdit~\citep{meng2022sdedit} enabled global edits through stochastic denoising, while Prompt-to-Prompt~\citep{hertz2023prompt} and Null-text Inversion~\citep{mokady2023nulltext} improved localized editing by manipulating cross-attention or optimized embeddings, but often required per-image optimization. Another line introduced explicit conditioning, including ControlNet~\citep{zhang2023controlnet} for spatial control and IP-Adapter~\citep{ye2023ipadapter} for image-prompt conditioning. Instruction tuning was popularized by InstructPix2Pix~\citep{brooks2023instructpix2pix}, followed by larger-scale supervised editors such as MagicBrush~\citep{zhang2023magicbrush}, AnyEdit~\citep{yu2025anyedit}, UltraEdit~\citep{zhao2024ultraedit}, and OmniGen~\citep{xiao2025omnigen}. More recent MMDiT-based editors, including FLUX.1~Kontext~\citep{labs2025flux}, Step1X-Edit~\citep{liu2025step1x}, ICEdit~\citep{zhang2025icedit}, OmniGen2~\citep{wu2025omnigen2}, UniWorld-V1~\citep{lin2025uniworld}, DreamOmni2~\citep{xia2025dreamomni2}, ChronoEdit~\citep{wu2025chronoedit}, JoyAI-Image-Edit~\citep{joyaiedit}, and FireRed-Image-Edit~\citep{firered2025}, have improved instruction following, identity preservation, and compositional editing, but often rely on substantially larger backbones. \mageedit instead provides a compact 4B instruction-based editor that retains strong editing quality while substantially lowering the cost of adaptation, fine-tuning, and downstream research.

\paragraph{Unified multimodal generation models.}
A parallel line of work studies unified multimodal models that combine image understanding, generation, and editing in a single backbone~\citep{zhang2025unified}. Janus-Pro~\citep{chen2025januspro} uses separate visual encoders for understanding and generation while sharing one autoregressive transformer. Transfusion~\citep{zhou2025transfusion} trains a single transformer over interleaved text and image data with both next-token prediction and diffusion objectives. BAGEL~\citep{deng2025bagel} adopts a decoder-only architecture for multimodal understanding and generation, while Emu3.5~\citep{cui2025emu35nativemultimodalmodels} further scales native multimodal modeling toward broad generation and editing capabilities. These models pursue general-purpose multimodal unification and are typically larger and architecturally distinct from MMDiT-based generators. \mageflow focuses on the complementary direction: specializing a 4B MMDiT stack for efficient, high-quality generation and editing, while treating understanding as an orthogonal capability that can be incorporated through external encoders or downstream systems.

\subsection{Tokenization for Image Generation}
\label{sec:related_tokenization}

Visual tokenization bridges raw pixels and generative backbones, and largely determines the
efficiency frontier of modern image generators. Continuous-latent VAEs, introduced by Latent
Diffusion Models~\citep{ldm} and later used in systems such as SD3~\citep{esser2024sd3} and
FLUX.1, remain the dominant choice for diffusion-based generation. In parallel, discrete
tokenizers such as VQ-VAE~\citep{vqvae} and VQ-GAN~\citep{vqgan} laid the foundation for
autoregressive image generation. However, most open generators inherit public VAEs optimized
primarily for pixel-level reconstruction, whose encoder and decoder costs grow rapidly with
resolution and can become a major latency and memory bottleneck in high-resolution generation
and repeated editing.

A related line of work rethinks tokenization through diffusion-based generative compression. Multi-step diffusion codecs \cite{psc, idempotence, resulic} can achieve strong perceptual quality at low bitrates, but their iterative sampling makes them too slow for real-time generative pipelines; one-step diffusion codecs \cite{oscar, onedc, stablecodec} improve latency, but often rely on heavy DiT- or U-Net-style backbones. CoD-Lite~\citep{codlite2026} provides an important observation for this regime: at small scales, compression-oriented pre-training transfers better to learned codecs than generation-oriented pre-training, and after distillation and adversarial training, lightweight fully-convolutional decoders are sufficient without global attention. Building on these insights, \magevae trains the tokenizer from scratch for downstream MMDiT generation. Its decoder is a fully-convolutional pixel-diffusion model pre-trained with a compression-oriented objective and distilled to a single step, avoiding the global attention blocks and multi-step computation that make conventional VAE decoders expensive at high resolution. Since auto-encoding is symmetric, we construct the encoder as the architectural dual of the decoder: a one-step diffusion model that generates latents conditioned on pixels, making encoding as efficient as decoding. Finally, \magevae replaces the standard Gaussian-prior KL with an anchor-latent KL that regularizes the posterior toward a fixed VAE latent distribution, producing a generation-ready latent space with $16\times$ spatial reduction and $128$ channels.

\subsection{Post-training for Image Generation}
\label{sec:related_posttraining}
Post-training is critical for aligning pre-trained diffusion and flow-matching generators with human preferences, prompt adherence, and aesthetic quality. One major direction extends offline preference optimization to diffusion models. Diffusion-DPO~\citep{wallace2023diffdpo} applies the DPO objective over diffusion trajectories, while D3PO~\citep{yang2024d3po} fine-tunes directly from preference comparisons without an explicit reward model. These methods are simple and stable because they operate on pre-collected preference data, and have become common components in modern text-to-image post-training pipelines.

A complementary direction uses online reinforcement learning to optimize non-differentiable or black-box rewards. DDPO~\citep{black2024ddpo} casts denoising as a sequential decision process and fine-tunes generators with learned or rule-based rewards, often using reward models such as ImageReward~\citep{xu2023imagereward}. Extending RL to flow matching is more challenging because ODE sampling is deterministic and lacks an explicit action distribution. Flow-GRPO~\citep{liu2025flowgrpo} addresses this by converting the ODE sampler into a stochastic SDE with matched marginals, whereas Diffusion-NFT~\citep{zheng2025diffusionnft} bypasses the reverse process and performs negative-aware fine-tuning on the forward process, requiring neither likelihood estimation nor a specific solver. In \mageflow and \mageedit, we adopt Diffusion-NFT as the post-training stage on top of supervised fine-tuning, applying it to curated capability-targeted generation and editing data to improve prompt adherence, text rendering, compositional control, and editing fidelity.

\begin{figure*}[t]
\centering
\includegraphics[width=\linewidth]{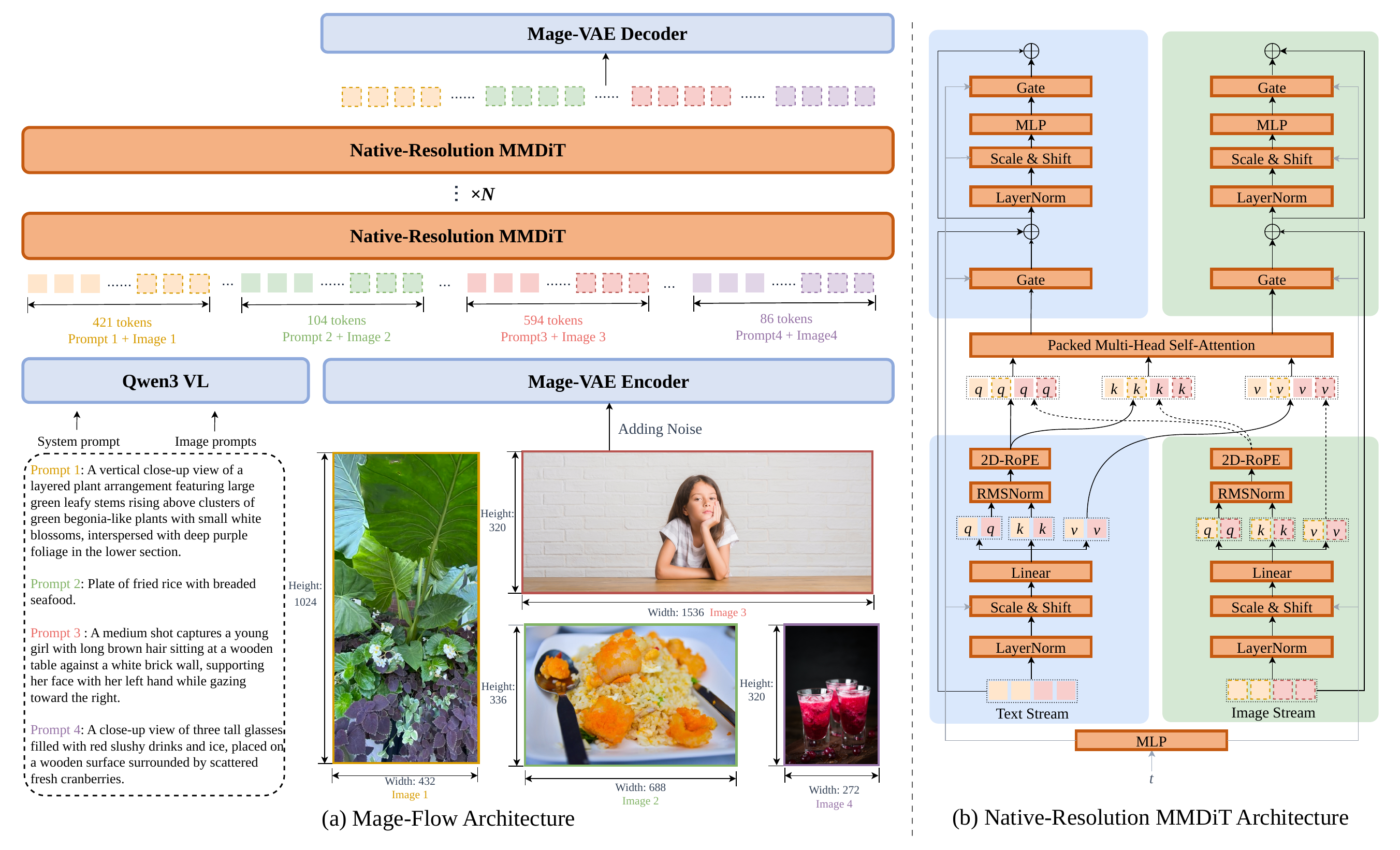}
\caption{\textbf{\mageflow architecture.}
\mageflow uses Qwen3-VL to encode text prompts and \magevae to map images into compact
transformer-ready latents. Images of different resolutions and aspect ratios are flattened into
variable-length latent token sequences and packed together with variable-length text tokens in a
single batch. The 4B Native-Resolution MMDiT processes these packed sequences with per-sample
2D rotary positional embeddings and FlashAttention variable-length attention, avoiding fixed
resolution buckets while preserving each sample's native spatial layout. The right panel
illustrates the MMDiT block, where text and image streams use modality-specific normalization and
projection layers but interact through joint self-attention.}
\label{fig:nr_mmdit}
\end{figure*}

\subsection{Distillation for Image Generation}
\label{sec:related_distillation}

Iterative diffusion sampling is costly at deployment, motivating extensive work on reducing the
number of sampling steps. Training-free solvers such as DDIM~\citep{song2021ddim} and higher-order
ODE solvers reduce step counts without retraining but often degrade under very small budgets.
Distillation methods push this further: progressive distillation~\citep{salimans2022progressive}
iteratively halves teacher trajectories; consistency models~\citep{song2023consistency} and latent
consistency models~\citep{luo2023lcm} learn few-step mappings through self-consistency along the
probability-flow ODE; and InstaFlow~\citep{liu2024instaflow} improves one-step generation by
straightening flow trajectories.

Another family distills at the distribution level. Distribution Matching Distillation
(DMD)~\citep{yin2024dmd} matches the student distribution to a teacher using an auxiliary score
model, while DMD2~\citep{yin2024dmd2} removes regression loss, introduces adversarial training, and
supports multi-step student sampling. Adversarial diffusion distillation~\citep{sauer2023add} and
Hyper-SD~\citep{ren2024hypersd} further combine score- or consistency-based objectives with adversarial losses to preserve visual detail at small step counts. More recently, Decoupled DMD~\citep{liu2025decoupleddmd} separates CFG augmentation from distribution matching and allows independent noise schedules, while SenseFlow~\citep{ge2025senseflow} introduces adversarial guidance based on frozen Vision Foundation Models (VFMs) for flow-based generators. Specifically, frozen VFMs such as CLIP~\citep{radford2021clip} and DINOv2~\citep{oquab2024dinov2} are used to extract semantic features from both generated and real images, and a lightweight feature discriminator trained on these representations provides adversarial gradients to improve realism and semantic alignment. \mageflow and \mageedit use Decoupled DMD as the main distillation backbone and add this VFM-based adversarial guidance as a complementary signal, producing 4-step Turbo variants for both generation and editing.

\section{Mage-Flow Stack}
\label{sec:model}

As shown in Fig.~\ref{fig:nr_mmdit}(a), the \mageflow stack is built around two core model components: \magevae, a lightweight
high-fidelity latent tokenizer, and a 4B Native-Resolution Multimodal Diffusion Transformer
(NR-MMDiT) trained with rectified flow matching. \magevae provides a compact generation-ready
latent space, while NR-MMDiT models packed latent sequences with flexible resolutions and aspect
ratios. The stack is supported by a training and inference infrastructure that makes native-resolution 4B-scale learning practical. We describe \magevae in \cref{sec:model_vae}, NR-MMDiT in \cref{sec:model_mmdit}, and the training infrastructure in \cref{sec:training_infra}.

%The \mageflow stack consists of two tightly coupled components: \magevae, a lightweight latent image tokenizer; a 4B Native-Resolution Multimodal Diffusion Transformer (NR-MMDiT) trained with rectified flow matching; and a training infrastructure that enables efficient native-resolution 4B-scale learning. \magevae provides a compact high-fidelity latent space for generation and editing, while NR-MMDiT operates directly on packed latent sequences of arbitrary resolution and aspect ratio. To make this design practical at scale, our training infrastructure combines native-resolution packing with fused CUDA kernels for the dominant repeated blocks in \magevae, the Qwen3-VL text encoder \cite{qwen3vl}, and NR-MMDiT, reducing padding overhead and memory-bound operator cost. We describe \magevae in \cref{sec:model_vae}, NR-MMDiT in \cref{sec:model_mmdit}, and the training infrastructure in \cref{sec:training_infra}.

\begin{figure*}[t]
\centering
\includegraphics[width=\linewidth]{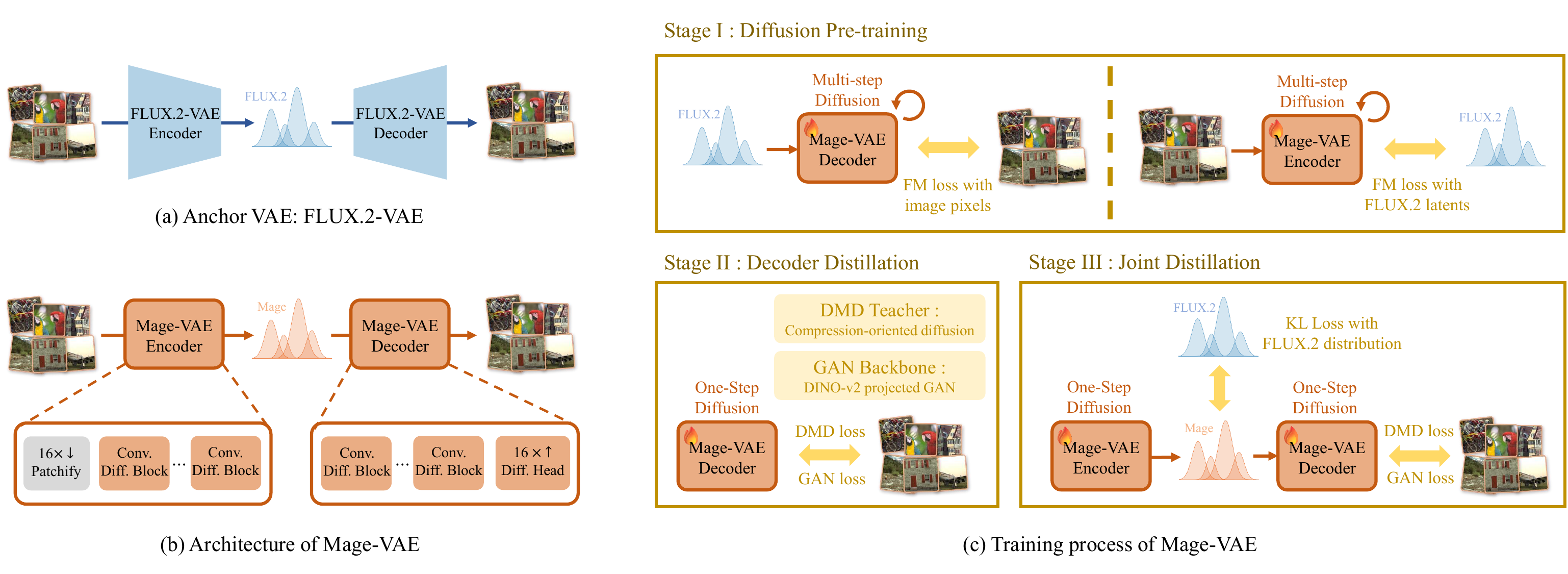}
\caption{\magevae is a lightweight pixel-diffusion-based VAE distilled from the FLUX.2-VAE latent space. FLUX.2-VAE provides the anchor latent distribution used for diffusion pre-training and anchor-latent regularization. The \magevae decoder is distilled to reconstruct images from FLUX.2-aligned latents, while the \magevae encoder is trained to produce latents compatible with the same anchor space. This design yields an efficient one-step encoder--decoder that preserves a generation-ready latent structure and can be interchanged with FLUX.2-VAE in downstream generators.}
\label{fig:mage_vae}
\end{figure*}

\subsection{Mage-VAE: Lightweight Image Tokenizer}
\label{sec:model_vae}

Public VAEs released alongside SD3~\citep{esser2024sd3} and the FLUX family~\citep{flux-2-2025} largely inherit the VQGAN~\citep{esser2021taming} and LDM~\citep{rombach2022high} autoencoder architectures, which were originally designed for low-resolution $256\times256$ images. When scaled to modern 1K and 2K image generation, inference latency and memory consumption grow unfavorably due to architectural components such as global attention and computationally expensive high-resolution blocks. As a result, VAE encoding and decoding can become a significant source of latency, approaching the diffusion Transformer itself in few-step high-resolution generation. For example, during the 4-step generation of 1K-resolution images using FLUX.2-Klein-4B, VAE decoding accounts for 14\% of the total generation time. \magevae is designed to remove this bottleneck while preserving reconstruction fidelity. It replaces heavy high-resolution autoencoder components with lightweight one-step diffusion-style encoding and decoding, and regularizes the latent space with an anchor-latent KL toward a strong public VAE. Its architecture and training pipeline are summarized in Fig.~\ref{fig:mage_vae}.

\subsubsection{Architecture}
\label{sec:model_vae_arch}

\paragraph{Lite codec-style decoder.}
Our decoder design is inspired by CoD-Lite~\citep{codlite2026}, a lightweight diffusion-based image codec for high-fidelity perceptual reconstruction. CoD-Lite shows that compression-oriented diffusion pre-training, followed by one-step distillation, can yield strong reconstruction quality without heavy DiT- or U-Net-style decoders. Following this principle, we formulate the \magevae decoder as a one-step pixel diffusion model. It uses stacked convolutional diffusion blocks~\citep{ai2026dico} and a decoupled pixel diffusion head~\citep{ma2026deco} to reconstruct RGB pixels directly from latents. This fully convolutional design avoids global attention and keeps decoding cost nearly linear in image resolution.

\paragraph{Lite symmetric encoder.}
Unlike image codecs, where decoding efficiency is often the primary concern, generative models also require efficient encoding: training needs to prepare image latents at scale, and editing models must repeatedly encode reference or source images. We therefore design the encoder as the architectural dual of the decoder. Since a decoder can be viewed as a pixel generator conditioned on latents, we view the encoder as a latent generator conditioned on pixels. Concretely, the encoder is also implemented as a one-step diffusion model, consisting of a patch embedding layer followed by stacked convolutional diffusion blocks. This symmetric design makes latent extraction as lightweight as pixel reconstruction.

\begin{table*}[t]\centering
\caption{Evaluation results of \magevae. The model parameters and computational complexity (measured by kMACs/pixel) of both encoding and decoding are reported. Reconstruction
quality is measured on CLIC 2020 testset at native resolution and FFHQ val~10k at $1024\times1024$ resolution.}
\label{tab:vae_evaluation_table}
\setlength{\tabcolsep}{3pt} 
\resizebox{0.9\textwidth}{!}{
\begin{tabular}{l| cc |cc |ccc |ccc}
\toprule
\multirow{2}{*}{Model} & \multicolumn{2}{c}{Params (M)} & \multicolumn{2}{|c|}{kMACs/px} & \multicolumn{3}{c}{CLIC 2020 Test} & \multicolumn{3}{|c}{FFHQ (val 10k)} \\
\cmidrule(lr){2-3}\cmidrule(lr){4-5}\cmidrule(lr){6-8}\cmidrule(lr){9-11}
 & Enc. & Dec. & Enc. & Dec. & PSNR$\uparrow$ & SSIM$\uparrow$ & LPIPS$\downarrow$ & PSNR$\uparrow$ & SSIM$\uparrow$ & LPIPS$\downarrow$ \\
\midrule
SD-VAE~\cite{rombach2022high}         & 34  & 49  & 2130 & 4796  & 30.10 & 0.8156 & 0.0565 & 33.00 & 0.8662 & 0.0461 \\
SD-3.5-VAE~\cite{esser2024sd3}    & 34  & 50  & 2132 & 4797  & 32.79 & 0.8896 & 0.0271 & 36.23 & 0.9330 & 0.0189 \\
FLUX-VAE~\cite{labs2025flux}      & 34  & 50  & 2132 & 4797  & 34.80 & 0.9264 & 0.0188 & 38.43 & 0.9587 & 0.0129 \\
FLUX.2-VAE~\cite{flux-2-2025}     & 34  & 50  & 2134 & 4798  & \textbf{36.88} & 0.9447 & \textbf{0.0139} & 40.47 & 0.9682 & \textbf{0.0102} \\
Qwen-Image-VAE~\cite{qwenimage}                    & 54  & 73  & 3378 & 5423  & 34.95 & 0.9129 & 0.0571 & 38.75 & 0.9515 & 0.0422 \\
HunyuanVideo-VAE~\cite{kong2024hunyuanvideo}       & 100 & 146 & 6198 & 14096 & 36.14 & 0.9305 & 0.0250 & 39.86 & 0.9610 & 0.0205 \\
HunyuanImage-3.0-VAE~\cite{hunyuanimage3}           & 389 & 871 & 28053 & 49836 & 33.80 & 0.8864 & 0.0546 & 36.84 & 0.9259 & 0.0468 \\
\midrule
\textbf{Mage-VAE}          & 49  & 52  & \textbf{173} & \textbf{215} & 36.61 & \textbf{0.9450} & 0.0148 & \textbf{40.67} & \textbf{0.9708} & 0.0107 \\
\bottomrule
\end{tabular}
} % 结束 resizebox
\end{table*}

\begin{figure*}[t]
\centering
\begin{minipage}[c]{\textwidth}
  \centering
  \includegraphics[width=0.495\linewidth]{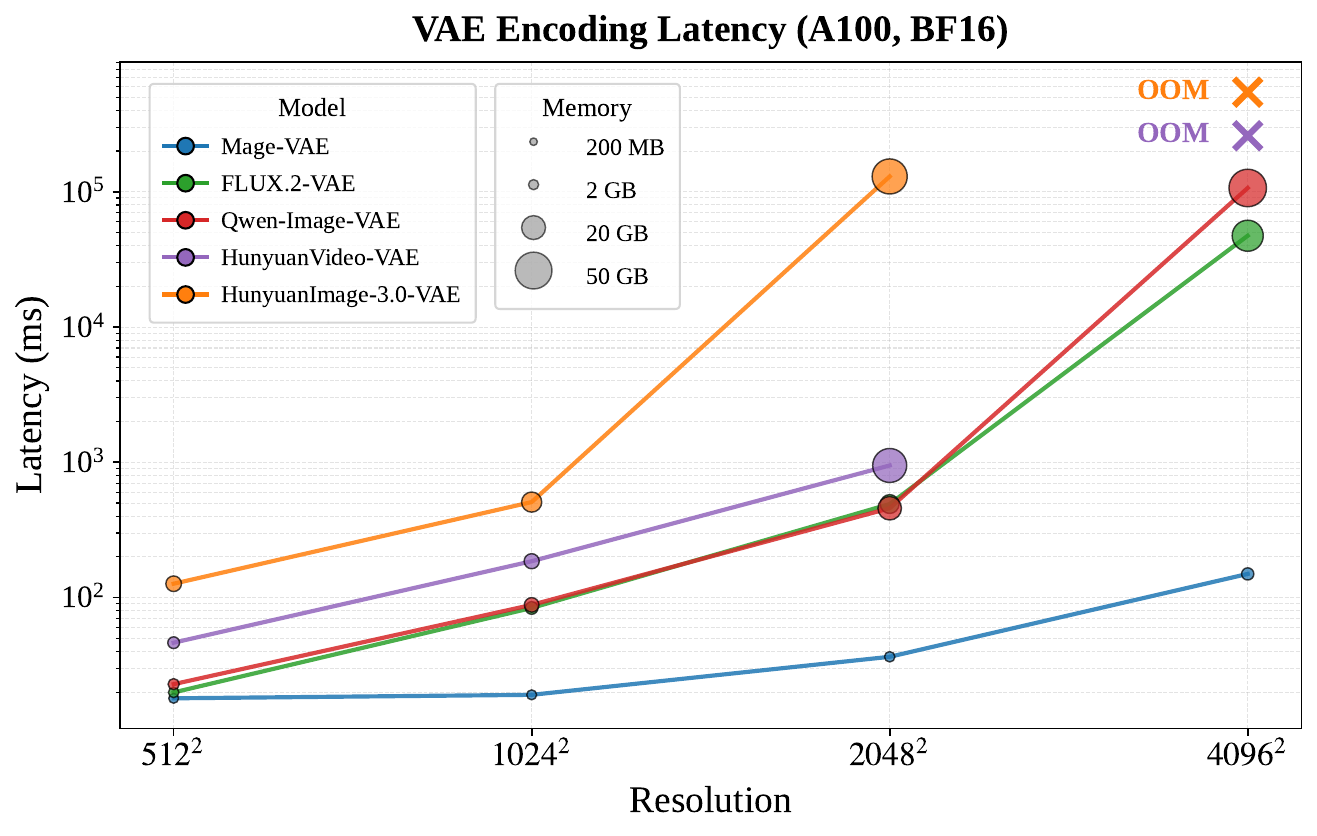}
  \includegraphics[width=0.495\linewidth]{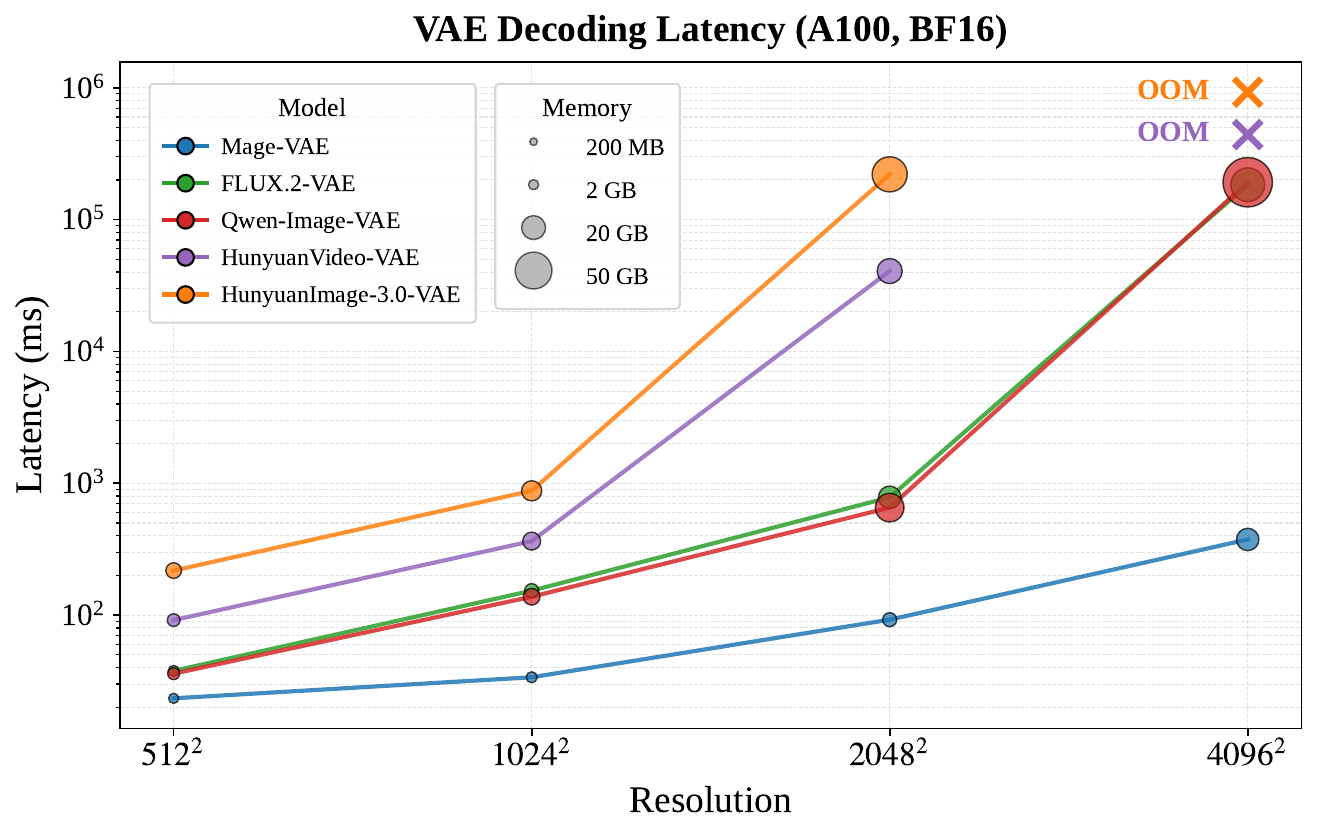}
  \captionof{figure}{Inference cost of \magevae. Encoding and decoding latency and memory is tested across resolutions on 80GB A100 GPU. SD-VAE, SD-3.5-VAE, and FLUX-VAE are not shown, as their model structures are almost identical to FLUX.2-VAE, causing their inference costs to completely overlap.}
  \label{fig:vae_speed}
\end{minipage}
\end{figure*}

\subsubsection{KL regularization with an anchor latent distribution}
\label{sec:model_vae_kl}
% CoD-Lite constrains its latent space through a bitrate objective. To obtain a standard variational autoencoder, we replace this bitrate constraint with a KL regularizer. Unlike conventional VAEs that compute the KL divergence against a standard Gaussian prior, we regularize the posterior toward the latent distribution induced by the FLUX.2 VAE~\cite{flux-2-2025}. This anchor latent space is also used to pre-train our encoder, as described in the following sections.
% The FLUX.2 VAE employs 32 latent channels with an $8\times$ spatial reduction. To improve inference efficiency, our VAE operates directly at a $16\times$ spatial reduction (so both the encoder and decoder work entirely in the $1/16$ resolution latents) and increases the latent dimensionality to 128 channels. The FLUX.2 latents are further patch-embedded by a factor of $2\times$ to match the latent shape of our model for KL regularization.

% Learned image codecs optimize a bitrate objective in latent space, which is equivalent to a KL-divergence term under the commonly adopted uniform-noise posterior approximation. From this perspective, a VAE can be interpreted as an image codec with a Gaussian latent posterior. Consequently, architectural advances developed for learned image compression naturally transfer to VAE design.

CoD-Lite constrains its latent space with a bitrate objective. Under the commonly used uniform-noise posterior approximation, this bitrate objective can be interpreted as a KL-style regularization term in latent space. From this perspective, a VAE can be viewed as a learned image codec with a Gaussian latent posterior, suggesting that codec architectures can be transferred to VAE design by replacing bitrate control with variational regularization.

Following this view, we convert the CoD-Lite-style codec into a generation-ready VAE by replacing the bitrate constraint with a KL regularizer. Unlike conventional VAEs that match the posterior to a standard Gaussian prior, we regularize the posterior toward an anchor latent distribution induced by the FLUX.2-VAE~\citep{flux-2-2025}. Since FLUX.2-VAE uses 32 latent channels with an $8\times$ spatial reduction and these latents are typically $2\times$ patchified before entering the diffusion Transformer, \magevae internalizes this patchification by directly producing $16\times$-downsampled latents with 128 channels. We use the patchified FLUX.2 latents as anchor targets for both encoder pre-training and KL regularization, allowing \magevae to inherit a generation-friendly latent structure while producing Transformer-ready latents for downstream diffusion training.

\subsubsection{Training}
\label{sec:model_vae_train}
% The training pipeline consists of three stages. In \textbf{Stage I}, we pre-train the encoder and decoder separately as diffusion models. Both are optimized with the standard flow-matching objective in the $\mathcal{X}$-prediction parameterization. For the encoder, the target latents are obtained from the FLUX.2 VAE.
% In \textbf{Stage II}, we distill decoder into a one-step architecture. We fine-tune the multi-step diffusion decoder using a DINOv2-projected GAN loss~\cite{sauer2021projected} and a DMD loss~\cite{yin2024dmd} with supervision from a compression-oriented diffusion teacher~\cite{jia2026cod}. 
% In \textbf{Stage III}, we jointly fine-tune the encoder and decoder with an additional KL regularization term with the FLUX.2 VAE.

The training pipeline contains three stages. In \textbf{Stage I}, we pre-train the encoder and decoder separately as multi-step diffusion models using a standard flow-matching objective in the $\mathcal{X}$-prediction parameterization. The decoder learns to reconstruct image pixels, while the encoder learns to predict the patchified FLUX.2-VAE anchor latents from pixels. In \textbf{Stage II}, we distill the decoder into a one-step model using reconstruction loss, a DMD loss~\citep{yin2024dmd} with a compression-oriented diffusion teacher~\citep{jia2026cod}, together with a DINOv2-projected GAN loss~\citep{sauer2021projected} to improve perceptual fidelity. In \textbf{Stage III}, we jointly fine-tune the one-step encoder and decoder with the anchor-latent KL regularizer, producing a compact VAE whose latent space is ready for \mageflow training. The detailed training process can be found in Appendix~\ref{app:vae_training}.

\subsubsection{Evaluation}
\label{sec:model_vae_evaluation}
%Table~\ref{tab:vae_evaluation_table} and Figure~\ref{fig:vae_speed} present the quantitative evaluation of \magevae. Compared to existing VAEs, our method achieves a significant reduction in computational complexity, requiring $12\times$ and $22\times$ fewer kMACs per pixel for encoding and decoding, respectively. Furthermore, \magevae delivers inference times and GPU memory cost that are orders of magnitude faster than baseline VAEs, with these advantages widening as image resolution increases. For reconstruction quality, we evaluate it on the high-quality general-domain CLIC 2020 test set~\cite{} at its native resolution ($\sim$2K) and the portrait FFHQ validation dataset~\cite{karras2019style} at $1024\times1024$. \magevae yields fidelity comparable to the FLUX.2 VAE while consistently outperforming other prior methods.

\begin{table*}[t]
\centering
\caption{\textbf{Anchor-latent tokenizer compatibility.}
We evaluate generation and editing backbones with different latent tokenizers. The comparison swaps between \magevae and FLUX.2-VAE while keeping the downstream backbone fixed. Higher is better for all metrics.}
\label{tab:vae_ablation}
\begingroup
\setlength{\tabcolsep}{4pt}
\renewcommand{\arraystretch}{1.05}
\small

\textbf{(a) Text-to-image generation.}

\vspace{2pt}
\begin{adjustbox}{width=\textwidth}
\begin{tabular}{l l c c c c c c c c c}
\toprule
\textbf{Backbone}
& \textbf{Tokenizer}
& \textbf{GenEval}
& \textbf{DPG-Bench}
& \textbf{TIIF-Short}
& \textbf{TIIF-Long}
& \textbf{CVTG-2K}
& \textbf{OneIG-EN}
& \textbf{OneIG-CN}
& \textbf{LongText-EN}
& \textbf{LongText-CN} \\
\midrule
\multirow{2}{*}{\mageturbo}
& \magevae
& 0.88 & 85.48 & 83.58 & 84.16 & 0.873 & 0.523 & 0.491 & 0.911 & 0.801 \\
& FLUX.2-VAE
& 0.88 & 85.49 & 81.18 & 81.09 & 0.869 & 0.523 & 0.491 & 0.909 & 0.800 \\
\midrule
\multirow{2}{*}{FLUX.2-Klein-4B~\cite{flux-2-2025}}
& FLUX.2-VAE
& 0.83 & 85.53 & 78.91 & 79.04 & 0.628 & 0.500 & 0.364 & 0.649 & 0.068  \\
& \magevae
& 0.83 & 85.52 & 77.40 & 79.43 & 0.623 & 0.502 & 0.361 & 0.654 & 0.068  \\
\bottomrule
\end{tabular}
\end{adjustbox}

\vspace{8pt}

\textbf{(b) Instruction-based editing.}

\vspace{2pt}
\begin{adjustbox}{width=\textwidth}
\begin{tabular}{l l c c c c c}
\toprule
\textbf{Backbone}
& \textbf{Tokenizer}
& \textbf{ImgEdit}
& \textbf{GEdit-EN}
& \textbf{GEdit-CN}
& \textbf{TextEdit-Syn}
& \textbf{TextEdit-Real} \\
\midrule
\multirow{2}{*}{\mageeditturbo}
& \magevae
& 4.38 & 8.271 & 8.264 & 12.77 & 15.41  \\
& FLUX.2-VAE
& 4.34 & 8.098 & 8.112 & 12.85 & 15.26 \\
\midrule
\multirow{2}{*}{FLUX.2-Klein-4B~\cite{flux-2-2025}}
& FLUX.2-VAE
& 4.01 & 7.717 & 7.750 & 11.84 & 14.46 \\
& \magevae
& 3.95 & 7.734 & 7.672 & 11.65 & 14.46 \\
\bottomrule
\end{tabular}
\end{adjustbox}

\endgroup
\end{table*}

% Table~\ref{tab:vae_evaluation_table} and Fig.~\ref{fig:vae_speed} evaluate \magevae from both reconstruction quality and inference efficiency. For reconstruction, we use two complementary settings: the high-quality general-domain CLIC 2020 test set~\cite{clic2020} at its native resolution ($\sim$2K), and the portrait-domain FFHQ validation set~\cite{karras2019style} at $1024\times1024$. \magevae achieves fidelity comparable to the strongest FLUX.2-VAE baseline on CLIC 2020, while obtaining the best or near-best reconstruction quality on FFHQ among the compared VAEs. This shows that the proposed lightweight design preserves high visual fidelity across both general and face-centric images.

% In terms of efficiency, \magevae reduces the computational complexity of FLUX.2-VAE by about $12.3\times$ for encoding and $22.3\times$ for decoding. As shown in Fig.~\ref{fig:vae_speed}, this translates into consistently lower latency and memory usage across all tested resolutions. The advantage becomes more pronounced at high resolution, where baseline VAEs become extremely slow or even run out of memory at $4096^2$, while \magevae remains efficient for both encoding and decoding. These results demonstrate that \magevae provides a substantially better quality-efficiency trade-off, making it well suited for high-resolution generation, repeated editing, and interactive deployment.

\paragraph{Reconstruction.}
In \cref{tab:vae_evaluation_table}, we evaluate the reconstruction performance of \magevae using two complementary settings: the high-quality general-domain CLIC 2020 test set~\citep{clic2020} at its native resolution ($\sim$2K), and the portrait-domain FFHQ validation set~\citep{karras2019style} at $1024\times1024$. \magevae achieves reconstruction fidelity comparable to the strongest FLUX.2-VAE baseline on CLIC 2020, while obtaining the best or near-best reconstruction quality on FFHQ among the compared VAEs. This shows that the proposed lightweight design preserves high visual fidelity across both general-domain and face-centric images.

\paragraph{Inference efficiency.}
\cref{tab:vae_evaluation_table} and Fig.~\ref{fig:vae_speed} further evaluate the efficiency. \magevae reduces the computational complexity of FLUX.2-VAE by about $12.3\times$ for encoding and $22.3\times$ for decoding. This translates into consistently lower latency and memory usage across all tested resolutions. The advantage becomes more pronounced at high resolution, where baseline VAEs become extremely slow or run out of memory at $4096\times4096$, while \magevae remains efficient for both encoding and decoding. These results demonstrate that \magevae provides a substantially better quality--efficiency trade-off, making it suitable for high-resolution generation, repeated editing, and interactive deployment.

\paragraph{Generation and editing.}
Beyond pixel-level reconstruction and efficiency, we further evaluate whether \magevae preserves the generation-ready latent structure of FLUX.2-VAE. Since \magevae is distilled from and regularized toward the latent distribution of the high-capacity but computationally expensive FLUX.2-VAE, the two tokenizers should remain compatible when used by downstream generators. Table~\ref{tab:vae_ablation} reports a cross-tokenizer ablation that swaps \magevae and FLUX.2-VAE while keeping the generation or editing backbone fixed. Replacing \magevae with FLUX.2-VAE in our Turbo models yields similar benchmark performance, and replacing FLUX.2-VAE with \magevae in FLUX.2-Klein-4B~\cite{flux-2-2025} also maintains comparable results. This suggests that \magevae not only reconstructs images well, but also preserves the latent geometry required by FLUX.2-style diffusion generators, allowing it to serve as an efficient substitute for the original FLUX.2-VAE in downstream generation and editing models.

\begin{promptbox}
\large
\textbf{\textit{Finding 1:}}
Anchor-latent supervision enables efficient VAE distillation without breaking the generation-ready latent space: by aligning a lightweight one-step encoder--decoder to the latent distribution of a strong but computationally expensive VAE, the tokenizer preserves reconstruction quality and cross-generator compatibility while substantially reducing encoding and decoding cost.
\end{promptbox}

\begin{table}[t]
\centering
\setlength{\tabcolsep}{3.5pt}
\caption{\textbf{Inference efficiency of packed CFG evaluation.}
By packing conditional and unconditional branches into a single batch, 
\mageflow{} reduces the classifier-free guidance overhead while preserving 
the original denoising trajectory. All configurations are benchmarked on a single NVIDIA A100 GPU.}
\label{tab:cfg_efficiency}
\begin{tabular}{l |c c c c}
\toprule
\textbf{Model} & Steps & Separate CFG & Packed CFG & Speedup \\
\midrule
\magebase & 30 & 7.5089s  & 6.5159s & 1.15$\times$\\
\mageflow & 20 & 5.0076s & 4.3680s &  1.15$\times$ \\
\mageeditbase & 30 & 11.5463s & 10.5582s &  1.09$\times$ \\
\mageedit & 30 & 11.5985s & 10.5475s & 1.10$\times$ \\
\bottomrule
\end{tabular}
\end{table}

\subsection{Native-Resolution MMDiT}
\label{sec:model_mmdit}

\subsubsection{Model Architecture and Native Packing}
\label{sec:model_mmdit_arch_packing}

The generative backbone of \mageflow is a 4B-parameter Native-Resolution Multimodal Diffusion
Transformer, as illustrated in Fig.~\ref{fig:nr_mmdit}(b). It follows the MMDiT block
design introduced by SD3~\citep{esser2024sd3}, where text and image tokens are concatenated and
processed by joint self-attention. Each block uses modality-specific normalization and projection
layers to preserve the distinct statistics of text and visual streams, while cross-modal
interaction is performed through self-attention over the combined sequence.

A key difference from standard MMDiT training is that NR-MMDiT does not restrict each optimization
step to a predefined resolution bucket. In conventional bucket-based training, images are assigned
to a finite set of fixed resolution and aspect-ratio buckets, and each step draws samples from only
one bucket so that all visual token grids share the same spatial shape. This simplifies batching,
but it discretizes the native resolution distribution, introduces bucket-quantization mismatch, and
limits the diversity of aspect ratios observed within each update. It also makes extremely wide or
tall outputs difficult to support unless corresponding buckets are explicitly added.
Inspired by NiT~\citep{wang2025nit}, we instead train directly on native-resolution sequences.
Images with different resolutions and aspect ratios are encoded by \magevae, flattened into
variable-length latent token sequences, and packed into a single contiguous batch under a fixed
token budget. We apply the same packing principle to text conditions: prompt embeddings of
different lengths are packed rather than padded to a common maximum length. With FlashAttention's
variable-length kernels~\citep{dao2023flashattention2,zadouri2026flashattention}, the packed text
and image sequences are processed using per-sample cumulative offsets, which restrict attention
within each sample without constructing explicit block-diagonal masks.

This native-resolution formulation has several practical advantages. First, samples are packed under a fixed token budget, so
training can naturally balance heterogeneous image sizes together with variable-length text
conditions. This improves batching flexibility and avoids unnecessary padding on the text side,
while allowing the image branch to preserve each sample's native latent grid.
Second, it removes the
single-bucket restriction during training: each optimization step can contain images with different
native resolutions and aspect ratios, rather than drawing all samples from one predefined
resolution bucket. This exposes the model to a richer and less discretized resolution distribution
within every update, and avoids the bucket-quantization mismatch introduced by mapping continuous
image sizes to a finite set of buckets. Therefore, a single checkpoint generalizes naturally to flexible output sizes at inference time. As shown in Fig.~\ref{fig:showcase}, both height and width can range from $512$ to $2048$, enabling not only standard square, portrait, and landscape outputs, but also extreme aspect ratios such as $512\times2048$ and $2048\times512$.

The same packing mechanism also improves inference efficiency. For classifier-free guidance, the conditional and unconditional branches can be packed together and evaluated in a single forward pass, avoiding the redundant computation introduced by separate CFG evaluation while preserving the original denoising trajectory. As shown in Table~\ref{tab:cfg_efficiency}, packed CFG consistently accelerates inference by 1.09$\times$--1.15$\times$ across  Mage-Flow and Mage-Edit variants. Therefore, native-resolution packing serves as both a training strategy for heterogeneous-resolution learning and an inference optimization for efficient CFG evaluation.

\begin{promptbox}
\large
\textbf{\textit{Finding 2:}}
Native-resolution packing turns resolution diversity into a training signal:
removing the single-bucket restriction allows each update to mix heterogeneous image sizes and
aspect ratios, improving resolution flexibility while also enabling efficient packed CFG inference.
\end{promptbox}

\begin{table}[t]
\centering
\setlength{\tabcolsep}{3.5pt}
\caption{\textbf{Training-efficiency ablation.}
All configurations are benchmarked on a single 8-GPU NVIDIA B200 node with
FlashAttention-4~\citep{zadouri2026flashattention}. The global batch size is $8$, with one packed
sample per GPU, and each sample contains a fixed packed sequence length of $50{,}000$ tokens.
We progressively replace FLUX.2-VAE with \magevae and enable fused CUDA kernels for \magevae, the
Qwen3-VL text encoder, and NR-MMDiT. Peak per-GPU memory, model FLOP utilization (MFU), per-step
wall-clock time, and relative speedup are reported.}
\label{tab:efficiency}
\begin{tabular}{l c c c c c c c}
\toprule
\textbf{Tokenizer} & \textbf{VAE Fuse} & \textbf{Text Fuse} & \textbf{DiT Fuse}
& \textbf{Mem. (GB)} & \textbf{MFU} & \textbf{Time (s)} & \textbf{Speedup} \\
\midrule
FLUX.2-VAE~\citep{flux-2-2025} & -- & -- & -- & 175.45 & 13.88\% & 1.9285 & $1.00\times$ \\
\magevae  & -- & -- & -- & 175.47 & 17.44\% & 1.3647 & $1.41\times$ \\
\magevae  & \checkmark & -- & -- & 175.47 & 17.41\% & 1.3609 & $1.42\times$ \\
\magevae  & \checkmark & \checkmark & -- & 175.47 & 17.88\% & 1.3258 & $1.45\times$ \\
\magevae  & \checkmark & \checkmark & \checkmark & 141.44 & 29.28\% & 0.7775 & $2.48\times$ \\
\bottomrule
\end{tabular}
\end{table}

\subsubsection{Conditioning and rectified-flow training}
\label{sec:model_mmdit_conditioning}

Text conditioning is provided by a frozen Qwen3-VL-4B-Instruct~\citep{qwen3vl} text encoder, which maps each prompt into contextual embeddings $\tau$. Given an image $x$, \magevae encodes it into latents $z=\magevae(x)$ at a $16\times$ spatial reduction with 128 channels. These latents are flattened into visual tokens and linearly projected to the NR-MMDiT hidden dimension. The model is trained in the \magevae latent space with the rectified
flow-matching objective~\citep{esser2024sd3},
\[
\mathcal{L}(\theta)
=
\mathbb{E}_{(x,\tau),t,\epsilon}
\left[
\left\|
v_\theta(z_t,t,\tau) - (z-\epsilon)
\right\|_2^2
\right],
\]
where
\[
z_t=(1-t)z+t\epsilon,\qquad \epsilon\sim\mathcal{N}(0,I).
\]

The same rectified-flow objective and NR-MMDiT backbone are used for both generation and editing, but with different conditioning inputs. For \mageflow, Qwen3-VL encodes the text prompt into $\tau$, and NR-MMDiT denoises the target latent tokens conditioned on $\tau$. For \mageedit, Qwen3-VL encodes the editing instruction together with the source image into multimodal conditioning embeddings $\tau$, while \magevae encodes the source and target images into $z^{\mathrm{src}}$ and $z^{\mathrm{tgt}}$. Thus, the NR-MMDiT input sequence concatenates $\tau$, $z^{\mathrm{src}}$, and the noisy target latent tokens.

To distinguish source and target visual tokens, \mageedit extends the 2D rotary positional embedding (RoPE) used in \mageflow with an additional frame dimension. Each visual token is assigned a 3D position index $(h,w,f)$, where $(h,w)$ denotes its native spatial location and $f$ denotes the image index, covering all source images and the target image. This frame-aware RoPE preserves source-target spatial correspondence while indicating token roles inside the shared attention sequence. The loss is computed only on target tokens, so \mageedit can be initialized directly from \mageflow without adding separate editing modules.

\subsection{Training Infrastructure}
\label{sec:training_infra}

The training efficiency of the \mageflow stack is determined by three repeatedly executed modules: the convolutional diffusion blocks in \magevae, the Transformer blocks in the frozen Qwen3-VL text encoder, and the 4B NR-MMDiT blocks. Although their main arithmetic comes from convolutions, matrix multiplications, and attention, each block also contains many memory-bound operator chains, including normalization, adaptive modulation, RoPE application, gating, activation, and residual addition. When launched as separate CUDA kernels, these operators repeatedly read and write large activation tensors, causing substantial memory traffic and kernel-launch overhead.

%The training efficiency of the \mageflow stack is determined by three repeatedly executed modules: the convolutional diffusion blocks in \magevae, the Transformer blocks in the frozen Qwen3-VL text encoder, and the 4B NR-MMDiT blocks. Although their dominant arithmetic comes from convolutions, matrix multiplications, and attention, each block also contains many memory-bound operator chains, including normalization, adaptive modulation, RoPE application, gating, activation, and residual addition. When executed as separate CUDA kernels, these operations repeatedly read and write large activation tensors, leading to substantial memory traffic and kernel-launch overhead.

To reduce this overhead, we fuse the dominant operator chains inside the repeated blocks of the full stack. In \magevae, we fuse normalization--activation--residual chains in the convolutional diffusion blocks. In the Qwen3-VL text encoder and NR-MMDiT, we fuse common Transformer-side operations such as adaptive normalization, rotary embedding application, and gated residual updates. These fused kernels keep intermediate values in on-chip memory and write back only the final outputs, reducing both activation memory movement and kernel launches. Because these blocks are executed many times during each forward and backward pass, local operator fusion translates into a substantial stack-level throughput gain.

%To reduce this overhead, we fuse the dominant operator chains inside the repeated blocks of the full stack. In \magevae, we fuse normalization--activation--residual chains in the convolutional diffusion blocks. In Qwen3-VL and NR-MMDiT, we fuse common Transformer-side operations such as adaptive normalization, rotary embedding application, and gated residual updates. These fused kernels keep intermediate values in on-chip memory and write back only the final outputs, reducing both activation memory movement and kernel launches. Since these blocks are executed many times during each forward and backward pass, local operator fusion translates into a large stack-level throughput gain.

As reported in \tableautorefname~\ref{tab:efficiency}, replacing FLUX.2-VAE with \magevae already reduces per-step time from $1.9285$\,s to $1.3647$\,s, giving a $1.41\times$ speedup and showing the importance of a lightweight tokenizer. Fusing \magevae and Qwen3-VL brings additional but moderate gains, while fusing the repeated NR-MMDiT blocks provides the largest improvement because the 4B diffusion backbone dominates the training step. Overall, the full system increases MFU from $13.88\%$ to $29.28\%$, reduces peak per-GPU memory from $175.45$\,GB to $141.44$\,GB, and improves per-step training speed by $2.48\times$. These results show that efficient native-resolution training requires not only an efficient tokenizer and architecture, but also stack-level kernel optimization that removes memory-bound overhead from the repeated blocks.

\begin{promptbox}
\large
\textbf{\textit{Finding 3:}}
Efficient native-resolution generation requires co-designing the model and the training system:
lightweight tokenization reduces the arithmetic cost, while stack-level kernel fusion removes
memory-bound overhead that would otherwise dominate repeated VAE, text-encoder, and MMDiT blocks.
\end{promptbox}

\section{Data Collection and Curation}
\label{sec:data}

\begin{figure*}[t]
\centering
\includegraphics[width=\linewidth]{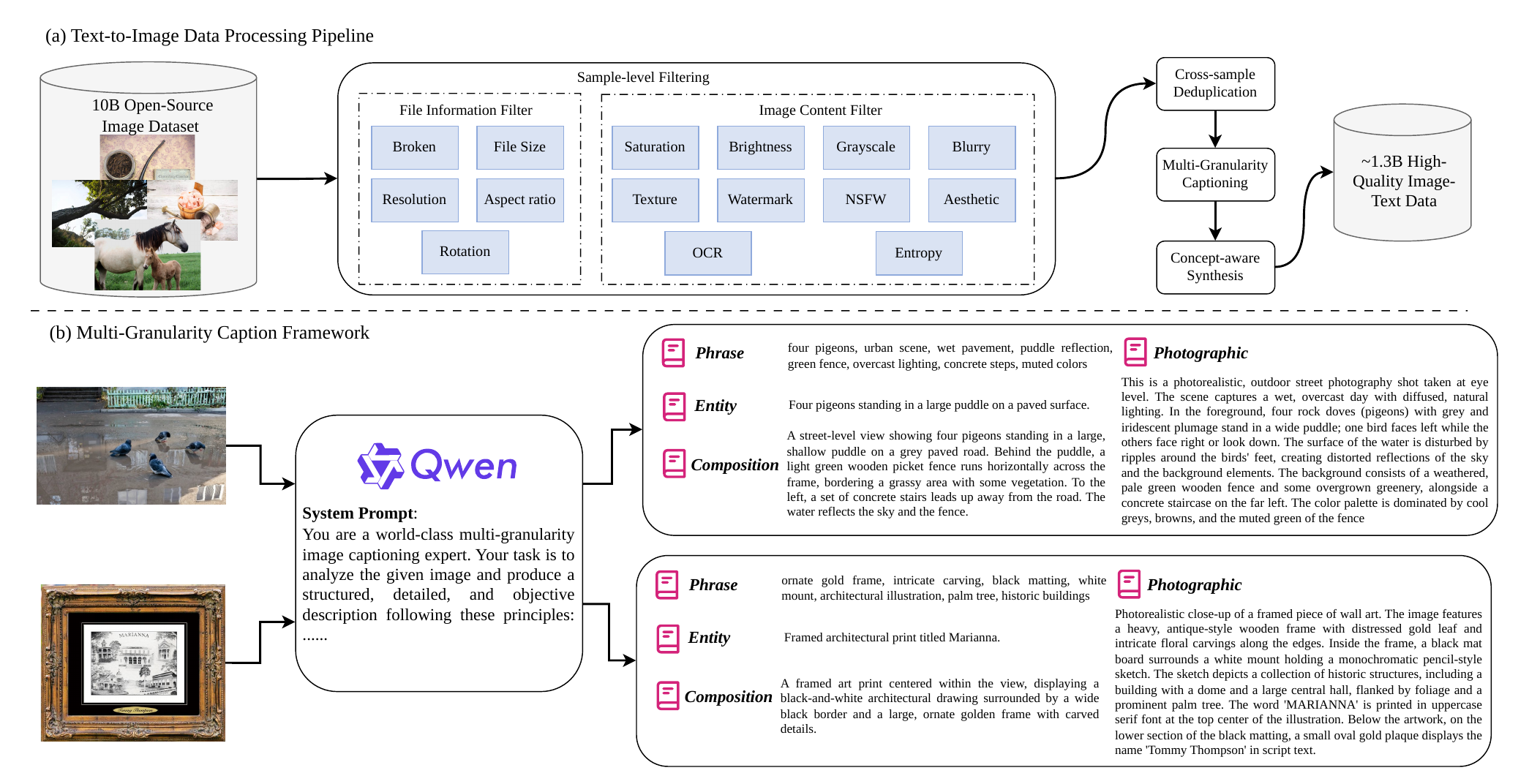}
\caption{\textbf{Text-to-image data processing and captioning pipeline.}
(a) Raw web image--text pairs are processed by sample-level filtering, cross-sample deduplication, multi-granularity captioning, and concept-aware synthesis, resulting in
a high-quality image--text corpus. (b) The multi-granularity captioning framework uses Qwen3-VL to produce phrase-level, entity-level, composition-level, and photographic captions, providing training prompts with different levels of semantic detail and visual specificity.}
\label{fig:filter}
\end{figure*}

The \mageflow stack relies on two complementary data pipelines: one for text-to-image generation and one for instruction-based image editing. The generation pipeline curates large-scale image--text pairs into high-quality and balanced prompt--image supervision, while the editing pipeline constructs and filters source-image, instruction, and target-image triples. Both pipelines are designed to improve visual quality, safety, semantic alignment, diversity, and coverage of capability-specific skills that are under-represented in raw web data.

\subsection{Text-to-Image Data Collection and Filtering}
\label{sec:data_t2i}

The \mageflow generation corpus is built from roughly $10$B raw image--text pairs collected from large-scale open-source datasets. As shown in Fig.~\ref{fig:filter}, the curation pipeline contains four major stages: sample-level filtering, cross-sample deduplication, multi-granularity captioning, and concept-aware synthesis. The sample-level filters remove corrupted, low-quality, unsafe, or visually unsuitable images; deduplication suppresses near-duplicate visual modes; multi-granularity captioning standardizes textual supervision; and concept-aware synthesis supplements long-tail images. After curation, approximately $1.3$B high-quality image--text pairs are retained, from which the stage-wise pre-training subsets are sampled.

\paragraph{Sample-level filtering.}
We first filter each image--caption pair independently using the file-information and image-content filters shown in Fig.~\ref{fig:filter}(a). File-information filters remove corrupted or near-empty files, images with insufficient resolution or pixel count, extreme aspect ratios, and samples with incorrect orientation metadata. Image-content filters then score decoded images for visual quality and safety: brightness and saturation filters remove over-exposed, under-exposed, or unnaturally saturated images; grayscale and blurry filters remove near-monochrome or low-sharpness samples; entropy and texture filters suppress near-empty images and texture-like non-semantic patterns; and watermark, aesthetic, OCR, and NSFW filters remove watermarked, low-quality, document-like, or unsafe images. Many filters are threshold-based rather than binary. As shown in \cref{tab:filter_stages}, the thresholds are progressively tightened across the $256^2$, $512^2$, $1024^2$, and SFT stages so that early training preserves broad coverage, while later stages focus on cleaner and higher-quality data.

\begin{table}[t]
\centering
\caption{\textbf{Sample-level filtering thresholds across pre-training and SFT stages.}
Thresholds are progressively tightened from the $256^2$ stage to the $1024^2$ and SFT stages.
Early stages retain broad visual coverage, while later stages emphasize higher resolution,
stronger aesthetics, lower watermark probability, and cleaner image content.}
\label{tab:filter_stages}
\small
\setlength{\tabcolsep}{6.5pt}
\begin{tabular}{l c c c c}
\toprule
 & $\mathbf{256^2}$ & $\mathbf{512^2}$ & $\mathbf{1024^2}$ & \textbf{SFT} \\
\midrule
Pixel count $h\times w$        & $\geq 256^2$      & $\geq 512^2$      & $\geq 1024^2$     & $\geq 1024^2$     \\
$\min(h, w)$                   & $\geq 128$        & $\geq 256$        & $\geq 512$        & $\geq 512$        \\
Aspect ratio                   & $[0.1, 10.0]$     & $[0.1, 10.0]$     & $[0.1, 10.0]$     & $[0.1, 10.0]$     \\
File size                      & $\geq 1$\,KB      & $\geq 1$\,KB      & $\geq 1$\,KB      & $\geq 1$\,KB      \\
NSFW score                     & $\leq 0.1$        & $\leq 0.1$        & $\leq 0.1$        & $\leq 0.1$        \\
Watermark score                & $< 0.5$           & $< 0.3$           & $< 0.1$           & $< 0.05$          \\
Aesthetic-V2.5 score           & $\geq 4.5$        & $\geq 5.5$        & $\geq 6.0$        & $\geq 6.5$        \\
OCR text-area ratio            & $\leq 0.3$        & $\leq 0.3$        & $\leq 0.3$        & $\leq 0.3$        \\
OCR num.\ regions              & $\leq 5$          & $\leq 5$          & $\leq 5$          & $\leq 5$          \\
\bottomrule
\end{tabular}
\end{table}

\paragraph{Cross-sample deduplication.}
Web-scale data is highly redundant both within individual sources and across datasets. After sample-level filtering, we deduplicate images at two levels. Each retained image is encoded into an SSCD copy-detection descriptor~\citep{pizzi2022self}, which is robust to re-encoding, cropping, resizing, and light edits. All descriptors are indexed with FAISS \cite{douze2024faiss} for efficient nearest-neighbor search. Within each dataset, image pairs with cosine similarity above $0.9$ are grouped as duplicates, and only the highest-quality representative is retained. Very large clusters are capped to suppress repeated web templates such as stock photos, banners, and product layouts. Across datasets, we maintain a persistent descriptor index of already accepted images and reject new samples that reproduce existing ones above the same similarity threshold. We also match against a held-out benchmark index to reduce evaluation contamination. Together, these steps improve diversity and prevent duplicated sources from dominating the training distribution.

\paragraph{Multi-granularity captioning.}
To standardize textual supervision, we caption retained images with Qwen3-VL-32B-Instruct~\citep{qwen3vl}. Following \citep{LongCat-Image}, each image is assigned captions at multiple levels: a phrase-level description for concept statistics, an entity-level caption describing major objects and attributes, a composition-level caption describing spatial layout and relations, and a photographic caption describing style, lighting, viewpoint, atmosphere, and fine visual details. During training, the model samples from these descriptive caption channels so that it learns to follow prompts with different lengths and specificity. For text-rich images, the captioner is explicitly prompted to recognize visible text and convert it into rendering instructions.

\begin{figure*}[t]
\centering
\captionsetup[subfigure]{justification=centering,singlelinecheck=true}

\begin{subfigure}[t]{0.48\textwidth}
    \centering
    \includegraphics[width=0.95\linewidth]{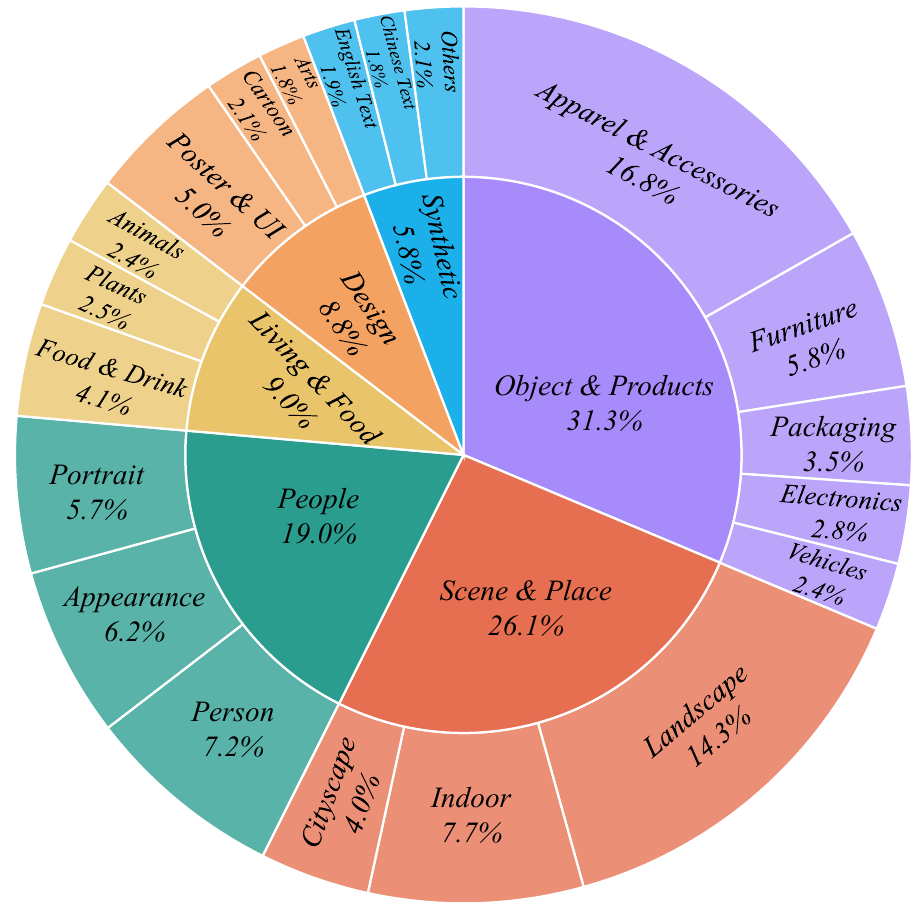}
    \caption{Generation pre-training data.}
    \label{fig:pretrain_data_gen}
\end{subfigure}
\hfill
\begin{subfigure}[t]{0.48\textwidth}
    \centering
    \includegraphics[width=0.95\linewidth]{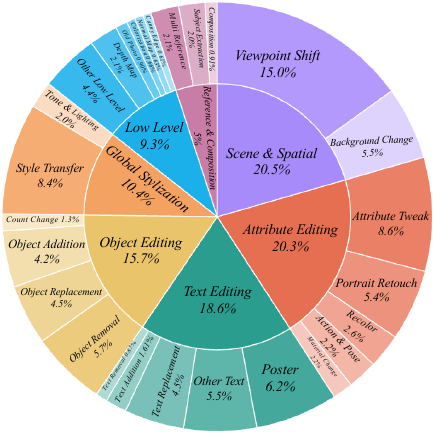}
    \caption{Editing pre-training data.}
    \label{fig:pretrain_data_edit}
\end{subfigure}

\caption{\textbf{Data composition for generation and editing pre-training.}
(a) Concept distribution of the curated generation corpus after merging filtered web data with targeted supplemental data. (b) Final composition of the editing pre-training mixture after adjusting sampling rates across constituent editing datasets.}
\label{fig:pretrain_data_composition}
\end{figure*}

\paragraph{Concept-aware synthesis and balancing.}
Although large-scale web data provides broad visual coverage, it remains sparse in several capability-critical areas, including long-text rendering, rare objects, uncommon attributes, structured layouts, and under-represented styles. Thus, we construct targeted supplemental data for these long-tail concepts, including synthetic text-rendering samples with diverse fonts, layouts, languages, colors, and backgrounds, as well as additional image--text pairs covering rare concepts and compositional cases. All supplemental samples are passed through the same filtering and quality-control pipeline before being merged into the curated pre-training corpus. Then, we use phrase-level captions to estimate the concept distribution of the combined corpus, as shown in Fig.~\ref{fig:pretrain_data_composition}(a). The resulting distribution remains long-tailed: \textit{Object \& Products} and \textit{Scene \& Place} form the two largest coarse domains, while \textit{Design} and \textit{Synthetic} provide important coverage for layout, product-style, poster-style, and text-rendering capabilities. During training, we apply concept-aware sampling to reduce the dominance of frequent objects, natural scenes, and common photorealistic styles. Across training stages, we progressively tighten filtering thresholds and strengthen reweighting, moving from broad visual-prior learning in early stages to cleaner and more capability-focused learning in later stages.

\subsection{Image-Editing Data Collection and Filtering}
\label{sec:data_edit}
\begin{figure*}[t]
\centering
\includegraphics[width=\linewidth]{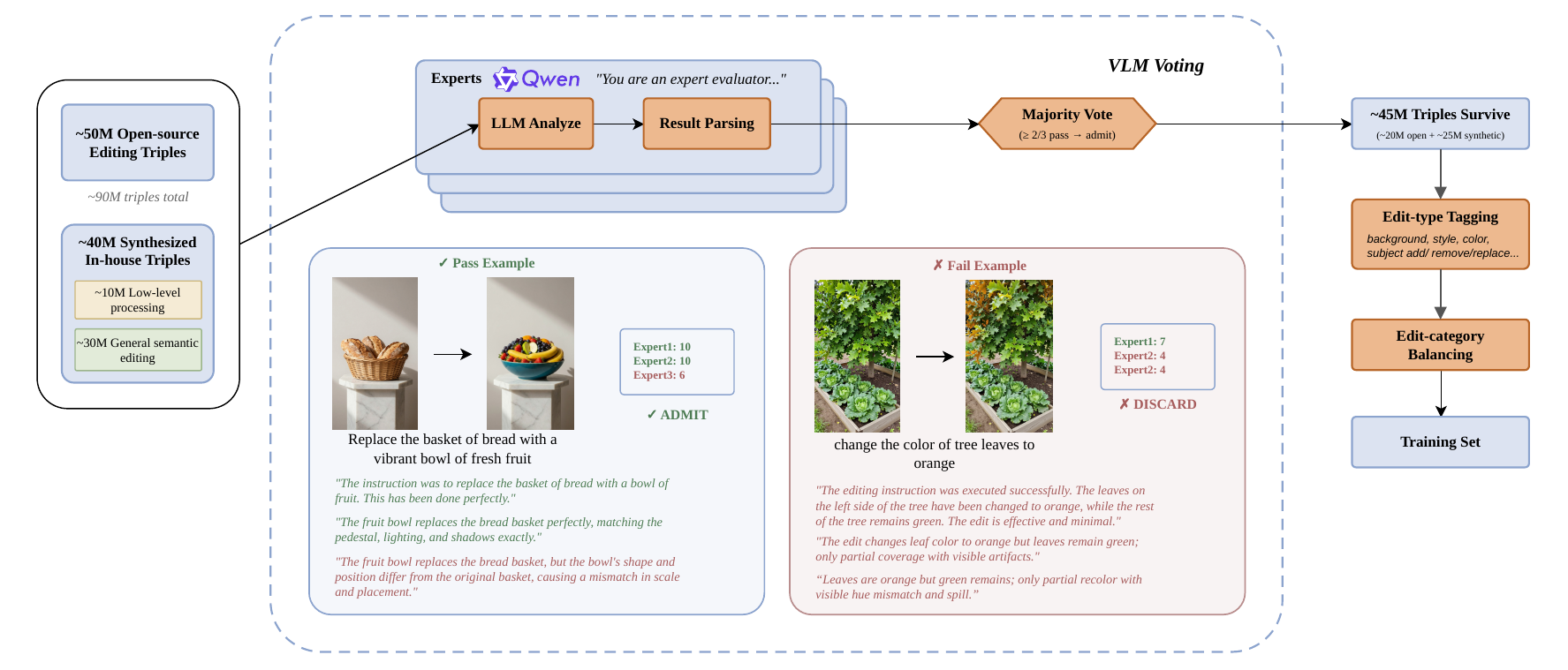}
\caption{\textbf{Editing data filtering pipeline.} Raw editing triples are collected from open-source editing datasets and in-house synthesis. Each triple is evaluated by three Qwen3.5-9B experts with different system prompts and partially overlapping rubrics. Each expert analyzes the source image, target image, and edit instruction, and its reasoning output is parsed into a pass/fail decision. A triple is retained only if it receives a majority vote from the three experts. The surviving 45M triples are then assigned to a manually defined edit-type taxonomy and reweighted across categories to form the final editing training set.}
\label{fig:edit_filter}
\end{figure*}

The \mageedit corpus consists of (source image, edit instruction, target image) triples. As shown in Fig.~\ref{fig:edit_filter}, the raw pool contains roughly $90$M triples from two complementary sources: about $50$M triples aggregated from open-source instruction-based editing datasets and about $40$M triples synthesized in-house. The synthesized split contains approximately $10$M low-level image-processing triples and $30$M general semantic-editing triples. Then, we apply a VLM-based voting filter, followed by edit-type tagging and category balancing, to construct the final editing training set.

\paragraph{Editing data synthesis.}
The in-house semantic-editing subset is constructed to cover a broad range of user-facing editing skills, including background replacement, color and material modification, tone and style transfer, subject addition, removal, and replacement, object-count change, motion change, viewpoint change, text editing, portrait retouching, old-photo restoration, and global adjustment. Each edit type is generated with a type-specific pipeline that combines off-the-shelf generation, inpainting, segmentation, image processing, and template-based instruction generation. This provides explicit source--target pairings and clear editing instructions, while supplementing edit categories that are sparse or unreliable in open-source datasets.

\paragraph{VLM-based dataset filtering.}
Raw editing triples often contain instruction-inconsistent or visually degraded samples. For example, the target image may fail to apply the requested edit, modify irrelevant regions, change the source identity or layout unnecessarily, or introduce visible artifacts. Therefore, we evaluate each triple with three independent Qwen3.5-9B experts~\cite{qwen3.5}. Each expert is configured with a different system prompt and evaluates a partially overlapping set of criteria, so that the three judgments are complementary rather than identical. Given the source image, target image, and edit instruction, each expert checks whether the requested edit is correctly executed, whether unrelated regions are preserved, and whether the edited result remains visually plausible. The reasoning output is parsed into a criterion-level assessment and then converted into a pass/fail decision using a predefined threshold. We retain a triple only when at least two of the three experts vote to pass. After filtering, roughly $20$M open-source triples and $25$M synthesized triples survive, forming a $45$M-triple retained pool for editing training.

\paragraph{Edit-type tagging and balancing.}
We manually define a taxonomy of 19 edit categories that covers all retained editing data. Instead of tagging each sample independently with a VLM, we determine the annotation unit from the organization of each data source. If a dataset contains a consistent editing operation, the whole dataset is treated as one unit. If a dataset is organized into semantically consistent sub-datasets, each sub-dataset is treated as a separate unit. If an explicit edit-type field is available, samples sharing the same field value form one unit. We then manually map every unit to one of the 19 edit categories, ensuring that each retained sample is assigned to the unified taxonomy. During balancing, we adjust the sampling rate of each constituent dataset and edit category so that the aggregate training mixture maintains broad and well-proportioned coverage across the taxonomy. This prevents frequent operations from dominating the gradient signal while keeping rare but important edit types sufficiently represented. The resulting editing pre-training composition is shown in Fig.~\ref{fig:pretrain_data_composition}(b).

\section{Training}
\label{sec:training}

\begin{figure*}[t]
\centering
\includegraphics[width=\linewidth]{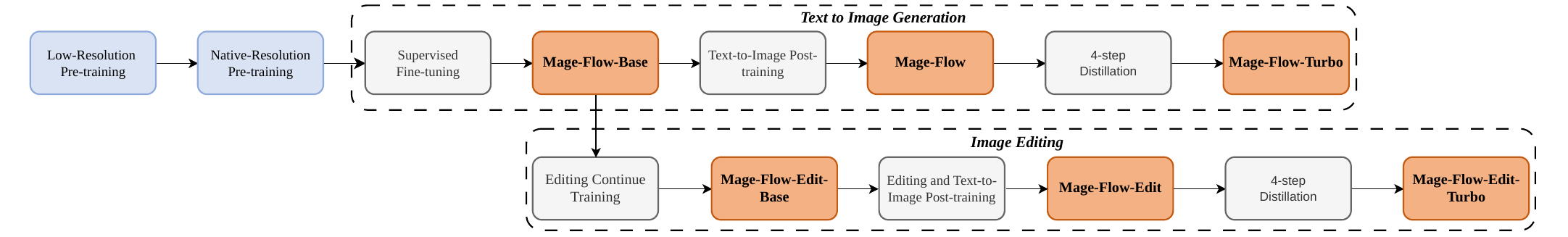}
\caption{\textbf{Overview of the \mageflow training pipeline.}
The shared backbone is first trained with progressive text-to-image pre-training, summarized as low-resolution pre-training followed by native-resolution pre-training, and then supervised fine-tuning produces the base generation checkpoint \magebase. From this checkpoint, the text-to-image branch applies Diffusion-NFT post-training to obtain \mageflow and 4-step distillation to obtain \mageturbo. The editing branch forks from \magebase: continued source-conditioned editing training produces \mageeditbase, mixed editing-and-generation post-training produces \mageedit, and 4-step distillation produces \mageeditturbo.}
\label{fig:training_pipeline}
\end{figure*}

We train the \mageflow model family with a unified recipe shared by text-to-image generation and instruction-based editing. Both tasks operate in the same \magevae latent space and use the same 4B Native-Resolution MMDiT backbone with a rectified-flow objective. The main differences are the conditioning format and data mixture: text-to-image generation is conditioned on prompts, while
editing is conditioned on editing instructions and source image(s). As shown in
Fig.~\ref{fig:training_pipeline}, we organize the recipe by training stage: progressive pre-training and supervised fine-tuning produce the base checkpoint, Diffusion-NFT post-training produces the aligned models, and few-step distillation produces the Turbo variants.

\subsection{Pre-training and Supervised Fine-tuning}
\label{sec:training_base}

\paragraph{Text-to-image generation.}
The generation model is trained with a progressive pre-training and SFT curriculum. As shown in \tableautorefname~\ref{tab:filter_stages}, pre-training contains three stages. First, we train on $1.2$B filtered and recaptioned image--text pairs at a fixed $256\times256$ resolution to learn broad visual--language alignment at low computational cost. Second, we move to a higher-quality $600$M subset under a $512$-pixel native-aspect-ratio regime, where each sample keeps an approximately $512\times512$ pixel budget while preserving its original aspect ratio. Third, we train on an even cleaner $300$M subset under the $1024$-pixel native-aspect-ratio regime to improve fine details, layout fidelity, text rendering, and aesthetic quality. All stages use the same rectified-flow objective in the \magevae latent space, while progressively increasing resolution, data quality, and concept-aware reweighting strength. We then perform supervised fine-tuning on a curated $150$M high-quality subset at the $1024$-pixel native-aspect-ratio regime, using stricter aesthetic, alignment, watermark, OCR, and duplication filters together with increased weights for capability-targeted data. This produces the text-to-image base checkpoint, \textbf{\magebase}.

\begin{figure*}[t]
\centering
\includegraphics[width=\linewidth]{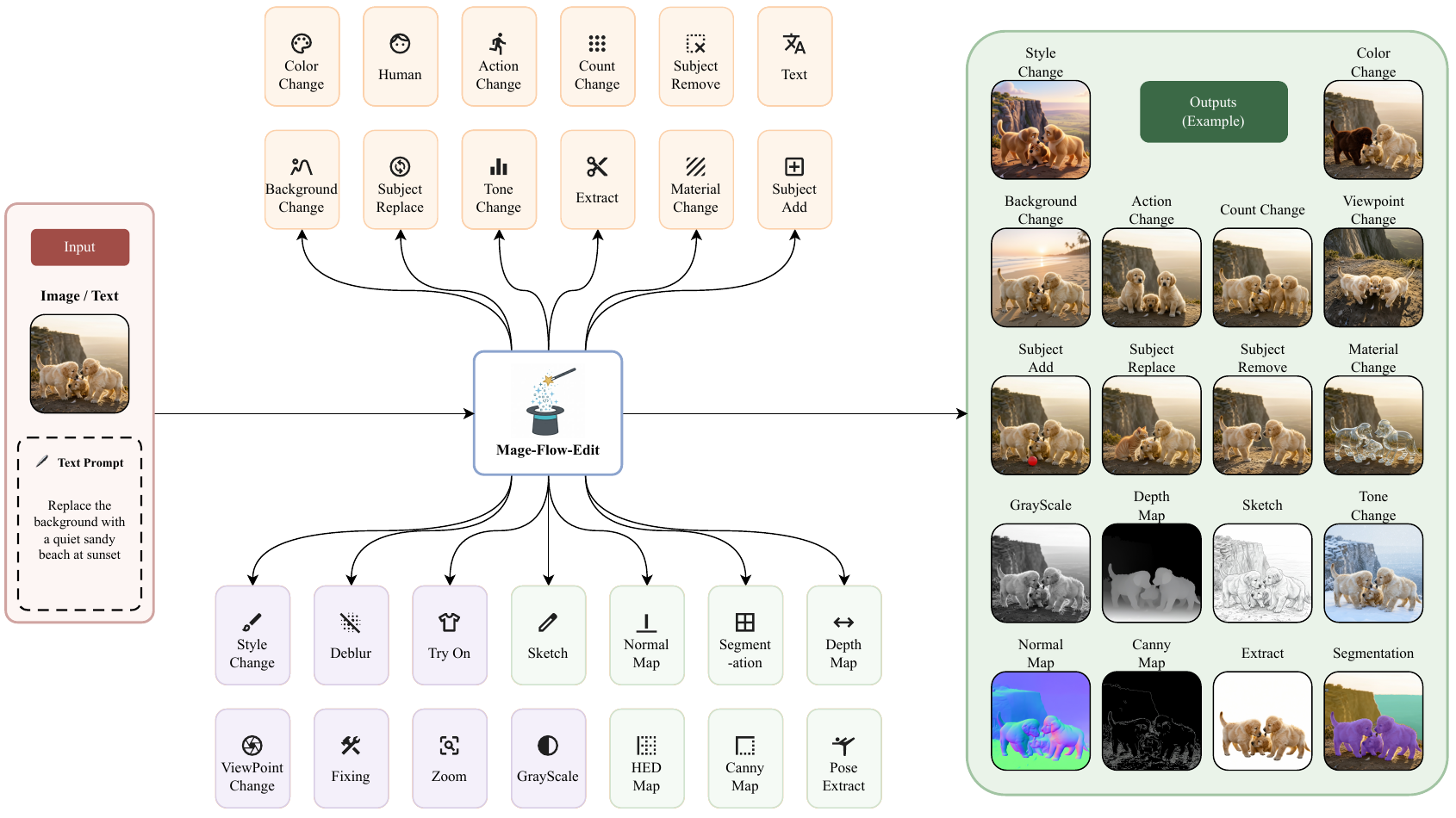}
\caption{\textbf{Unified image-editing capabilities and representative outputs.} \mageedit supports semantic content editing, appearance transformation, image restoration, and structure-aware outputs within a unified image-and-text-conditioned model. The displayed prompt illustrates background replacement, while the remaining outputs correspond to task-specific instructions or output requests.}
\label{fig:edit_diversity}
\end{figure*}

\paragraph{Instruction-based editing.}
The editing model is initialized from \magebase and trained with a two-stage editing adaptation recipe. In the first stage, we train on a balanced mixture of $35$M editing triples and $35$M generation pairs, adapting the model to source-conditioned instruction following while preserving the visual prior and open-ended synthesis capability inherited from \magebase. In the second stage, we further train on a higher-quality mixture of $20$M editing triples and $10$M generation pairs to improve editing fidelity and robustness. In both stages, multi-image editing examples account for no more than $0.5\%$ of the editing data, while the majority are single-image edits. This two-stage adaptation produces the editing base checkpoint, \textbf{\mageeditbase}.
Fig.~\ref{fig:edit_diversity} illustrates the breadth of editing capabilities learned through this adaptation recipe. Given the same source image, \mageedit can follow diverse text instructions to produce semantic edits, appearance transformations, restoration results, and structure-aware outputs, demonstrating that the editing checkpoint supports a unified one-to-many editing interface rather than a collection of task-specific models.

\subsection{Diffusion-NFT Post-training}
\label{sec:training_nft}

Starting from the Base checkpoints, we apply Diffusion-NFT~\citep{zheng2025diffusionnft} as the post-training stage for both generation and editing. Diffusion-NFT operates directly on the forward process of flow-matching generators and performs negative-aware fine-tuning with online samples, requiring no likelihood estimation and remaining compatible with arbitrary black-box samplers. We use the same Diffusion-NFT objective for \mageflow and \mageedit, while adapting the condition format, data mixture, and reward models to each task.

\paragraph{Text-to-image generation.}
For \mageflow, we curate a compact RL prompt pool of approximately $20$K prompts spanning three capability groups: about $10$K text-rendering prompts, $4$K aesthetic-quality prompts, and $6$K semantic-understanding prompts. Each prompt is assigned a capability tag that determines both its evaluation rubric and its reward evaluator. Text-rendering prompts are scored by PaddleOCR-VL-1.5~\citep{cui2026paddleocr}, which reads the generated image and compares the recognized strings against the target text to measure OCR fidelity, scene text, typography, and text--object interactions. Aesthetic-quality and semantic-understanding prompts are scored by Qwen3.5-27B~\citep{qwen3.5} with two different rubric system prompts. The aesthetic rubric evaluates photographic quality, lighting, composition, color harmony, texture, and artistic style, while the semantic rubric decomposes prompts into checkable questions covering compositional reasoning, object attributes, spatial relations, counting, actions, and multi-concept scenes. Details of the reward evaluators, rubrics, and OCR scoring formula are provided in Appendix~\ref{app:reward_models}.

\begin{figure}[t]
\centering
\includegraphics[width=\linewidth]{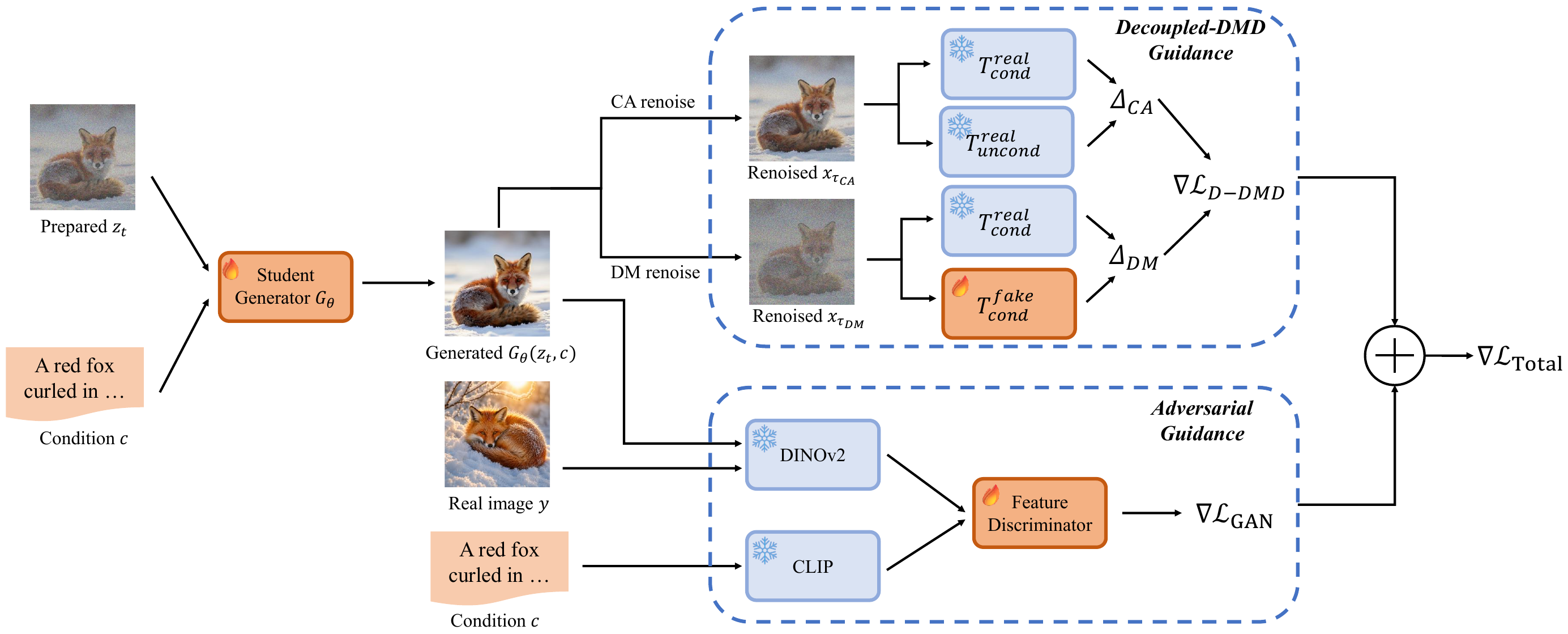}
\caption{\textbf{Few-step distillation framework.}
The student maps a prepared noised latent $z_t$ and condition $c$ to a predicted clean sample. Decoupled DMD re-noises the student output at two independent noise levels: the classifier-free augmentation branch queries the frozen teacher with and without the condition to form $\Delta_{\mathrm{CA}}$, while the distribution-matching branch contrasts the teacher with a trainable fake-score model to form $\Delta_{\mathrm{DM}}$. In parallel, generated and real images are encoded by frozen vision foundation models such as DINOv2 and CLIP, and a lightweight feature discriminator provides the adversarial gradient $\nabla\mathcal{L}_{\mathrm{GAN}}$. The Decoupled DMD and adversarial gradients are combined to train the 4-step Turbo student.}
\label{fig:distill_arch}
\end{figure}

Optimization is performed with online rollout groups. We use a global batch size of $48$, and each optimizer step contains prompts from all three capability groups. Although a step mixes heterogeneous prompts, each prompt is routed to exactly one evaluator according to its capability tag; its reward is computed only by that evaluator and is never summed or averaged across capabilities. For each prompt, we sample a group of candidates from the current generator using $10$ denoising steps with guidance scale $5.0$, and score the candidates with the assigned evaluator. Since different evaluators have different score distributions, advantages are normalized separately within each reward type: candidates scored by one evaluator are normalized only against candidates from the same evaluator. This per-type normalization produces the optimality probabilities \(r_i^{(s)}\in[0,1]\) used by the Diffusion-NFT loss.

We use a two-stage schedule that shifts the capability mixture over time while keeping all three groups present throughout. Stage~1 runs for $140$ optimizer steps on a balanced mixture \(\mathcal{P}_{\text{aes}}{:}\mathcal{P}_{\text{text}}{:}\mathcal{P}_{\text{sem}}=1{:}1{:}1\), with text-rendering prompts that emphasize stable OCR cases including single words, short phrases, simple signs, logos, and short scene text, so that the model sharpens character- and word-level rendering while jointly improving visual quality and semantic alignment. Stage~2 then continues from the best Stage-1 checkpoint for a further $60$ steps, up-weighting the harder text-rendering prompts to \(\mathcal{P}_{\text{aes}}{:}\mathcal{P}_{\text{text}}{:}\mathcal{P}_{\text{sem}}=2{:}4{:}1\) and targeting complete sentences, dense captions, multi-line layouts, punctuation-rich text, and text embedded in complex scenes; retaining aesthetic and semantic prompts throughout prevents over-specialization to OCR and preserves the general generation quality built up in the first stage. Together, the two stages produce the aligned generation checkpoint, \textbf{\mageflow}.

\paragraph{Instruction-based editing.}
For \mageedit, we apply Diffusion-NFT on top of \mageeditbase using a joint stream of editing and generation data. This sharpens source-conditioned instruction following while preserving the generation prior. The two streams are interleaved at a $4{:}1$ ratio of four editing updates per generation update, and share the same Diffusion-NFT objective, global batch size, and optimizer as the text-to-image run.

The generation stream reuses the text-to-image RL prompt pool and its capability-routed rewards. This stream helps preserve the text-rendering, compositional, and open-ended generation ability of \mageedit during editing post-training. The editing stream draws instructions from a curated editing RL pool of approximately $30$K prompts, uniformly sampled across edit tasks in the editing pre-training corpus, including object replacement, object removal, background replacement, spatial editing, style transfer, and general instruction-based editing. This prevents any single edit type from dominating the update. Each editing rollout is scored by RationalRewards~\citep{wang2026rationalrewards}, a reasoning reward model that first produces a multi-dimensional critique and then emits a scalar preference. The critique evaluates four aspects: instruction adherence, preservation of source content outside the edited region, physical and perceptual plausibility, and the quality of rendered text. The four aspect scores are averaged into a normalized reward in $[0,1]$. The joint post-training runs for $300$ optimizer steps and yields the aligned editing checkpoint, \textbf{\mageedit}.

\paragraph{Diffusion-NFT objective.}
For both tasks, each condition \(c\) is used to sample online candidates from the current generator.
The candidates are scored by the task-specific rubric and reward evaluator, and the raw rewards are
normalized within each prompt group to obtain an optimality probability
\(r_i^{(s)}\in[0,1]\), where larger values indicate higher-quality samples. Diffusion-NFT then
optimizes a reward-weighted flow-matching objective with implicit positive and negative policies:
\[
\mathcal{L}_{\mathrm{NFT}}^{(s)}(\theta)
=
\mathbb{E}_{c,\,x_{0,i}\sim\pi_{\mathrm{old}}(\cdot|c),\,t}
\left[
\underbrace{
r_i^{(s)}
\left\|
v_{\theta}^{+}(x_{i,t},t,c)-v_{i,t}
\right\|_2^2
}_{\text{positive match}}
+
\underbrace{
\left(1-r_i^{(s)}\right)
\left\|
v_{\theta}^{-}(x_{i,t},t,c)-v_{i,t}
\right\|_2^2
}_{\text{negative match}}
\right],
\]
where \(v_{i,t}\) denotes the target velocity of the forward process, and
\(v_{\theta}^{+}\) and \(v_{\theta}^{-}\) denote the implicit positive and negative policies
defined by Diffusion-NFT. The positive branch pulls the model toward high-reward samples, while
the negative branch suppresses low-reward samples. For \mageflow, \(c\) is a text prompt and the
reward comes from the rubric-selected generation evaluator. For \mageedit, \(c\) can be either a
generation prompt or an editing condition; the former uses the same generation reward evaluators,
while the latter uses RationalRewards on the source image, instruction, and edited result.

\begin{table*}[t]
\centering
\caption{\textbf{Effect of adversarial perceptual guidance in 4-step distillation.}
Adversarial guidance consistently improves generation and benefits text editing under the highly
compressed four-step trajectory, while changes on general editing benchmarks are mixed. Higher
scores indicate better performance.}
\label{tab:adv_guidance_ablation}

\textbf{(a) Text-to-image generation.}

\vspace{2pt}
\begin{adjustbox}{width=\textwidth}
\begin{tabular}{l c c c c c c c c c}
\toprule
\textbf{Model}
& \textbf{GenEval}
& \textbf{DPG-Bench}
& \textbf{TIIF-Short}
& \textbf{TIIF-Long}
& \textbf{CVTG-2K}
& \textbf{OneIG-EN}
& \textbf{OneIG-CN}
& \textbf{LongText-EN}
& \textbf{LongText-CN} \\
\midrule
\mageturbo
& 0.88 & 85.48 & 83.58 & 84.16
& 0.873 & 0.523 & 0.491
& 0.911 & 0.801 \\
\mageturbo w/o Adv.
& 0.89 & 85.37 & 80.99 & 82.03
& 0.847 & 0.518 & 0.486
& 0.882 & 0.783 \\
\bottomrule
\end{tabular}
\end{adjustbox}

\vspace{8pt}

\textbf{(b) Instruction-based editing.}

\vspace{2pt}
\begin{adjustbox}{width=\textwidth}
\begin{tabular}{l c c c c c}
\toprule
\textbf{Model}
& \textbf{ImgEdit}
& \textbf{GEdit-EN}
& \textbf{GEdit-CN}
& \textbf{TextEdit-Syn}
& \textbf{TextEdit-Real} \\
\midrule
\mageeditturbo
& 4.38 & 8.271 & 8.264 & 12.77 & 15.41 \\
\mageeditturbo w/o Adv.
& 4.29 & 8.003 & 8.025 & 11.64 & 14.77 \\
\bottomrule
\end{tabular}
\end{adjustbox}
\end{table*}

\subsection{Few-step Distillation}
\label{sec:training_distill}
To reduce inference cost, we distill the RL-aligned checkpoints into 4-step Turbo models: \textbf{\mageturbo} for text-to-image generation and \textbf{\mageeditturbo} for instruction-based editing. In both tasks, the student is initialized from the corresponding frozen teacher checkpoint. As illustrated in Fig.~\ref{fig:distill_arch}, our distillation framework combines Decoupled DMD~\citep{liu2025decoupleddmd} with adversarial perceptual guidance inspired by SenseFlow~\citep{ge2025senseflow}.

The distillation objective builds on Decoupled DMD, which separates two signals that are coupled in standard distribution-matching distillation: a classifier-free-guidance augmentation (CA) term that steers the student toward the guided teacher direction, and a distribution-matching (DM) regularizer defined through a trainable fake-score network. Instead of sharing a single timestep distribution, Decoupled DMD assigns independent noise schedules to the two terms, improving
stability in the few-step regime.

However, compressing the sampling trajectory to only four denoising steps makes perceptual quality more fragile. As shown in \tableautorefname~\ref{tab:adv_guidance_ablation}, adversarial perceptual guidance provides clear gains for generation and text editing, while its effect on general editing benchmarks is more mixed: it improves ImgEdit slightly, but does not uniformly improve GEdit. We therefore add an adversarial perceptual term based on a feature discriminator operating in frozen DINOv2 and CLIP feature spaces~\citep{oquab2024dinov2,radford2021clip}. This term acts as a lightweight regularizer on top of Decoupled DMD, with the generator updated once every five discriminator updates. Overall, the student is optimized by the total gradient:
\begin{equation}
\begin{gathered}
\nabla\mathcal{L}_{\mathrm{Total}}
=
\underbrace{\Delta_{\mathrm{CA}}+\Delta_{\mathrm{DM}}}_{\nabla\mathcal{L}_{\mathrm{D-DMD}}}
+
\lambda_{\mathrm{GAN}}\,\nabla\mathcal{L}_{\mathrm{GAN}},\\[2pt]
\Delta_{\mathrm{CA}}
=
(w-1)\big(
T_{cond}^{real}(x_{\tau_{\mathrm{ca}}})
-
T_{uncond}^{real}(x_{\tau_{\mathrm{ca}}})
\big),
\qquad
\Delta_{\mathrm{DM}}
=
T_{cond}^{real}(x_{\tau_{\mathrm{dm}}})
-
T_{cond}^{fake}(x_{\tau_{\mathrm{dm}}}).
\end{gathered}
\end{equation}
Here, $\Delta_{\mathrm{CA}}$ is the CFG-augmentation term formed from the teacher's conditional and unconditional predictions $T_{cond}^{real}$ and $T_{uncond}^{real}$, while $\Delta_{\mathrm{DM}}$ is the distribution-matching term against the fake-score prediction $T_{cond}^{fake}$. The two terms are evaluated at independent noise levels $\tau_{\mathrm{ca}}$ and $\tau_{\mathrm{dm}}$. We use guidance scale $w=7.5$ and adversarial weight $\lambda_{\mathrm{GAN}}=0.13$.

\begin{promptbox}
\large
\textbf{\textit{Finding 4:}}
Adversarial perceptual guidance is most beneficial for preserving generation and text-editing quality under a highly compressed four-step trajectory; its gains on general editing benchmarks are benchmark-dependent rather than uniform.
\end{promptbox}

\begin{table}[t]
\centering
\setlength{\tabcolsep}{3.5pt}
\caption{\textbf{Effect of mixing generation data during editing training.}
We compare editing models trained with and without generation data. The effect is modest for the
full-step editor; for the 4-step Turbo editor, generation data improves ImgEdit, while GEdit changes
are mixed.}
\label{tab:mixture_gen_data}
\begin{adjustbox}{width=\textwidth,center}
\begin{tabular}{l | c c c c c c}
\toprule
\textbf{Model} & \textbf{Generation Data} & \textbf{ImgEdit} &
\textbf{GEdit-EN} & \textbf{GEdit-CN} &
\textbf{TextEdit-Syn} & \textbf{TextEdit-Real} \\
\midrule
\mageedit & \checkmark & 4.34 & 8.127 & 8.123 & 14.14 & 16.26  \\
\mageedit & - & 4.34 & 7.991 & 8.062 & 14.29 & 15.95 \\
\mageeditturbo & \checkmark & 4.38 & 8.271 & 8.264 & 12.77 & 15.41  \\
\mageeditturbo & - & 4.20 & 7.984 & 8.050 & 12.66 & 15.20 \\
\bottomrule
\end{tabular}
\end{adjustbox}
\end{table}

\begin{figure*}[t]
\centering
\captionsetup[subfigure]{justification=centering,singlelinecheck=true}

\begin{subfigure}[t]{0.48\textwidth}
    \centering
    \includegraphics[width=0.95\linewidth]{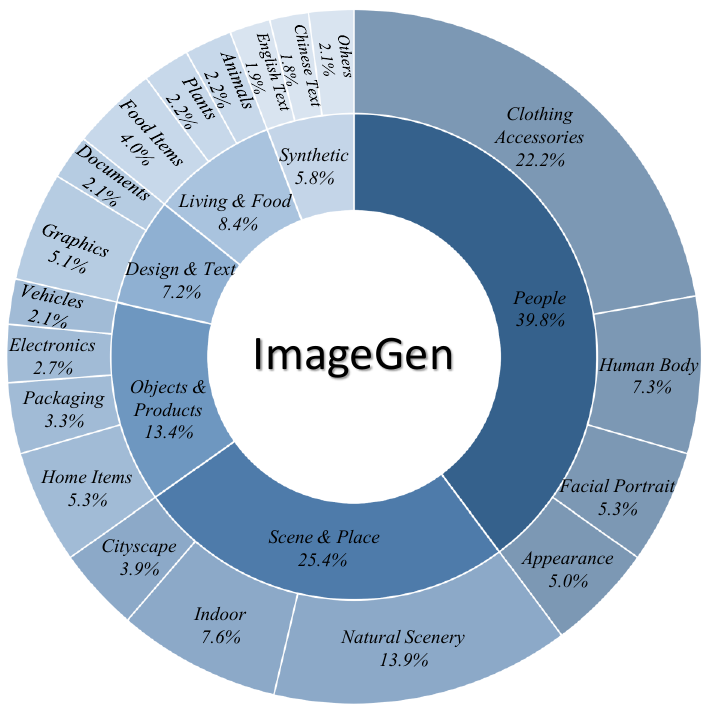}
    \caption{Generation distillation set.}
    \label{fig:distill_data_gen}
\end{subfigure}
\hfill
\begin{subfigure}[t]{0.48\textwidth}
    \centering
    \includegraphics[width=0.95\linewidth]{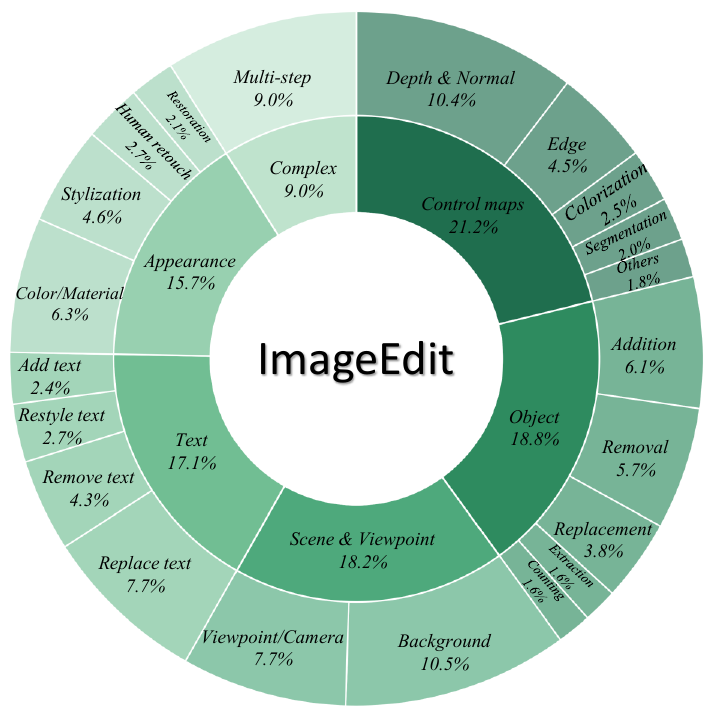}
    \caption{Editing distillation set.}
    \label{fig:distill_data_edit}
\end{subfigure}

\caption{\textbf{Composition of the distillation sets.}
The inner rings show the main categories, and the outer rings show the corresponding sub-categories. (a) The generation distillation set covers diverse content categories such as people, scenes, products, design, food, and synthetic data. (b) The editing distillation set covers various edit-type categories such as object editing, scene and viewpoint changes, text editing, appearance modification, control-map editing, and complex edits.}
\label{fig:distill_data}
\end{figure*}

\paragraph{Generation distillation.}
For \mageturbo, we distill the student from the frozen \mageflow teacher on roughly $200$K curated high-quality prompt--image pairs. The set spans six broad content categories: People, Scene \& Place, Objects \& Products, Living \& Food, Design \& Text, and Synthetic, whose distribution is shown in Fig.~\ref{fig:distill_data}(a). With the combined Decoupled-DMD and adversarial perceptual objective, \mageturbo preserves the visual quality and prompt-following behavior of the multi-step teacher while reducing sampling to a four-step rectified-flow trajectory.

\paragraph{Editing distillation.}
For \mageeditturbo, we use the same distillation framework with two task-specific adjustments. First, the real branch of the feature discriminator uses target images from editing triples rather than images from a separate real-image corpus, aligning the adversarial signal with the edited-image distribution. Second, the student is trained on a $3{:}1$ mixture of editing and generation data. The editing samples teach the shortened model to follow edit instructions, while the retained generation samples preserve open-ended generation ability and improve robustness on edits that require large visual changes. In total, the editing student is distilled on roughly $250$K editing samples spanning six categories: Object, Scene \& Viewpoint, Text, Appearance, Control maps, and Complex. The category distribution before mixing with generation samples is shown in Fig.~\ref{fig:distill_data}(b).

To further verify the role of generation data during editing training, we compare models trained with and without generation data in \tableautorefname~\ref{tab:mixture_gen_data}. The full-step editor changes only marginally. For the Turbo variant, adding generation data improves ImgEdit from $4.20$ to $4.38$, whereas the two GEdit splits remain comparable and do not move uniformly. These results suggest that generation data primarily helps broad category-wise editing robustness under four-step compression, rather than uniformly improving every editing metric.

Overall, this unified recipe allows both \mageturbo and \mageeditturbo to inherit the quality and alignment of their multi-step teachers while reducing inference to only 4 steps.

\begin{promptbox}
\large
\textbf{\textit{Finding 5:}}
Mixing generation data during editing training has its clearest effect on broad editing robustness for the four-step model, as reflected by ImgEdit, while GEdit remains comparable and does not improve uniformly.
\end{promptbox}

\section{Experiments}
\label{sec:experiment}

\subsection{Experimental Setup}
\label{sec:exp_setup}
\paragraph{Text-to-image generation.}
We evaluate \mageflow on eight widely used text-to-image benchmarks covering prompt following, fine-grained generation, and text rendering. GenEval~\citep{ghosh2023geneval}, DPG-Bench~\citep{hu2024ella}, and TIIF-Bench~\citep{wei2025tiif} measure prompt following and compositional understanding. OneIG~\citep{chang2026oneig} evaluates fine-grained generation on English and Chinese splits. CVTG-2K~\citep{du2025textcrafter} focuses on multi-region text rendering, and LongText~\citep{geng2025x} evaluates long-form English and Chinese text rendering. All results are computed at native $1024^2$ resolution using the official evaluation protocols. Unless otherwise specified, \magebase uses $30$ denoising steps, \mageflow uses $20$ steps, and \mageturbo uses $4$ steps.

\begin{table*}[t]
\centering
\caption{\textbf{Summary of text-to-image generation results} across eight benchmarks. Steps is the number of denoising steps at evaluation. Models are grouped into closed-source and open-source. The best open-source score in each column is in \textbf{bold}. The second-best results are \underline{underlined}.}
\label{tab:gen_summary}
\begingroup
\setlength{\tabcolsep}{3.5pt}
\renewcommand{\arraystretch}{1.05}
\scriptsize
\begin{adjustbox}{width=\textwidth,center}
\begin{tabular}{c l c c c c c c c c c c c}
\toprule
Type & \textbf{Model} & \#Params & Steps & GenEval & DPG & TIIF-Short & TIIF-Long & CVTG-2K & OneIG-EN & OneIG-CN & LongText-EN & LongText-CN \\
\midrule
\multirow{5}{*}{\makecell{Closed-\\Source}}
& Seedream 3.0~\cite{gao2025seedream} & -- & -- & 0.84 & 88.27 & 86.02 & 84.31 & 0.592 & 0.530 & 0.528 & 0.896 & 0.878 \\
& Seedream 4.0~\cite{seedream2025seedream} & -- & -- & 0.84 & 88.63 & -- & -- & 0.892 & 0.573 & 0.554 & 0.936 & 0.946 \\
& GPT-Image-1~\cite{gptimage1} & -- & -- & 0.84 & 85.15 & 89.15 & 88.29 & 0.857 & 0.533 & 0.474 & 0.956 & 0.619 \\
& Nano-Banana-Pro~\cite{nanopro} & -- & -- & 0.83 & 87.16 & -- & -- & 0.779 & 0.580 & 0.570 & 0.981 & 0.949 \\
& Kolors 2.0~\cite{kolors} & -- & -- & -- & -- & -- & -- & -- & 0.434 & 0.426 & 0.258 & 0.329 \\
\cmidrule{1-13}
\multirow{7}{*}{\makecell{Open-Source\\Unified}}
& BAGEL~\cite{deng2025bagel} & 14B & 50 & 0.82 & 85.07 & 71.50 & 71.70 & 0.356 & 0.361 & 0.370 & 0.373 & 0.310 \\
& Janus-Pro-7B~\cite{chen2025januspro} & 7B & -- & 0.80 & 84.19 & 66.50 & 65.02 & -- & -- & -- & 0.019 & 0.006 \\
& Emu3-Gen~\cite{wang2024emu3} & 8B & -- & 0.54 & 80.60 & -- & -- & -- & -- & -- & -- & -- \\
& OmniGen2~\cite{wu2025omnigen2} & 4B & 50 & 0.80 & 83.57 & -- & -- & -- & 0.475 & -- & 0.561 & 0.059 \\
& InternVL-U~\cite{tian2026internvl} & 4B & 20 & 0.85 & 85.18 & 74.90 & 73.90 & 0.623 & 0.500 & 0.500 & 0.738 & 0.860 \\
& UniWorld-V1~\cite{lin2025uniworld} & 19B & 30 & 0.80 & 81.38 & -- & -- & -- & -- & -- & -- & -- \\
& Ovis-U1~\cite{wang2025ovis} & 3.6B & 50 & \underline{0.89} & 83.72 & 66.70 & 68.20 & 0.093 & 0.340 & 0.340 & 0.030 & 0.051 \\
\cmidrule{1-13}
\multirow{21}{*}{\makecell{Open-Source\\Specialist}}
& SD3.5-Large~\cite{esser2024sd3} & 8B & 28 & 0.71 & 84.96 & 76.10 & 68.71 & 0.655 & 0.462 & 0.248 & 0.473 & 0.009 \\
& HiDream-I1-Full~\cite{cai2025hidream} & 17B & 50 & 0.83 & 85.89 & 79.33 & 71.88 & 0.740 & 0.477 & 0.337 & 0.543 & 0.024 \\
& FLUX.1-dev~\cite{flux2024} & 12B & 50 & 0.66 & 83.84 & 71.09 & 71.78 & 0.496 & 0.434 & 0.245 & 0.607 & 0.005 \\
& FLUX.1-Krea-dev~\cite{fluxkrea} & 12B & 50 & 0.72 & 86.59 & 80.36 & 81.67 & 0.444 & 0.443 & 0.271 & 0.693 & 0.002 \\
& FLUX.2-dev~\cite{flux-2-2025} & 32B & 50 & 0.87 & 87.57 & \textbf{88.82} & \underline{88.10} & \textbf{0.893} & \textbf{0.551} & 0.516 & \textbf{0.963} & 0.757 \\
& FLUX.2-Klein-Base-4B~\cite{flux-2-2025} & 4B & 50 & 0.78 & 83.02 & 79.94 & 80.01 & 0.656 & 0.485 & 0.366 & 0.554 & 0.071 \\
& FLUX.2-Klein-Base-9B~\cite{flux-2-2025} & 9B & 50 & 0.83 & 85.29 & 81.47 & 84.52 & 0.655 & 0.544 & 0.400 & 0.872 & 0.227 \\
& FLUX.2-Klein-4B~\cite{flux-2-2025} & 4B & 4 & 0.83 & 85.53 & 78.91 & 79.04 & 0.628 & 0.500 & 0.364 & 0.649 & 0.068 \\
& FLUX.2-Klein-9B~\cite{flux-2-2025} & 9B & 4 & 0.86 & 86.20 & 85.22 & 84.13 & 0.424 & 0.538 & 0.406 & 0.872 & 0.226 \\
& Qwen-Image~\cite{qwenimage} & 20B & 50 & 0.87 & \textbf{88.32} & \underline{86.14} & 86.83 & 0.829 & 0.539 & \textbf{0.548} & 0.943 & 0.946 \\
& JoyAI-Image~\cite{song2026joyai} & 16B & 50 & -- & 88.05 & -- & -- & 0.874 & 0.542 & 0.521 & \textbf{0.963} & \textbf{0.963} \\
& HunyuanImage-3.0~\cite{hunyuanimage3} & 80B & 50 & 0.72 & 86.10 & -- & -- & 0.765 & -- & -- & -- & -- \\
& LongCat-Image~\cite{LongCat-Image} & 6B & 50 & 0.87 & 86.80 & 80.93 & 81.30 & 0.866 & 0.516 & 0.518 & 0.885 & \underline{0.956} \\
& Z-Image-Base~\cite{zimage} & 6B & 50 & 0.84 & 88.14 & 80.20 & 83.04 & 0.867 & \underline{0.546} & \underline{0.535} & 0.935 & 0.936 \\
& Z-Image-Turbo~\cite{zimage} & 6B & 8 & 0.82 & 84.86 & 77.73 & 80.05 & 0.859 & 0.528 & 0.507 & 0.917 & 0.926 \\
& Lens-Base~\cite{lens2026} & 3.8B & 50 & 0.70 & 85.51 & 80.33 & 83.49 & 0.635 & 0.527 & 0.500 & 0.830 & 0.741 \\
& Lens-RL~\cite{lens2026} & 3.8B & 20 & 0.85 & \underline{88.19} & 84.23 & 84.92 & 0.843 & 0.526 & 0.497 & 0.901 & 0.817 \\
& Lens-Turbo~\cite{lens2026} & 3.8B & 4 & 0.83 & 87.13 & 82.20 & 81.81 & 0.882 & 0.520 & 0.489 & 0.909 & 0.860 \\
& \cellcolor{red!10}\textbf{\magebase} & \cellcolor{red!10}4B & \cellcolor{red!10}30 & \cellcolor{red!10}0.79 & \cellcolor{red!10}86.26 & \cellcolor{red!10}82.50 & \cellcolor{red!10}83.19 & \cellcolor{red!10}0.851 & \cellcolor{red!10}0.542 & \cellcolor{red!10}0.509 & \cellcolor{red!10}0.904 & \cellcolor{red!10}0.792 \\
& \cellcolor{red!10}\textbf{\mageflow} & \cellcolor{red!10}4B & \cellcolor{red!10}20 & \cellcolor{red!10}\textbf{0.90} & \cellcolor{red!10}86.49 & \cellcolor{red!10}82.19 & \cellcolor{red!10}84.70 & \cellcolor{red!10}\underline{0.887} & \cellcolor{red!10}0.536 & \cellcolor{red!10}0.505 & \cellcolor{red!10}\underline{0.944} & \cellcolor{red!10}0.823 \\
& \cellcolor{red!10}\textbf{\mageturbo} & \cellcolor{red!10}4B & \cellcolor{red!10}4 & \cellcolor{red!10}0.88 & \cellcolor{red!10}85.48 & \cellcolor{red!10}83.58 & \cellcolor{red!10}84.16 & \cellcolor{red!10}0.873 & \cellcolor{red!10}0.523 & \cellcolor{red!10}0.491 & \cellcolor{red!10}0.911 & \cellcolor{red!10}0.801 \\
\bottomrule
\end{tabular}
\end{adjustbox}
\endgroup
\end{table*}

\begin{table}[t]
\centering
\captionsetup{justification=centering}
\caption{Quantitative results on GenEval~\cite{ghosh2023geneval} and DPG-Bench~\cite{hu2024ella}.}
\label{tab:gen_geneval}\label{tab:gen_dpg}
\scriptsize
\setlength{\tabcolsep}{4pt}
\begin{adjustbox}{width=\textwidth,center}
\begin{tabular}{c l c|c c c c c c c|c c c c c c}
\toprule
\multirow{2}{*}{Type} & \multirow{2}{*}{Model} & \multirow{2}{*}{\#Params} & \multicolumn{7}{c|}{\textbf{GenEval}} & \multicolumn{6}{c}{\textbf{DPG-Bench}} \\
\cmidrule(lr){4-10} \cmidrule(lr){11-16}
& & & Single & Two & Counting & Colors & Position & Attr.Bind & Overall & Global & Entity & Attribute & Relation & Other & Overall \\
\midrule
\multirow{4}{*}{\makecell{Closed-\\Source}}
& Seedream 3.0~\cite{gao2025seedream} & -- & 0.99 & 0.96 & 0.91 & 0.93 & 0.47 & 0.80 & 0.84 & 94.31 & 92.65 & 91.36 & 92.78 & 88.24 & 88.27 \\
& Seedream 4.0~\cite{seedream2025seedream} & -- & 0.99 & 0.92 & 0.72 & 0.91 & 0.76 & 0.74 & 0.84 & 87.17 & 92.41 & 92.29 & 93.33 & 95.48 & 88.63 \\
& GPT-Image-1~\cite{gptimage1} & -- & 0.99 & 0.92 & 0.85 & 0.92 & 0.75 & 0.61 & 0.84 & 88.89 & 88.94 & 89.84 & 92.63 & 90.96 & 85.15 \\
& Nano-Banana-Pro~\cite{nanopro} & -- & 1.00 & 0.96 & 0.71 & 0.84 & 0.86 & 0.65 & 0.83 & 91.00 & 92.85 & 91.56 & 92.39 & 89.93 & 87.16 \\
\cmidrule{1-16}
\multirow{8}{*}{\makecell{Open-Source\\Unified}}
& BAGEL~\cite{deng2025bagel} & 14B & \underline{0.99} & 0.94 & 0.81 & 0.88 & 0.64 & 0.63 & 0.82 & 88.94 & 90.37 & 91.29 & 90.82 & 88.67 & 85.07 \\
& Janus-Pro-7B~\cite{chen2025januspro} & 7B & \underline{0.99} & 0.89 & 0.59 & 0.90 & 0.79 & 0.66 & 0.80 & 86.90 & 88.90 & 89.40 & 89.32 & 89.48 & 84.19 \\
& Emu3-Gen~\cite{wang2024emu3} & 8B & 0.98 & 0.71 & 0.34 & 0.81 & 0.17 & 0.21 & 0.54 & 85.21 & 86.68 & 86.84 & 90.22 & 83.15 & 80.60 \\
& Show-o2~\cite{xie2025showo2} & 7B & \textbf{1.00} & 0.87 & 0.58 & \underline{0.92} & 0.52 & 0.62 & 0.76 & -- & -- & -- & -- & -- & -- \\
& OmniGen2~\cite{wu2025omnigen2} & 4B & \textbf{1.00} & 0.95 & 0.64 & 0.88 & 0.55 & 0.76 & 0.80 & 88.81 & 88.83 & 90.18 & 89.37 & 90.27 & 83.57 \\
& InternVL-U~\cite{tian2026internvl} & 4B & \underline{0.99} & 0.94 & 0.74 & 0.91 & 0.77 & 0.74 & 0.85 & 90.39 & 90.78 & 90.68 & 90.29 & 88.77 & 85.18 \\
& UniWorld-V1~\cite{lin2025uniworld} & 19B & \underline{0.99} & 0.93 & 0.79 & 0.89 & 0.49 & 0.70 & 0.80 & 83.64 & 88.39 & 88.44 & 89.27 & 87.22 & 81.38 \\
& Ovis-U1~\cite{wang2025ovis} & 3.6B & 0.98 & \underline{0.98} & \textbf{0.90} & \underline{0.92} & 0.79 & 0.75 & \underline{0.89} & 82.37 & 90.08 & 88.68 & 93.35 & 85.20 & 83.72 \\
\cmidrule{1-16}
\multirow{21}{*}{\makecell{Open-Source\\Specialist}}
& SD3.5-Large~\cite{esser2024sd3} & 8B & 0.98 & 0.89 & 0.73 & 0.83 & 0.34 & 0.47 & 0.71 & 84.75 & 89.93 & 88.19 & 93.23 & 89.70 & 84.96 \\
& HiDream-I1-Full~\cite{cai2025hidream} & 17B & \textbf{1.00} & \underline{0.98} & 0.79 & 0.91 & 0.60 & 0.72 & 0.83 & 76.44 & 90.22 & 89.48 & 93.74 & 91.83 & 85.89 \\
& FLUX.1-dev~\cite{flux2024} & 12B & 0.98 & 0.81 & 0.74 & 0.79 & 0.22 & 0.45 & 0.66 & 74.35 & 90.00 & 88.96 & 90.87 & 88.33 & 83.84 \\
& FLUX.1-Krea-dev~\cite{fluxkrea} & 12B & \underline{0.99} & 0.93 & 0.69 & 0.82 & 0.30 & 0.59 & 0.72 & 87.54 & 92.08 & 89.54 & \underline{94.85} & 87.20 & 86.59 \\
& FLUX.2-dev~\cite{flux-2-2025} & 32B & \textbf{1.00} & \textbf{0.99} & 0.79 & \textbf{0.93} & 0.73 & \textbf{0.78} & 0.87 & 92.20 & 91.36 & \textbf{93.28} & 93.52 & 89.72 & 87.57 \\
& FLUX.2-Klein-Base-4B~\cite{flux-2-2025} & 4B & \underline{0.99} & 0.87 & 0.81 & 0.90 & 0.53 & 0.59 & 0.78 & 91.61 & 88.72 & 90.30 & 91.23 & 88.48 & 83.02 \\
& FLUX.2-Klein-Base-9B~\cite{flux-2-2025} & 9B & \textbf{1.00} & 0.90 & 0.87 & \textbf{0.93} & 0.65 & 0.63 & 0.83 & 89.76 & 90.34 & 90.66 & 93.31 & 86.58 & 85.29 \\
& FLUX.2-Klein-4B~\cite{flux-2-2025} & 4B & \textbf{1.00} & 0.92 & 0.88 & 0.86 & 0.67 & 0.64 & 0.83 & 86.54 & 90.18 & 91.80 & 90.85 & 91.51 & 85.53 \\
& FLUX.2-Klein-9B~\cite{flux-2-2025} & 9B & \underline{0.99} & 0.96 & \underline{0.89} & 0.91 & 0.70 & 0.69 & 0.86 & 88.94 & 91.95 & 89.53 & 92.74 & 92.18 & 86.20 \\
& Qwen-Image~\cite{qwenimage} & 20B & \underline{0.99} & 0.92 & \underline{0.89} & 0.88 & 0.76 & \underline{0.77} & 0.87 & 91.32 & 91.56 & 92.02 & 94.31 & \underline{92.73} & \textbf{88.32} \\
& JoyAI-Image~\cite{song2026joyai} & 16B & -- & -- & -- & -- & -- & -- & -- & -- & -- & -- & -- & -- & 88.05 \\
& HunyuanImage-3.0~\cite{hunyuanimage3} & 80B & \textbf{1.00} & 0.92 & 0.48 & 0.82 & 0.42 & 0.63 & 0.72 & 92.12 & 92.53 & 89.13 & 92.13 & 91.92 & 86.10 \\
& LongCat-Image~\cite{LongCat-Image} & 6B & \underline{0.99} & \underline{0.98} & 0.86 & 0.86 & 0.75 & 0.73 & 0.87 & 89.10 & 92.54 & 92.00 & 93.28 & 87.50 & 86.80 \\
& Z-Image-Base~\cite{zimage} & 6B & \textbf{1.00} & 0.94 & 0.78 & \textbf{0.93} & 0.62 & \underline{0.77} & 0.84 & \textbf{93.39} & 91.22 & \underline{93.16} & 92.22 & 91.52 & 88.14 \\
& Z-Image-Turbo~\cite{zimage} & 6B & \textbf{1.00} & 0.95 & 0.77 & 0.89 & 0.65 & 0.68 & 0.82 & 91.29 & 89.59 & 90.14 & 92.16 & 88.68 & 84.86 \\
& Lens-Base~\cite{lens2026} & 3.8B & \underline{0.99} & 0.79 & 0.58 & 0.83 & 0.52 & 0.46 & 0.70 & 88.68 & 91.24 & 91.93 & 91.89 & 89.65 & 85.51 \\
& Lens-RL~\cite{lens2026} & 3.8B & \textbf{1.00} & 0.95 & 0.83 & 0.89 & 0.73 & 0.72 & 0.85 & 91.07 & \underline{92.94} & 91.93 & 93.39 & \textbf{92.78} & \underline{88.19} \\
& Lens-Turbo~\cite{lens2026} & 3.8B & \underline{0.99} & 0.94 & 0.84 & 0.86 & 0.66 & 0.68 & 0.83 & 92.19 & \textbf{93.44} & 90.34 & \textbf{95.11} & 90.89 & 87.13 \\
& \cellcolor{red!10}\textbf{\magebase} & \cellcolor{red!10}4B & \cellcolor{red!10}\underline{0.99} & \cellcolor{red!10}0.88 & \cellcolor{red!10}0.68 & \cellcolor{red!10}0.90 & \cellcolor{red!10}0.63 & \cellcolor{red!10}0.65 & \cellcolor{red!10}0.79 & \cellcolor{red!10}\underline{92.66} & \cellcolor{red!10}92.38 & \cellcolor{red!10}89.34 & \cellcolor{red!10}88.33 & \cellcolor{red!10}91.87 & \cellcolor{red!10}86.26 \\
& \cellcolor{red!10}\textbf{\mageflow} & \cellcolor{red!10}4B & \cellcolor{red!10}\textbf{1.00} & \cellcolor{red!10}0.97 & \cellcolor{red!10}\underline{0.89} & \cellcolor{red!10}0.89 & \cellcolor{red!10}\textbf{0.93} & \cellcolor{red!10}0.73 & \cellcolor{red!10}\textbf{0.90} & \cellcolor{red!10}91.57 & \cellcolor{red!10}92.41 & \cellcolor{red!10}90.04 & \cellcolor{red!10}91.17 & \cellcolor{red!10}91.70 & \cellcolor{red!10}86.49 \\
& \cellcolor{red!10}\textbf{\mageturbo} & \cellcolor{red!10}4B & \cellcolor{red!10}\textbf{1.00} & \cellcolor{red!10}0.97 & \cellcolor{red!10}0.80 & \cellcolor{red!10}0.88 & \cellcolor{red!10}\underline{0.90} & \cellcolor{red!10}0.73 & \cellcolor{red!10}0.88 & \cellcolor{red!10}80.88 & \cellcolor{red!10}89.42 & \cellcolor{red!10}91.87 & \cellcolor{red!10}91.34 & \cellcolor{red!10}91.19 & \cellcolor{red!10}85.48 \\
\bottomrule
\end{tabular}
\end{adjustbox}
\end{table}

\begin{table*}[t]
\centering
\captionsetup{justification=centering}
\caption{Quantitative Results on TIIF-Bench testmini~\cite{wei2025tiif}.}
\label{tab:tiif}
\renewcommand{\arraystretch}{1.15}
\setlength{\tabcolsep}{3pt}
\scriptsize
\begin{adjustbox}{width=\textwidth,center}
\begin{tabular}{c l|cc|cccccccc|cccccccc|cccccc}
\toprule
\multirow{3}{*}{\textbf{Type}} & \multirow{3}{*}{\textbf{Model}}
  & \multicolumn{2}{c|}{\multirow{2}{*}{\textbf{Overall}}}
  & \multicolumn{8}{c|}{\textbf{Basic Following}}
  & \multicolumn{8}{c|}{\textbf{Advanced Following}}
  & \multicolumn{6}{c}{\textbf{Designer}} \\
\cmidrule(lr){5-12} \cmidrule(lr){13-20} \cmidrule(lr){21-26}
& & & &
  \multicolumn{2}{c}{Avg}
  & \multicolumn{2}{c}{Attribute}
  & \multicolumn{2}{c}{Relation}
  & \multicolumn{2}{c|}{Reasoning}
  & \multicolumn{2}{c}{Avg}
  & \multicolumn{2}{c}{\makecell{Attribute\\+Relation}}
  & \multicolumn{2}{c}{\makecell{Attribute\\+Reasoning}}
  & \multicolumn{2}{c|}{\makecell{Relation\\+Reasoning}}
  & \multicolumn{2}{c}{Style}
  & \multicolumn{2}{c}{Text}
  & \multicolumn{2}{c}{\makecell{Real\\World}} \\
& & short & long & short & long & short & long & short & long & short & long & short & long & short & long & short & long & short & long & short & long & short & long & short & long \\
\midrule
\multirow{4}{*}{\makecell{Closed-\\Source}}
& Seedream 3.0~\cite{gao2025seedream} & 86.02 & 84.31 & 87.07 & 84.93 & 90.50 & 90.00 & 89.85 & 85.94 & 80.86 & 78.86 & 79.16 & 80.60 & 79.76 & 81.82 & 77.23 & 78.85 & 75.64 & 78.64 & 100.00 & 93.33 & 97.17 & 87.78 & 83.21 & 83.58 \\
& GPT-Image-1~\cite{gptimage1} & 89.15 & 88.29 & 90.75 & 89.66 & 91.33 & 87.08 & 84.57 & 84.57 & 96.32 & 97.32 & 88.55 & 88.35 & 87.07 & 89.44 & 87.22 & 83.96 & 85.59 & 83.21 & 90.00 & 93.33 & 89.83 & 86.83 & 89.73 & 93.46 \\
& DALL-E 3~\cite{dalle3} & 74.96 & 70.81 & 78.72 & 78.50 & 79.50 & 79.83 & 80.82 & 78.82 & 75.82 & 76.82 & 73.39 & 67.27 & 73.45 & 67.20 & 72.01 & 71.34 & 63.59 & 60.72 & 89.66 & 86.67 & 66.83 & 54.83 & 72.93 & 60.99 \\
& MidJourney v7~\cite{midjourney} & 68.74 & 65.69 & 77.41 & 76.00 & 77.58 & 81.83 & 82.07 & 76.82 & 72.57 & 69.32 & 64.66 & 60.53 & 67.20 & 62.70 & 81.22 & 71.59 & 60.72 & 64.59 & 83.33 & 80.00 & 24.83 & 20.83 & 68.83 & 63.61 \\
\cmidrule{1-26}
\multirow{4}{*}{\makecell{Open-Source\\Unified}}
& BAGEL~\cite{deng2025bagel} & 71.50 & 71.70 & 81.80 & 80.10 & 82.50 & 83.50 & 83.00 & 79.90 & 79.90 & 76.80 & 70.20 & 72.20 & 74.40 & 75.00 & 67.40 & 70.10 & 72.00 & 74.90 & 86.70 & 83.30 & 29.40 & 33.90 & 68.30 & 67.90 \\
& Janus-Pro-7B~\cite{chen2025januspro} & 66.50 & 65.02 & 79.33 & 78.25 & 79.33 & 82.33 & 78.32 & 73.32 & 80.32 & 79.07 & 59.71 & 58.82 & 66.07 & 56.20 & 70.46 & 70.84 & 67.22 & 59.97 & 60.00 & 70.00 & 28.83 & 33.83 & 65.84 & 60.25 \\
& InternVL-U~\cite{tian2026internvl} & 74.90 & 73.90 & 82.30 & 81.50 & 86.00 & 81.50 & 84.10 & 82.20 & 76.70 & 80.90 & 73.50 & 72.70 & 75.30 & 76.20 & 70.40 & 67.60 & 75.50 & 75.80 & 93.30 & 83.30 & 47.50 & 50.70 & 65.30 & 66.80 \\
& Ovis-U1~\cite{wang2025ovis} & 66.70 & 68.20 & 77.80 & 79.40 & 83.50 & 81.50 & 80.10 & 81.40 & 69.90 & 75.20 & 67.40 & 67.80 & 71.80 & 68.30 & 66.80 & 73.80 & 69.00 & 65.90 & 83.30 & 86.70 & 8.10 & 12.70 & 67.20 & 68.70 \\
\cmidrule{1-26}
\multirow{15}{*}{\makecell{Open-Source\\Specialist}}
& FLUX.1-dev~\cite{flux2024} & 71.09 & 71.78 & 83.12 & 78.65 & 87.05 & 83.17 & 87.25 & 80.39 & 75.01 & 72.39 & 65.79 & 68.54 & 67.07 & 73.69 & 73.84 & 73.34 & 69.09 & 71.59 & 66.67 & 66.67 & 43.83 & 52.83 & 70.72 & 71.47 \\
& FLUX.1-Krea-dev~\cite{fluxkrea} & 80.36 & 81.67 & 84.26 & 83.76 & 86.00 & 84.00 & 82.93 & 86.18 & \underline{83.47} & 80.99 & 73.85 & 76.15 & 79.89 & 77.01 & 74.75 & 79.21 & 76.43 & 78.98 & 73.33 & 80.00 & 66.52 & 70.14 & 93.66 & \underline{94.78} \\
& FLUX.2-Klein-Base-4B~\cite{flux-2-2025} & 79.94 & 80.01 & 82.99 & 80.21 & 89.33 & 84.83 & 83.74 & 85.37 & 74.38 & 69.42 & 73.60 & 76.28 & 77.01 & 78.16 & 68.32 & 77.23 & 67.52 & 71.97 & \underline{93.33} & 80.00 & 77.38 & 76.47 & \underline{94.03} & 90.67 \\
& FLUX.2-Klein-Base-9B~\cite{flux-2-2025} & 81.47 & 84.52 & 82.49 & 85.28 & 84.00 & 85.33 & 83.74 & 87.80 & 79.34 & 82.64 & 78.32 & \underline{82.02} & 78.16 & 77.59 & 75.74 & 78.53 & 73.89 & \underline{83.44} & 76.67 & 80.00 & 84.16 & 87.78 & 89.18 & 90.67 \\
& FLUX.2-Klein-4B~\cite{flux-2-2025} & 78.91 & 79.04 & 81.22 & 80.96 & 79.33 & 80.00 & 83.74 & 83.74 & 80.99 & 79.34 & 73.34 & 73.56 & 77.01 & 79.31 & 72.28 & 74.11 & 67.52 & 75.16 & 76.67 & 73.33 & 75.11 & 67.42 & 91.79 & 92.16 \\
& FLUX.2-Klein-9B~\cite{flux-2-2025} & \underline{85.22} & 84.13 & \underline{88.32} & 85.17 & 90.67 & 84.00 & 86.99 & 90.24 & \textbf{86.78} & 81.36 & \textbf{80.87} & 81.12 & 78.24 & 77.01 & 74.75 & 78.71 & 76.43 & 81.53 & 93.10 & 90.00 & 90.05 & 85.07 & 93.28 & 91.42 \\
& Qwen-Image~\cite{qwenimage} & \textbf{86.14} & \textbf{86.83} & 86.18 & 87.22 & 90.50 & \underline{91.50} & 88.22 & \textbf{90.78} & 79.81 & 79.38 & 79.30 & 80.88 & 79.21 & 78.94 & \underline{78.85} & 81.69 & 75.57 & 78.59 & \textbf{100.00} & \textbf{100.00} & \underline{92.76} & \underline{89.14} & 90.30 & 91.42 \\
& Z-Image-Base~\cite{zimage} & 80.20 & 83.04 & 78.36 & 82.79 & 79.50 & 86.50 & 80.45 & 79.94 & 75.13 & 81.94 & 72.89 & 77.02 & 72.91 & 77.56 & 66.99 & 73.82 & 73.89 & 75.62 & 90.00 & \underline{93.33} & \textbf{94.84} & \textbf{93.21} & 88.06 & 85.45 \\
& Z-Image-Turbo~\cite{zimage} & 77.73 & 80.05 & 81.85 & 81.59 & 86.50 & 87.00 & 82.88 & 79.99 & 76.17 & 77.77 & 68.32 & 74.69 & 72.04 & 75.24 & 60.22 & 73.33 & 68.90 & 71.92 & 83.33 & \underline{93.33} & 83.71 & 84.62 & 85.82 & 77.24 \\
& Lens-Base~\cite{lens2026} & 80.33 & 83.49 & 82.90 & 83.70 & 86.50 & 83.00 & 82.37 & 85.20 & 79.81 & 8.98 & 74.35 & 80.22 & 71.24 & 80.18 & 75.87 & 82.70 & 74.40 & 76.96 & 90.00 & \underline{93.33} & 71.04 & 73.76 & 91.79 & 93.28 \\
& Lens-RL~\cite{lens2026} & 84.23 & \underline{84.92} & 88.03 & \textbf{89.21} & \underline{91.00} & 89.50 & \textbf{91.18} & \underline{90.54} & 81.90 & \textbf{87.58} & \underline{80.75} & \textbf{83.27} & \textbf{84.53} & \underline{82.82} & \textbf{79.30} & \textbf{85.01} & \underline{78.97} & \textbf{83.90} & 66.67 & 66.67 & 90.50 & 84.62 & \underline{94.03} & 93.66 \\
& Lens-Turbo~\cite{lens2026} & 82.20 & 81.81 & \textbf{88.48} & 88.41 & \textbf{92.00} & 91.00 & \underline{90.49} & 88.17 & 82.94 & \underline{86.06} & 78.49 & 79.62 & 81.35 & \textbf{83.34} & 74.45 & 76.72 & \textbf{81.24} & 81.49 & 56.67 & 56.67 & 87.78 & 81.00 & 92.91 & 91.79 \\
& \cellcolor{red!10}\textbf{\magebase} & \cellcolor{red!10}82.50 & \cellcolor{red!10}83.19 & \cellcolor{red!10}83.36 & \cellcolor{red!10}85.34 & \cellcolor{red!10}87.00 & \cellcolor{red!10}90.50 & \cellcolor{red!10}88.86 & \cellcolor{red!10}85.63 & \cellcolor{red!10}74.21 & \cellcolor{red!10}79.90 & \cellcolor{red!10}77.04 & \cellcolor{red!10}80.81 & \cellcolor{red!10}78.62 & \cellcolor{red!10}80.10 & \cellcolor{red!10}77.48 & \cellcolor{red!10}\underline{82.98} & \cellcolor{red!10}72.86 & \cellcolor{red!10}79.74 & \cellcolor{red!10}86.67 & \cellcolor{red!10}83.33 & \cellcolor{red!10}84.62 & \cellcolor{red!10}75.11 & \cellcolor{red!10}92.16 & \cellcolor{red!10}91.42 \\
& \cellcolor{red!10}\textbf{\mageflow} & \cellcolor{red!10}82.19 & \cellcolor{red!10}84.70 & \cellcolor{red!10}83.44 & \cellcolor{red!10}85.72 & \cellcolor{red!10}88.00 & \cellcolor{red!10}86.00 & \cellcolor{red!10}86.19 & \cellcolor{red!10}86.67 & \cellcolor{red!10}76.12 & \cellcolor{red!10}84.50 & \cellcolor{red!10}77.35 & \cellcolor{red!10}80.86 & \cellcolor{red!10}79.61 & \cellcolor{red!10}79.51 & \cellcolor{red!10}76.72 & \cellcolor{red!10}81.97 & \cellcolor{red!10}74.66 & \cellcolor{red!10}80.28 & \cellcolor{red!10}80.00 & \cellcolor{red!10}80.00 & \cellcolor{red!10}83.26 & \cellcolor{red!10}88.24 & \cellcolor{red!10}\textbf{95.15} & \cellcolor{red!10}\textbf{95.15} \\
& \cellcolor{red!10}\textbf{\mageturbo} & \cellcolor{red!10}83.58 & \cellcolor{red!10}84.16 & \cellcolor{red!10}85.11 & \cellcolor{red!10}\underline{88.45} & \cellcolor{red!10}88.00 & \cellcolor{red!10}\textbf{92.00} & \cellcolor{red!10}87.52 & \cellcolor{red!10}88.86 & \cellcolor{red!10}79.81 & \cellcolor{red!10}84.50 & \cellcolor{red!10}77.07 & \cellcolor{red!10}78.70 & \cellcolor{red!10}\underline{82.32} & \cellcolor{red!10}80.90 & \cellcolor{red!10}74.28 & \cellcolor{red!10}77.39 & \cellcolor{red!10}71.86 & \cellcolor{red!10}76.61 & \cellcolor{red!10}86.67 & \cellcolor{red!10}80.00 & \cellcolor{red!10}89.59 & \cellcolor{red!10}86.88 & \cellcolor{red!10}92.16 & \cellcolor{red!10}90.30 \\
\bottomrule
\end{tabular}
\end{adjustbox}
\end{table*}

\begin{table}[t]
\centering
\captionsetup{justification=centering}
\caption{Quantitative results of multi-region English text rendering on CVTG-2K~\cite{du2025textcrafter}.}
\label{tab:gen_cvtg}
\scriptsize
\setlength{\tabcolsep}{4pt}
\begin{tabular}{c l c c c c c c c c}
\toprule
Type & Model & \#Params & 2-reg. & 3-reg. & 4-reg. & 5-reg. & Avg & NED & CLIP \\
\midrule
\multirow{4}{*}{\makecell{Closed-\\Source}}
& Seedream 3.0~\cite{gao2025seedream} & -- & 0.628 & 0.596 & 0.604 & 0.561 & 0.592 & 0.854 & 0.782 \\
& Seedream 4.0~\cite{seedream2025seedream} & -- & 0.890 & 0.915 & 0.899 & 0.887 & 0.892 & 0.951 & 0.785 \\
& GPT-Image-1~\cite{gptimage1} & -- & 0.878 & 0.866 & 0.873 & 0.822 & 0.857 & 0.948 & 0.798 \\
& Nano-Banana-Pro~\cite{nanopro} & -- & 0.737 & 0.775 & 0.786 & 0.793 & 0.779 & 0.875 & 0.737 \\
\cmidrule{1-10}
\multirow{3}{*}{\makecell{Open-Source\\Unified}}
& BAGEL~\cite{deng2025bagel} & 14B & 0.498 & 0.391 & 0.332 & 0.291 & 0.356 & 0.657 & 0.779 \\
& InternVL-U~\cite{tian2026internvl} & 4B & 0.729 & 0.660 & 0.618 & 0.549 & 0.623 & 0.804 & 0.816 \\
& Ovis-U1~\cite{wang2025ovis} & 3.6B & 0.133 & 0.109 & 0.091 & 0.065 & 0.093 & 0.477 & 0.725 \\
\cmidrule{1-10}
\multirow{20}{*}{\makecell{Open-Source\\Specialist}}
& SD3.5-Large~\cite{esser2024sd3} & 8B & 0.729 & 0.682 & 0.657 & 0.594 & 0.655 & 0.847 & 0.780 \\
& FLUX.1-dev~\cite{flux2024} & 12B & 0.609 & 0.553 & 0.466 & 0.432 & 0.496 & 0.688 & 0.740 \\
& FLUX.1-Krea-dev~\cite{fluxkrea} & 12B & 0.527 & 0.454 & 0.440 & 0.407 & 0.444 & 0.737 & 0.773 \\
& FLUX.2-dev~\cite{flux-2-2025} & 32B & \textbf{0.926} & 0.890 & \underline{0.899} & \textbf{0.873} & \textbf{0.893} & 0.948 & 0.810 \\
& FLUX.2-Klein-Base-4B~\cite{flux-2-2025} & 4B & 0.679 & 0.653 & 0.667 & 0.636 & 0.656 & 0.825 & 0.757 \\
& FLUX.2-Klein-Base-9B~\cite{flux-2-2025} & 9B & 0.676 & 0.653 & 0.667 & 0.636 & 0.655 & 0.825 & 0.757 \\
& FLUX.2-Klein-4B~\cite{flux-2-2025} & 4B & 0.637 & 0.641 & 0.639 & 0.601 & 0.628 & 0.829 & 0.773 \\
& FLUX.2-Klein-9B~\cite{flux-2-2025} & 9B & 0.464 & 0.443 & 0.416 & 0.399 & 0.424 & 0.733 & 0.759 \\
& Qwen-Image~\cite{qwenimage} & 20B & 0.837 & 0.836 & 0.831 & 0.816 & 0.829 & 0.912 & 0.802 \\
& JoyAI-Image~\cite{song2026joyai} & 16B & -- & -- & -- & -- & 0.874 & 0.937 & 0.799 \\
& HunyuanImage-3.0~\cite{hunyuanimage3} & 80B & 0.830 & 0.763 & 0.738 & 0.728 & 0.765 & 0.876 & 0.812 \\
& LongCat-Image~\cite{LongCat-Image} & 6B & \underline{0.913} & 0.874 & 0.856 & 0.831 & 0.866 & 0.936 & 0.786 \\
& Z-Image-Base~\cite{zimage} & 6B & 0.901 & 0.872 & 0.865 & 0.851 & 0.867 & 0.937 & 0.797 \\
& Z-Image-Turbo~\cite{zimage} & 6B & 0.887 & 0.866 & 0.863 & 0.835 & 0.859 & 0.928 & 0.805 \\
& Lens-Base~\cite{lens2026} & 3.8B & 0.708 & 0.656 & 0.645 & 0.577 & 0.635 & 0.814 & 0.777 \\
& Lens-RL~\cite{lens2026} & 3.8B & 0.871 & 0.867 & 0.847 & 0.806 & 0.843 & 0.935 & 0.797 \\
& Lens-Turbo~\cite{lens2026} & 3.8B & 0.911 & \underline{0.899} & 0.883 & \underline{0.853} & 0.882 & \textbf{0.955} & 0.801 \\
& \cellcolor{red!10}\textbf{\magebase} & \cellcolor{red!10}4B & \cellcolor{red!10}0.908 & \cellcolor{red!10}0.878 & \cellcolor{red!10}0.866 & \cellcolor{red!10}0.789 & \cellcolor{red!10}0.851 & \cellcolor{red!10}0.933 & \cellcolor{red!10}0.815 \\
& \cellcolor{red!10}\textbf{\mageflow} & \cellcolor{red!10}4B & \cellcolor{red!10}0.902 & \cellcolor{red!10}\textbf{0.902} & \cellcolor{red!10}\textbf{0.902} & \cellcolor{red!10}0.850 & \cellcolor{red!10}\underline{0.887} & \cellcolor{red!10}\underline{0.950} & \cellcolor{red!10}\underline{0.826} \\
& \cellcolor{red!10}\textbf{\mageturbo} & \cellcolor{red!10}4B & \cellcolor{red!10}0.892 & \cellcolor{red!10}0.883 & \cellcolor{red!10}0.878 & \cellcolor{red!10}0.851 & \cellcolor{red!10}0.873 & \cellcolor{red!10}0.945 & \cellcolor{red!10}\textbf{0.831} \\
\bottomrule
\end{tabular}
\end{table}

\begin{table*}[t]
\centering
\captionsetup{justification=centering}
\caption{Quantitative results on OneIG-EN and OneIG-CN \cite{chang2026oneig}.}
\label{tab:gen_oneig}
\scriptsize
\setlength{\tabcolsep}{4pt}
\begin{adjustbox}{width=\textwidth,center}
\begin{tabular}{c l c|c c c c c c|c c c c c c}
\toprule
\multirow{2}{*}{Type} & \multirow{2}{*}{Model} & \multirow{2}{*}{\#Params} & \multicolumn{6}{c|}{\textbf{OneIG-EN}} & \multicolumn{6}{c}{\textbf{OneIG-CN}} \\
\cmidrule(lr){4-9} \cmidrule(lr){10-15}
& & & Alignment & Text & Reasoning & Style & Diversity & Overall & Alignment & Text & Reasoning & Style & Diversity & Overall \\
\midrule
\multirow{5}{*}{\makecell{Closed-\\Source}}
& Seedream 3.0~\cite{gao2025seedream} & -- & 0.818 & 0.865 & 0.275 & 0.413 & 0.277 & 0.530 & 0.793 & 0.928 & 0.281 & 0.397 & 0.243 & 0.528 \\
& Seedream 4.0~\cite{seedream2025seedream} & -- & 0.892 & 0.983 & 0.347 & 0.453 & 0.191 & 0.573 & 0.836 & 0.986 & 0.304 & 0.443 & 0.200 & 0.554 \\
& GPT-Image-1~\cite{gptimage1} & -- & 0.851 & 0.857 & 0.345 & 0.462 & 0.151 & 0.533 & 0.812 & 0.650 & 0.300 & 0.449 & 0.159 & 0.474 \\
& Nano-Banana-Pro~\cite{nanopro} & -- & 0.890 & 0.940 & 0.330 & 0.480 & 0.250 & 0.580 & 0.840 & 0.980 & 0.310 & 0.460 & 0.240 & 0.570 \\
& Kolors 2.0~\cite{kolors} & -- & 0.820 & 0.427 & 0.262 & 0.360 & 0.300 & 0.434 & 0.738 & 0.502 & 0.226 & 0.331 & 0.333 & 0.426 \\
\cmidrule{1-15}
\multirow{5}{*}{\makecell{Open-Source\\Unified}}
& BAGEL~\cite{deng2025bagel} & 14B & 0.769 & 0.244 & 0.173 & 0.367 & \underline{0.251} & 0.361 & 0.672 & 0.365 & 0.186 & 0.357 & 0.268 & 0.370 \\
& Show-o2~\cite{xie2025showo2} & 7B & 0.798 & 0.002 & 0.219 & 0.317 & 0.186 & 0.304 & -- & -- & -- & -- & -- & -- \\
& OmniGen2~\cite{wu2025omnigen2} & 4B & 0.804 & 0.680 & 0.271 & 0.377 & 0.242 & 0.475 & -- & -- & -- & -- & -- & -- \\
& InternVL-U~\cite{tian2026internvl} & 4B & 0.820 & 0.740 & 0.270 & 0.400 & 0.250 & 0.500 & 0.750 & 0.900 & 0.230 & 0.370 & 0.260 & 0.500 \\
& Ovis-U1~\cite{wang2025ovis} & 3.6B & 0.810 & 0.030 & 0.220 & \textbf{0.450} & 0.180 & 0.340 & 0.720 & 0.150 & 0.210 & \textbf{0.430} & 0.200 & 0.340 \\
\cmidrule{1-15}
\multirow{19}{*}{\makecell{Open-Source\\Specialist}}
& SD3.5-Large~\cite{esser2024sd3} & 8B & 0.809 & 0.629 & 0.294 & 0.353 & 0.225 & 0.462 & 0.137 & 0.200 & 0.141 & 0.243 & \underline{0.519} & 0.248 \\
& HiDream-I1-Full~\cite{cai2025hidream} & 17B & 0.829 & 0.707 & \textbf{0.317} & 0.347 & 0.186 & 0.477 & 0.620 & 0.205 & 0.256 & 0.304 & 0.300 & 0.337 \\
& FLUX.1-dev~\cite{flux2024} & 12B & 0.786 & 0.523 & 0.253 & 0.368 & 0.238 & 0.434 & 0.152 & 0.218 & 0.116 & 0.282 & 0.456 & 0.245 \\
& FLUX.1-Krea-dev~\cite{fluxkrea} & 12B & 0.842 & 0.483 & 0.282 & 0.377 & 0.232 & 0.443 & 0.105 & 0.234 & 0.114 & 0.263 & \textbf{0.640} & 0.271 \\
& FLUX.2-Klein-Base-4B~\cite{flux-2-2025} & 4B & 0.830 & 0.688 & 0.262 & 0.414 & 0.230 & 0.485 & 0.744 & 0.211 & 0.240 & 0.384 & 0.250 & 0.366 \\
& FLUX.2-Klein-Base-9B~\cite{flux-2-2025} & 9B & 0.867 & 0.897 & 0.299 & 0.442 & 0.217 & \underline{0.544} & 0.795 & 0.310 & 0.250 & 0.417 & 0.229 & 0.400 \\
& FLUX.2-Klein-4B~\cite{flux-2-2025} & 4B & 0.862 & 0.779 & 0.272 & 0.420 & 0.169 & 0.500 & 0.794 & 0.208 & 0.247 & 0.386 & 0.183 & 0.364 \\
& FLUX.2-Klein-9B~\cite{flux-2-2025} & 9B & \textbf{0.884} & 0.898 & \underline{0.306} & \underline{0.445} & 0.158 & 0.538 & 0.818 & 0.354 & \textbf{0.280} & \underline{0.418} & 0.162 & 0.406 \\
& Qwen-Image~\cite{qwenimage} & 20B & \underline{0.882} & 0.891 & \underline{0.306} & 0.418 & 0.197 & 0.539 & \textbf{0.825} & 0.963 & 0.267 & 0.405 & 0.279 & \textbf{0.548} \\
& JoyAI-Image~\cite{song2026joyai} & 16B & -- & -- & -- & -- & -- & 0.542 & -- & -- & -- & -- & -- & 0.521 \\
& LongCat-Image~\cite{LongCat-Image} & 6B & 0.854 & 0.923 & 0.216 & 0.365 & 0.222 & 0.516 & 0.812 & 0.948 & 0.253 & 0.355 & 0.224 & 0.518 \\
& Z-Image-Base~\cite{zimage} & 6B & 0.881 & \underline{0.987} & 0.280 & 0.387 & 0.194 & \textbf{0.546} & 0.793 & \textbf{0.988} & 0.266 & 0.386 & 0.243 & \underline{0.535} \\
& Z-Image-Turbo~\cite{zimage} & 6B & 0.840 & \textbf{0.994} & 0.298 & 0.368 & 0.139 & 0.528 & 0.782 & \underline{0.982} & \underline{0.276} & 0.361 & 0.134 & 0.507 \\
& Lens-Base~\cite{lens2026} & 3.8B & 0.842 & 0.908 & 0.241 & 0.384 & \textbf{0.262} & 0.527 & 0.781 & 0.904 & 0.256 & 0.358 & 0.200 & 0.500 \\
& Lens-RL~\cite{lens2026} & 3.8B & 0.877 & 0.965 & 0.279 & 0.348 & 0.162 & 0.526 & \underline{0.824} & 0.944 & 0.259 & 0.316 & 0.141 & 0.497 \\
& Lens-Turbo~\cite{lens2026} & 3.8B & 0.868 & 0.968 & 0.291 & 0.314 & 0.160 & 0.520 & 0.802 & 0.952 & 0.260 & 0.284 & 0.148 & 0.489 \\
& \cellcolor{red!10}\textbf{\magebase} & \cellcolor{red!10}4B & \cellcolor{red!10}0.863 & \cellcolor{red!10}0.950 & \cellcolor{red!10}0.305 & \cellcolor{red!10}0.431 & \cellcolor{red!10}0.159 & \cellcolor{red!10}0.542 & \cellcolor{red!10}0.800 & \cellcolor{red!10}0.904 & \cellcolor{red!10}0.279 & \cellcolor{red!10}0.396 & \cellcolor{red!10}0.163 & \cellcolor{red!10}0.509 \\
& \cellcolor{red!10}\textbf{\mageflow} & \cellcolor{red!10}4B & \cellcolor{red!10}0.869 & \cellcolor{red!10}0.969 & \cellcolor{red!10}\underline{0.309} & \cellcolor{red!10}0.411 & \cellcolor{red!10}0.124 & \cellcolor{red!10}0.536 & \cellcolor{red!10}0.812 & \cellcolor{red!10}0.926 & \cellcolor{red!10}\textbf{0.282} & \cellcolor{red!10}0.385 & \cellcolor{red!10}0.119 & \cellcolor{red!10}0.505 \\
& \cellcolor{red!10}\textbf{\mageturbo} & \cellcolor{red!10}4B & \cellcolor{red!10}0.861 & \cellcolor{red!10}0.949 & \cellcolor{red!10}0.307 & \cellcolor{red!10}0.391 & \cellcolor{red!10}0.105 & \cellcolor{red!10}0.523 & \cellcolor{red!10}0.803 & \cellcolor{red!10}0.909 & \cellcolor{red!10}\underline{0.281} & \cellcolor{red!10}0.363 & \cellcolor{red!10}0.099 & \cellcolor{red!10}0.491 \\
\bottomrule
\end{tabular}
\end{adjustbox}
\end{table*}

\begin{table*}[t]
\centering
\caption{\textbf{Summary of image editing results} across three benchmarks.
Steps denotes the number of denoising steps used during inference. Models are grouped into closed-source and open-source. The best open-source score in each column is in \textbf{bold}. The second-best results are \underline{underlined}.}
\label{tab:edit_summary}
\begingroup
\setlength{\tabcolsep}{4pt}
\renewcommand{\arraystretch}{1.05}
\footnotesize
\begin{adjustbox}{width=\textwidth,center}
\begin{tabular}{c l|cc|ccccc}
\toprule
Type & \textbf{Model} & \#Params & Steps & ImgEdit & GEdit-EN & GEdit-CN & TextEdit-Syn & TextEdit-Real \\
\midrule
\multirow{4}{*}{\makecell{Closed-\\Source}}
& Nano-Banana~\cite{nanopro} & -- & -- & 4.29 & 7.291 & 7.399 & 16.54 & 18.22 \\
& Seedream4.0~\cite{seedream2025seedream} & -- & -- & 4.30 & 7.701 & 7.692 & 14.90 & 18.54 \\
& Seedream4.5~\cite{seedream2025seedream} & -- & -- & 4.32 & 7.820 & 7.800 & -- & -- \\
& Nano-Banana-Pro~\cite{nanopro} & -- & -- & 4.37 & 7.738 & 7.799 & -- & -- \\
\cmidrule{1-9}
\multirow{4}{*}{\makecell{Open-Source\\Unified}}
& BAGEL~\cite{deng2025bagel} & 14B & 50 & 3.20 & -- & -- & 10.41 & 12.83 \\
& OmniGen2~\cite{wu2025omnigen2} & 4B & 50 & 3.44 & -- & -- & 8.52 & 10.99 \\
& UniWorld-V1~\cite{lin2025uniworld} & 19B & 30 & 3.26 & -- & -- & -- & -- \\
& Emu3.5~\cite{cui2025emu35nativemultimodalmodels} & 34B & -- & 4.41 & -- & -- & 9.34 & 14.05 \\
\cmidrule{1-9}
\multirow{23}{*}{\makecell{Open-Source\\Specialist}}
& Instruct-Pix2Pix~\cite{brooks2023instructpix2pix} & 1B & 10 & 1.88 & -- & -- & 5.51 & 5.35 \\
& MagicBrush~\cite{zhang2023magicbrush} & 1B & 100 & 1.90 & -- & -- & 3.86 & 4.12 \\
& AnyEdit~\cite{yu2025anyedit} & 1B & 50 & 2.45 & -- & -- & -- & -- \\
& UltraEdit~\cite{zhao2024ultraedit} & 2B & 50 & 2.70 & -- & -- & -- & -- \\
& OmniGen~\cite{xiao2025omnigen} & 3.8B & 50 & 2.96 & -- & -- & -- & -- \\
& ICEdit~\cite{zhang2025icedit} & 12B & 28 & 3.05 & -- & -- & -- & -- \\
& Step1X-Edit-v1.2~\cite{liu2025step1x} & 19B & 50 & 3.95 & 7.480 & 7.467 & 9.26 & 12.02 \\
& ChronoEdit~\cite{wu2025chronoedit} & 14B & 50 & 4.42 & -- & -- & -- & -- \\
& FLUX.1-Kontext-dev~\cite{labs2025flux} & 12B & 28 & 3.71 & 6.462 & 1.857 & 12.14 & 14.31 \\
& FLUX.2-dev~\cite{flux-2-2025} & 32B & 50 & 4.35 & 7.413 & 7.278 & 11.86 & 14.71 \\
& FLUX.2-Klein-Base-4B~\cite{flux-2-2025} & 4B & 50 & 3.80 & 7.081 & 7.102 & 11.01 & 13.79 \\
& FLUX.2-Klein-4B~\cite{flux-2-2025} & 4B & 4 & 4.01 & 7.717 & 7.750 & 11.84 & 14.46 \\
& FLUX.2-Klein-Base-9B~\cite{flux-2-2025} & 9B & 50 & 4.05 & 7.740 & 7.745 & 12.76 & 15.65 \\
& FLUX.2-Klein-9B~\cite{flux-2-2025} & 9B & 4 & 4.18 & 8.040 & 8.055 & 12.73 & 15.75 \\
& Z-Image-Edit~\cite{zimage} & 6B & 50 & 4.30 & 7.570 & 7.540 & -- & -- \\
& Qwen-Image-Edit-2509~\cite{qwenimage} & 20B & 50 & 4.31 & 7.480 & 7.467 & 13.40 & 15.81 \\
& Qwen-Image-Edit-2511~\cite{qwenimage} & 20B & 50 & \underline{4.51} & 7.877 & 7.819 & 13.53 & 16.81 \\
& LongCat-Image-Edit~\cite{LongCat-Image} & 6B & 50 & 4.45 & 7.748 & 7.731 & 12.46 & 14.89 \\
& FireRed-Image-Edit-1.0~\cite{firered2025} & 20B & 50 & \textbf{4.56} & 7.943 & 7.887 & \textbf{15.19} & \textbf{17.23} \\
& JoyAI-Image-Edit~\cite{song2026joyai} & 16B & 50 & 4.46 & \textbf{8.276} & \underline{8.125} & \underline{14.80} & \underline{17.23} \\
& \cellcolor{red!10}\mageeditbase & \cellcolor{red!10}4B & \cellcolor{red!10}30 & \cellcolor{red!10}4.28 & \cellcolor{red!10}7.860 & \cellcolor{red!10}7.970 & \cellcolor{red!10}13.63 & \cellcolor{red!10}15.57 \\
& \cellcolor{red!10}\mageedit & \cellcolor{red!10}4B & \cellcolor{red!10}30 & \cellcolor{red!10}4.34 & \cellcolor{red!10}8.127 & \cellcolor{red!10}8.123 & \cellcolor{red!10}14.14 & \cellcolor{red!10}16.26 \\
& \cellcolor{red!10}\mageeditturbo & \cellcolor{red!10}4B & \cellcolor{red!10}4 & \cellcolor{red!10}4.38 & \cellcolor{red!10}\underline{8.271} & \cellcolor{red!10}\textbf{8.264} & \cellcolor{red!10}12.77 & \cellcolor{red!10}15.41 \\
\bottomrule
\end{tabular}
\end{adjustbox}
\endgroup
\end{table*}

\paragraph{Instruction-based image editing.}
We evaluate \mageedit on three instruction-based image editing benchmarks.
ImgEdit-Bench~\citep{ye2026imgedit} covers nine representative editing skills, including Add, Adjust, Extract, Replace, Remove, Background, Style, Hybrid, and Action, and reports category-level performance on a $0$--$5$ scale.
GEdit-Bench~\citep{liu2025step1x} evaluates editing quality from three perspectives: Semantic Consistency, Perceptual Quality, and Overall Score, on both English and Chinese subsets, with each metric measured on a $0$--$10$ scale.
TextEdit-Bench~\citep{gui2025texteditbench} focuses on text-centric editing and evaluates five dimensions: Instruction Following, Text Accuracy, Visual Consistency, Layout Preservation, and Semantic Expectation. Each dimension is scored on a $0$--$5$ scale, with the overall score computed by summing the five metric means (out of $25$).
Unless otherwise specified, \mageeditbase and \mageedit are evaluated with $30$ denoising steps, while \mageeditturbo uses only $4$ denoising steps.

%We evaluate \mageedit on three instruction-based editing benchmarks. ImgEdit-Bench~\citep{ye2026imgedit} covers nine editing skills, including Add, Adjust, Extract, Replace, Remove, Background, Style, Hybrid, and Action, and reports category-wise scores on a $0$--$5$ scale. GEdit-Bench~\citep{liu2025step1x} evaluates semantic consistency, perceptual quality, and overall quality on English and Chinese splits, each on a $0$--$10$ scale. TextEdit-Bench~\citep{gui2025texteditbench} focuses on text editing and reports five dimensions: Instruction Following, Text Accuracy, Visual Consistency, Layout Preservation, and Structure/Style Editing. Each dimension is scored on a $0$--$5$ scale, with an overall score ranging from $0$ to $25$. Unless otherwise specified, \mageeditbase and \mageedit use $30$ denoising steps, while \mageeditturbo uses $4$ steps.

\subsection{Quantitative Results}
\label{sec:quantitative}

\begin{table}[t]
\centering
\caption{Quantitative results on ImgEdit-Bench~\cite{ye2026imgedit}. All metrics are evaluated by GPT-4.1 on 0--5 scale. Overall is the average of all scores across tasks.}
\label{tab:imgedit_benchmark}
\begingroup
\setlength{\tabcolsep}{2.2pt}
\renewcommand{\arraystretch}{1.05}
\scriptsize
\begin{adjustbox}{width=\textwidth,center}
\begin{tabular}{c l|c|ccccccccc|c}
\toprule
Type & \textbf{Model} & \#Params & \textbf{Add} & \textbf{Adjust} & \textbf{Extract} & \textbf{Replace} & \textbf{Remove} & \textbf{Background} & \textbf{Style} & \textbf{Hybrid} & \textbf{Action} & \textbf{Overall}$\uparrow$ \\
\midrule
\multirow{4}{*}{\makecell{Closed-\\Source}}
& Nano-Banana~\cite{nanopro} & -- & 4.62 & 4.41 & 3.68 & 4.34 & 4.39 & 4.40 & 4.18 & 3.72 & 4.83 & 4.29 \\
& Seedream4.0~\cite{seedream2025seedream} & -- & 4.33 & 4.38 & 3.89 & 4.65 & 4.57 & 4.35 & 4.22 & 3.71 & 4.61 & 4.30 \\
& Seedream4.5~\cite{seedream2025seedream} & -- & 4.57 & 4.65 & 2.97 & 4.66 & 4.46 & 4.37 & 4.92 & 3.71 & 4.56 & 4.32 \\
& Nano-Banana-Pro~\cite{nanopro} & -- & 4.44 & 4.62 & 3.42 & 4.60 & 4.63 & 4.32 & 4.97 & 3.64 & 4.69 & 4.37 \\
\cmidrule{1-13}
\multirow{4}{*}{\makecell{Open-Source\\Unified}}
& BAGEL~\cite{deng2025bagel} & 14B & 3.56 & 3.31 & 1.70 & 3.30 & 2.62 & 3.24 & 4.49 & 2.38 & 4.17 & 3.20 \\
& OmniGen2~\cite{wu2025omnigen2} & 4B & 3.57 & 3.06 & 1.77 & 3.74 & 3.20 & 3.57 & 4.81 & 2.52 & 4.68 & 3.44 \\
& UniWorld-V1~\cite{lin2025uniworld} & 19B & 3.82 & 3.64 & 2.27 & 3.47 & 3.24 & 2.99 & 4.21 & 2.96 & 2.74 & 3.26 \\
& Emu3.5~\cite{cui2025emu35nativemultimodalmodels} & 34B & \textbf{4.61} & 4.32 & 3.96 & \textbf{4.84} & 4.58 & 4.35 & 4.79 & 3.69 & 4.57 & 4.41 \\
\cmidrule{1-13}
\multirow{23}{*}{\makecell{Open-Source\\Specialist}}
& Instruct-Pix2Pix~\cite{brooks2023instructpix2pix} & 1B & 2.45 & 1.83 & 1.44 & 2.01 & 1.50 & 1.44 & 3.55 & 1.20 & 1.46 & 1.88 \\
& MagicBrush~\cite{zhang2023magicbrush} & 1B & 2.84 & 1.58 & 1.51 & 1.97 & 1.58 & 1.75 & 2.38 & 1.62 & 1.22 & 1.90 \\
& AnyEdit~\cite{yu2025anyedit} & 1B & 3.18 & 2.95 & 1.88 & 2.47 & 2.23 & 2.24 & 2.85 & 1.56 & 2.65 & 2.45 \\
& UltraEdit~\cite{zhao2024ultraedit} & 2B & 3.44 & 2.81 & 2.13 & 2.96 & 1.45 & 2.83 & 3.76 & 1.91 & 2.98 & 2.70 \\
& OmniGen~\cite{xiao2025omnigen} & 3.8B & 3.47 & 3.04 & 1.71 & 2.94 & 2.43 & 3.21 & 4.19 & 2.24 & 3.38 & 2.96 \\
& ICEdit~\cite{zhang2025icedit} & 12B & 3.58 & 3.39 & 1.73 & 3.15 & 2.93 & 3.08 & 3.84 & 2.04 & 3.68 & 3.05 \\
& Step1X-Edit-v1.2~\cite{liu2025step1x} & 19B & 3.91 & 4.04 & 2.68 & 4.48 & 4.26 & 3.90 & 4.82 & 3.23 & 4.22 & 3.95 \\
& ChronoEdit~\cite{wu2025chronoedit} & 14B & 4.48 & 4.39 & 3.49 & 4.66 & \underline{4.67} & \textbf{4.57} & 4.91 & 3.82 & 4.83 & 4.42 \\
& FLUX.1-Kontext-dev~\cite{labs2025flux} & 12B & 3.99 & 3.88 & 2.19 & 4.27 & 3.13 & 3.98 & 4.51 & 3.23 & 4.18 & 3.71 \\
& FLUX.2-dev~\cite{flux-2-2025} & 32B & 4.50 & 4.18 & 3.83 & 4.65 & 4.65 & 4.31 & 4.88 & 3.46 & 4.70 & 4.35 \\
& FLUX.2-Klein-Base-4B~\cite{flux-2-2025} & 4B & 4.40 & 4.04 & 1.96 & 4.00 & 3.10 & 4.23 & 4.76 & 3.14 & 4.74 & 3.80 \\
& FLUX.2-Klein-4B~\cite{flux-2-2025} & 4B & 4.45 & 4.37 & 2.04 & 4.21 & 3.93 & 4.29 & 4.87 & 3.49 & \underline{4.89} & 4.01 \\
& FLUX.2-Klein-Base-9B~\cite{flux-2-2025} & 9B & 4.40 & 4.22 & 2.31 & 4.50 & 3.84 & 4.26 & 4.89 & 3.27 & \underline{4.89} & 4.05 \\
& FLUX.2-Klein-9B~\cite{flux-2-2025} & 9B & 4.50 & 4.50 & 2.32 & 4.44 & 4.32 & 4.42 & 4.90 & 3.36 & \textbf{4.96} & 4.18 \\
& Z-Image-Edit~\cite{zimage} & 6B & 4.40 & 4.14 & 4.30 & 4.57 & 4.13 & 4.14 & 4.85 & 3.63 & 4.50 & 4.30 \\
& Qwen-Image-Edit-2509~\cite{qwenimage} & 20B & 4.34 & 4.27 & 3.42 & 4.73 & 4.36 & 4.37 & 4.91 & 3.56 & 4.80 & 4.31 \\
& Qwen-Image-Edit-2511~\cite{qwenimage} & 20B & 4.54 & \underline{4.57} & 4.13 & 4.70 & 4.46 & 4.36 & 4.89 & \textbf{4.16} & 4.81 & \underline{4.51} \\
& LongCat-Image-Edit~\cite{LongCat-Image} & 6B & 4.44 & 4.53 & 3.83 & \underline{4.80} & 4.60 & 4.33 & \underline{4.92} & 3.75 & 4.82 & 4.45 \\
& FireRed-Image-Edit-1.0~\cite{firered2025} & 20B & \underline{4.55} & \textbf{4.66} & \textbf{4.34} & 4.75 & 4.58 & \underline{4.45} & \textbf{4.97} & \underline{4.07} & 4.71 & \textbf{4.56} \\
& JoyAI-Image-Edit~\cite{song2026joyai} & 16B & 4.47 & 4.48 & \underline{4.31} & 4.57 & \textbf{4.75} & 4.33 & 4.79 & 3.72 & 4.69 & 4.46 \\
& \cellcolor{red!10}\textbf{\mageeditbase} & \cellcolor{red!10}4B & \cellcolor{red!10}4.28 & \cellcolor{red!10}4.22 & \cellcolor{red!10}4.08 & \cellcolor{red!10}4.45 & \cellcolor{red!10}4.39 & \cellcolor{red!10}4.01 & \cellcolor{red!10}4.82 & \cellcolor{red!10}3.40 & \cellcolor{red!10}4.21 & \cellcolor{red!10}4.28 \\
& \cellcolor{red!10}\textbf{\mageedit} & \cellcolor{red!10}4B & \cellcolor{red!10}4.34 & \cellcolor{red!10}4.27 & \cellcolor{red!10}4.03 & \cellcolor{red!10}4.48 & \cellcolor{red!10}4.50 & \cellcolor{red!10}4.06 & \cellcolor{red!10}4.89 & \cellcolor{red!10}3.49 & \cellcolor{red!10}4.51 & \cellcolor{red!10}4.34 \\
& \cellcolor{red!10}\textbf{\mageeditturbo} & \cellcolor{red!10}4B & \cellcolor{red!10}4.38 & \cellcolor{red!10}4.48 & \cellcolor{red!10}3.94 & \cellcolor{red!10}4.48 & \cellcolor{red!10}4.39 & \cellcolor{red!10}4.26 & \cellcolor{red!10}4.91 & \cellcolor{red!10}3.58 & \cellcolor{red!10}4.48 & \cellcolor{red!10}4.38 \\
\bottomrule
\end{tabular}
\end{adjustbox}
\endgroup
\end{table}

\subsubsection{Text-to-image generation}
\label{sec:t2i_quantitative}

\paragraph{Overall comparison.}
\cref{tab:gen_summary} summarizes the main text-to-image results across prompt following, fine-grained generation, and text rendering benchmarks. With only $4$B parameters, \mageflow achieves the best GenEval score among all compared systems and delivers competitive performance against substantially larger open-source specialist models on DPG-Bench, TIIF-Bench, OneIG, and LongText. It also obtains one of the strongest open-source results on CVTG-2K, approaching the best 32B open-source specialist model FLUX.2-dev \cite{flux-2-2025} while using a much smaller backbone. Compared with unified understanding--generation models, \mageflow consistently performs better across nearly all benchmarks, showing that a compact native-resolution MMDiT can provide strong generation quality without scaling to tens of billions of parameters. The distilled \mageturbo variant preserves most of this performance with only four denoising steps.

\paragraph{Prompt following and compositional understanding.}
\cref{tab:gen_geneval} shows that \mageflow achieves the strongest overall GenEval score, with particularly strong results on single-object generation, two-object generation, counting, and positional relations. On DPG-Bench, reported in \cref{tab:gen_dpg}, \mageflow remains competitive with strong open-source specialist models and clearly outperforms most unified baselines of similar or larger scale. \cref{tab:tiif} further shows the same trend under both short- and long-prompt settings: \mageflow maintains strong instruction-following ability, while \mageturbo retains competitive performance despite using only four sampling steps.

\paragraph{Text rendering and bilingual generation.}
\cref{tab:gen_cvtg} reports the CVTG-2K results, where \mageflow achieves near-best open-source multi-region text-rendering accuracy and substantially outperforms most unified generation models. On LongText, \mageflow performs strongly on English long-text rendering and remains competitive on Chinese, although the Chinese split still leaves room for further data supplementation. \cref{tab:gen_oneig} show that \magebase, \mageflow, and \mageturbo maintain strong alignment and text-rendering ability on both English and Chinese OneIG splits, indicating that the multi-granularity captioning and bilingual data pipeline transfer well to fine-grained generation settings.

\subsubsection{Instruction-based image editing}
\label{sec:edit_quantitative}

\begin{table}[t]
\centering
\caption{
Comparison of Semantic Consistency (G\_SC), Perceptual Quality (G\_PQ), and Overall Score (G\_O) on GEdit-Bench~\cite{liu2025step1x}. GPT-4.1 is used as the evaluator, with all scores normalized to a 0--10 scale. G\_O is calculated as the geometric mean of G\_SC and G\_PQ, averaged overall samples.}
\label{tab:gedit_benchmark}
\scriptsize
\begin{adjustbox}{width=\textwidth,center}
\begin{tabular}{c l|c|ccc|ccc}
\toprule
\multirow{2}{*}{Type} & \multirow{2}{*}{\textbf{Model}} & \multirow{2}{*}{\#Params} & \multicolumn{3}{c|}{\textbf{GEdit-Bench-EN}} & \multicolumn{3}{c}{\textbf{GEdit-Bench-CN}} \\
\cmidrule(lr){4-6} \cmidrule(lr){7-9}
& & & \textbf{G\_SC}$\uparrow$ & \textbf{G\_PQ}$\uparrow$ & \textbf{G\_O}$\uparrow$ & \textbf{G\_SC}$\uparrow$ & \textbf{G\_PQ}$\uparrow$ & \textbf{G\_O}$\uparrow$ \\
\midrule
\multirow{4}{*}{\makecell{Closed-\\Source}}
& Nano-Banana~\cite{nanopro} & -- & 7.396 & 8.454 & 7.291 & 7.540 & 8.424 & 7.399 \\
& Seedream4.0~\cite{seedream2025seedream} & -- & 8.143 & 8.124 & 7.701 & 8.159 & 8.074 & 7.692 \\
& Seedream4.5~\cite{seedream2025seedream} & -- & 8.268 & 8.167 & 7.820 & 8.254 & 8.167 & 7.800 \\
& Nano-Banana-Pro~\cite{nanopro} & -- & 8.102 & 8.344 & 7.738 & 8.135 & 8.306 & 7.799 \\
\cmidrule{1-9}
\multirow{16}{*}{\makecell{Open-Source\\Specialist}}
& Step1X-Edit-v1.2~\cite{liu2025step1x} & 19B & 7.974 & 7.714 & 7.480 & 7.988 & 7.679 & 7.467 \\
& FLUX.1-Kontext-dev~\cite{labs2025flux} & 12B & 7.045 & 7.206 & 6.462 & 1.606 & 7.863 & 1.857 \\
& FLUX.2-dev~\cite{flux-2-2025} & 32B & 7.835 & 8.064 & 7.413 & 7.697 & 8.046 & 7.278 \\
& FLUX.2-Klein-Base-4B~\cite{flux-2-2025} & 4B & 7.574 & 7.640 & 7.081 & 7.526 & 7.759 & 7.102 \\
& FLUX.2-Klein-4B~\cite{flux-2-2025} & 4B & 8.142 & 8.041 & 7.717 & 8.119 & 8.101 & 7.750 \\
& FLUX.2-Klein-Base-9B~\cite{flux-2-2025} & 9B & 8.300 & 7.870 & 7.740 & 8.243 & 7.960 & 7.745 \\
& FLUX.2-Klein-9B~\cite{flux-2-2025} & 9B & 8.592 & 7.993 & 8.040 & 8.550 & 8.106 & 8.055 \\
& Z-Image-Edit~\cite{zimage} & 6B & 8.110 & 7.720 & 7.570 & 8.030 & 7.800 & 7.540 \\
& Qwen-Image-Edit-2509~\cite{qwenimage} & 20B & 7.974 & 7.714 & 7.480 & 7.988 & 7.679 & 7.467 \\
& Qwen-Image-Edit-2511~\cite{qwenimage} & 20B & 8.297 & \underline{8.202} & 7.877 & 8.252 & \underline{8.134} & 7.819 \\
& LongCat-Image-Edit~\cite{LongCat-Image} & 6B & 8.128 & 8.177 & 7.748 & 8.141 & 8.117 & 7.731 \\
& FireRed-Image-Edit-1.0~\cite{firered2025} & 20B & 8.363 & \textbf{8.245} & 7.943 & 8.287 & \textbf{8.227} & 7.887 \\
& JoyAI-Image-Edit~\cite{song2026joyai} & 16B & 8.829 & 8.120 & \textbf{8.276} & 8.618 & 8.110 & \underline{8.125} \\
& \cellcolor{red!10}\textbf{\mageeditbase} & \cellcolor{red!10}4B & \cellcolor{red!10}8.685 & \cellcolor{red!10}7.495 & \cellcolor{red!10}7.860 & \cellcolor{red!10}8.772 & \cellcolor{red!10}7.584 & \cellcolor{red!10}7.970 \\
& \cellcolor{red!10}\textbf{\mageedit} & \cellcolor{red!10}4B & \cellcolor{red!10}\underline{8.893} & \cellcolor{red!10}7.766 & \cellcolor{red!10}8.127 & \cellcolor{red!10}\underline{8.927} & \cellcolor{red!10}7.685 & \cellcolor{red!10}8.123 \\
& \cellcolor{red!10}\textbf{\mageeditturbo} & \cellcolor{red!10}4B & \cellcolor{red!10}\textbf{8.965} & \cellcolor{red!10}7.970 & \cellcolor{red!10}\underline{8.271} & \cellcolor{red!10}\textbf{8.948} & \cellcolor{red!10}8.002 & \cellcolor{red!10}\textbf{8.264} \\
\bottomrule
\end{tabular}
\end{adjustbox}
\end{table}

\begin{table*}[t]
\centering
\caption{
Comparison of Instruction Following (IF), Text Accuracy (TA), Visual Consistency (VC), Layout Preservation (LP), and Semantic Expectation (SE) on the synthetic and real-world subsets of TextEdit-Bench~\cite{gui2025texteditbench}. 
All metrics are scored by GPT-4o on a 0--5 scale, and Overall denotes the sum of the five metric means (out of 25). 
For the synthetic subset, we use reference-based evaluation with ground-truth edited images, measuring preservation fidelity only on non-edited regions. 
For the real-world subset, where paired references are unavailable, we adopt non-reference evaluation by comparing outputs with original inputs outside the masked regions to quantify unintended changes.
}
\label{tab:tebench}
\scriptsize
\setlength{\tabcolsep}{2.8pt}
\begin{adjustbox}{width=\textwidth,center}
\begin{tabular}{c l|c|cccccc|cccccc}
\toprule
\multirow{2}{*}{Type} & \multirow{2}{*}{\textbf{Model}} & \multirow{2}{*}{\#Params} & \multicolumn{6}{c|}{\textbf{Synthetic}} & \multicolumn{6}{c}{\textbf{Real-World}} \\
\cmidrule(lr){4-9} \cmidrule(lr){10-15}
& & & \textbf{IF}$\uparrow$ & \textbf{TA}$\uparrow$ & \textbf{VC}$\uparrow$ & \textbf{LP}$\uparrow$ & \textbf{SE}$\uparrow$ & \textbf{Overall}$\uparrow$ & \textbf{IF}$\uparrow$ & \textbf{TA}$\uparrow$ & \textbf{VC}$\uparrow$ & \textbf{LP}$\uparrow$ & \textbf{SE}$\uparrow$ & \textbf{Overall}$\uparrow$ \\
\midrule
\multirow{2}{*}{\makecell{Closed-\\Source}}
& Nano-Banana~\cite{nanopro} & -- & 2.90 & 3.25 & 3.40 & 4.46 & 2.53 & 16.54 & 3.18 & 3.60 & 3.94 & 4.77 & 2.73 & 18.22 \\
& Seedream4.0~\cite{seedream2025seedream} & -- & 2.67 & 3.44 & 2.97 & 3.25 & 2.57 & 14.90 & 3.64 & 3.96 & 3.90 & 4.42 & 2.62 & 18.54 \\
\cmidrule{1-15}
\multirow{3}{*}{\makecell{Open-Source\\Unified}}
& BAGEL~\cite{deng2025bagel} & 14B & 1.81 & 2.31 & 1.83 & 2.94 & 1.52 & 10.41 & 2.21 & 2.50 & 2.75 & 4.22 & 1.15 & 12.83 \\
& OmniGen2~\cite{wu2025omnigen2} & 4B & 1.12 & 1.74 & 1.72 & 3.08 & 0.86 & 8.52 & 1.67 & 2.22 & 2.45 & 3.63 & 1.02 & 10.99 \\
& Emu3.5~\cite{cui2025emu35nativemultimodalmodels} & 34B & 1.40 & 2.41 & 1.98 & 1.90 & 1.65 & 9.34 & 2.82 & 3.00 & 3.10 & 3.73 & 1.40 & 14.05 \\
\cmidrule{1-15}
\multirow{17}{*}{\makecell{Open-Source\\Specialist}}
& Instruct-Pix2Pix~\cite{brooks2023instructpix2pix} & 1B & 0.90 & 1.14 & 1.14 & 1.61 & 0.72 & 5.51 & 0.56 & 1.22 & 1.09 & 1.69 & 0.79 & 5.35 \\
& MagicBrush~\cite{zhang2023magicbrush} & 1B & 0.73 & 0.73 & 0.60 & 1.18 & 0.62 & 3.86 & 0.59 & 0.80 & 0.64 & 1.42 & 0.67 & 4.12 \\
& Step1X-Edit-v1.2~\cite{liu2025step1x} & 19B & 1.40 & 1.44 & 1.88 & 3.40 & 1.14 & 9.26 & 1.96 & 2.11 & 2.81 & 4.06 & 1.08 & 12.02 \\
& LongCat-Image-Edit~\cite{LongCat-Image} & 6B & 2.18 & 2.27 & 2.51 & 3.87 & 1.64 & 12.46 & 2.92 & 2.96 & 3.32 & 4.43 & 1.26 & 14.89 \\
& FLUX.1-Kontext-dev~\cite{labs2025flux} & 12B & 1.80 & 1.92 & 2.44 & 4.43 & 1.55 & 12.14 & 2.64 & 2.73 & 3.18 & 4.64 & 1.12 & 14.31 \\
& FLUX.2-dev~\cite{flux-2-2025} & 32B & 1.81 & 1.99 & 2.44 & 4.14 & 1.49 & 11.86 & 2.68 & 2.81 & 3.39 & 4.42 & 1.41 & 14.71 \\
& FLUX.2-Klein-Base-4B~\cite{flux-2-2025} & 4B & 1.62 & 1.82 & 2.20 & 4.12 & 1.25 & 11.01 & 2.48 & 2.62 & 3.06 & 4.54 & 1.10 & 13.79 \\
& FLUX.2-Klein-4B~\cite{flux-2-2025} & 4B & 1.84 & 1.90 & 2.39 & 4.21 & 1.51 & 11.84 & 2.58 & 2.76 & 3.30 & 4.60 & 1.22 & 14.46 \\
& FLUX.2-Klein-Base-9B~\cite{flux-2-2025} & 9B & 2.09 & 2.21 & 2.58 & 4.25 & 1.64 & 12.76 & 3.03 & 3.04 & 3.51 & 4.68 & 1.40 & 15.65 \\
& FLUX.2-Klein-9B~\cite{flux-2-2025} & 9B & 2.13 & 2.25 & 2.63 & 4.12 & 1.61 & 12.73 & 3.02 & 3.14 & 3.58 & 4.62 & 1.39 & 15.75 \\
& Qwen-Image-Edit-2509~\cite{qwenimage} & 20B & 2.35 & 2.41 & 2.76 & 4.20 & 1.68 & 13.40 & 3.13 & 3.14 & 3.59 & 4.65 & 1.30 & 15.81 \\
& Qwen-Image-Edit-2511~\cite{qwenimage} & 20B & 2.49 & 2.54 & 2.74 & 4.01 & 1.75 & 13.53 & 3.46 & \underline{3.48} & 3.67 & 4.61 & \underline{1.58} & \underline{16.81} \\
& FireRed-Image-Edit-1.0~\cite{firered2025} & 20B & \textbf{2.88} & \textbf{2.85} & \textbf{3.17} & 4.46 & \underline{1.85} & \textbf{15.19} & \textbf{3.54} & \textbf{3.53} & \textbf{3.95} & \underline{4.74} & 1.47 & \textbf{17.23} \\
& JoyAI-Image-Edit~\cite{song2026joyai} & 16B & \underline{2.68} & \underline{2.76} & \underline{2.92} & \textbf{4.49} & \textbf{1.95} & \underline{14.80} & \underline{3.52} & 3.42 & \underline{3.79} & \textbf{4.83} & \textbf{1.67} & \textbf{17.23} \\
& \cellcolor{red!10}\textbf{\mageeditbase} & \cellcolor{red!10}4B & \cellcolor{red!10}2.53 & \cellcolor{red!10}2.55 & \cellcolor{red!10}2.65 & \cellcolor{red!10}4.20 & \cellcolor{red!10}1.70 & \cellcolor{red!10}13.63 & \cellcolor{red!10}3.18 & \cellcolor{red!10}3.14 & \cellcolor{red!10}3.41 & \cellcolor{red!10}4.61 & \cellcolor{red!10}1.23 & \cellcolor{red!10}15.57 \\
& \cellcolor{red!10}\textbf{\mageedit} & \cellcolor{red!10}4B & \cellcolor{red!10}2.55 & \cellcolor{red!10}2.59 & \cellcolor{red!10}2.75 & \cellcolor{red!10}\underline{4.48} & \cellcolor{red!10}1.77 & \cellcolor{red!10}14.14 & \cellcolor{red!10}3.24 & \cellcolor{red!10}3.24 & \cellcolor{red!10}3.55 & \cellcolor{red!10}\underline{4.74} & \cellcolor{red!10}1.48 & \cellcolor{red!10}16.26 \\
& \cellcolor{red!10}\textbf{\mageeditturbo} & \cellcolor{red!10}4B & \cellcolor{red!10}2.26 & \cellcolor{red!10}2.31 & \cellcolor{red!10}2.52 & \cellcolor{red!10}3.98 & \cellcolor{red!10}1.69 & \cellcolor{red!10}12.77 & \cellcolor{red!10}3.17 & \cellcolor{red!10}3.11 & \cellcolor{red!10}3.42 & \cellcolor{red!10}4.56 & \cellcolor{red!10}1.15 & \cellcolor{red!10}15.41 \\
\bottomrule
\end{tabular}
\end{adjustbox}
\end{table*}

\paragraph{Overall comparison.}
\cref{tab:edit_summary} summarizes the main instruction-based editing results across ImgEdit-Bench, GEdit-Bench, and TextEdit-Bench. With only $4$B parameters, \mageedit delivers competitive editing quality against substantially larger open-source specialist models while using a much smaller backbone. Our editing models attain the best open-source GEdit-Bench Chinese score and a near-best English score, and remain competitive on ImgEdit-Bench and TextEdit-Bench against models with several times more parameters. The distilled \mageeditturbo variant preserves most of this performance with only four denoising steps.

\paragraph{ImgEdit-Bench.}
\cref{tab:imgedit_benchmark} reports category-wise editing results. The Base, RL-aligned, and Turbo variants reach overall scores of 4.28, 4.34, and 4.38, respectively, outperforming earlier instruction-tuned models and remaining competitive with recent MMDiT-based editing systems. The Turbo variant is strongest among our models on this benchmark despite using only four denoising steps, indicating that few-step distillation preserves the teacher's broad editing behavior.

\paragraph{GEdit-Bench.}
As shown in \cref{tab:gedit_benchmark}, all three variants achieve strong overall scores on both language splits. \mageedit reaches 8.127 on English and 8.123 on Chinese, while \mageeditturbo further reaches 8.271 and 8.264 with only four denoising steps. Both variants outperform Qwen-Image-Edit-2511 \cite{qwenimage}, FireRed-Image-Edit-1.0 \cite{firered2025}, and LongCat-Image-Edit \cite{LongCat-Image} under the same benchmark protocol.

\paragraph{TextEdit-Bench.}
TextEdit-Bench isolates text-editing ability, which is only partially reflected by general editing benchmarks such as ImgEdit-Bench and GEdit-Bench. \cref{tab:tebench} consolidates the available results into one table. \mageedit is the strongest of our variants, reaching 14.142 on Synthetic and 16.255 on Real-World, while precise layout preservation and complex text replacement remain important directions for improvement.

\subsection{Qualitative Results}
\label{sec:qualitative}

\subsubsection{Text-to-image visualizations}
\label{sec:t2i_qualitative}

\begin{figure*}[p]
\centering
\includegraphics[width=\textwidth,height=0.9\textheight,keepaspectratio]{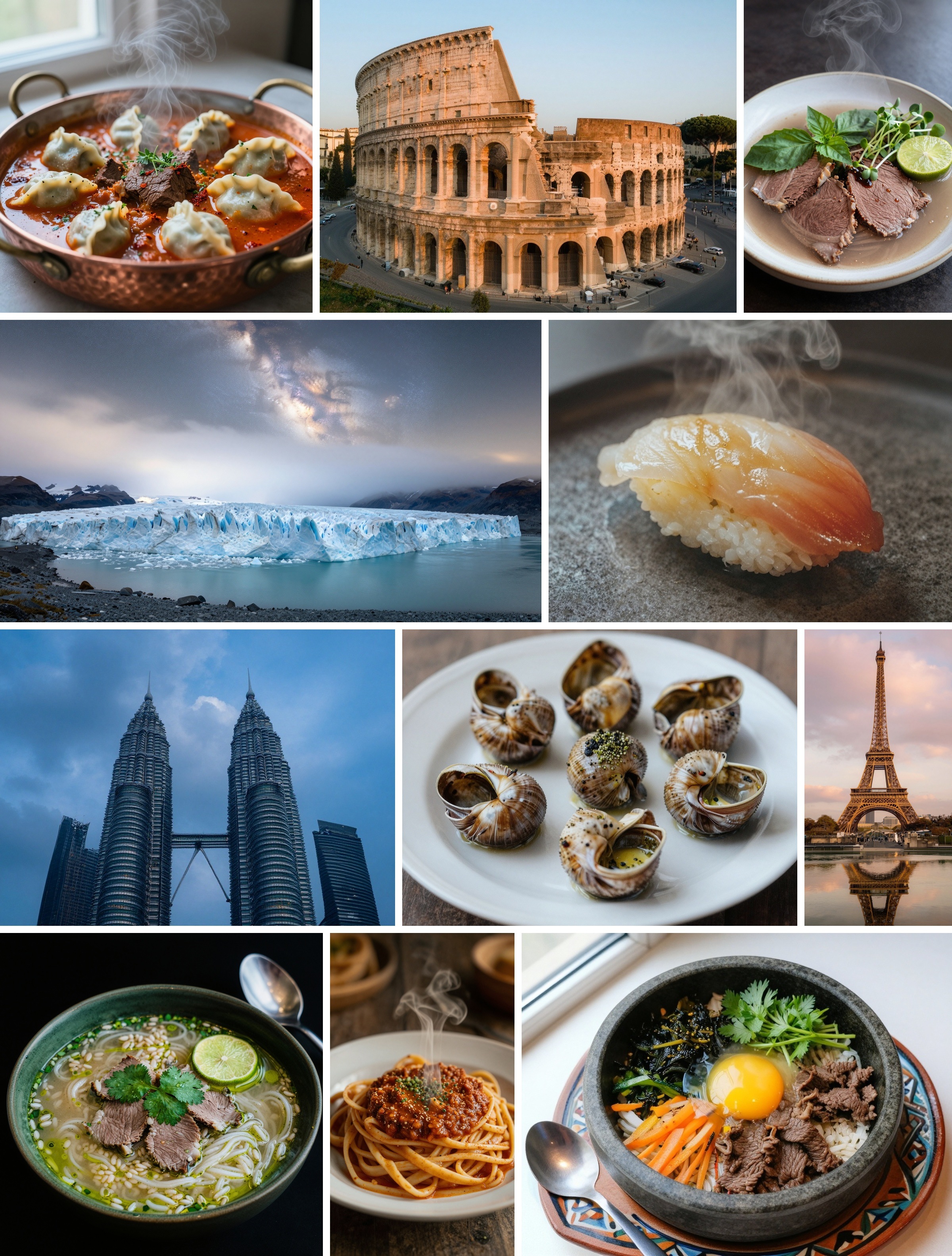}
\caption{\textbf{\mageflow{} gallery: cuisine \& world landmarks.} Native-resolution \magebase{}
samples of plated dishes from a range of cuisines alongside iconic architecture, justified into a
mosaic and shown uncropped at their generated aspect ratios. Consistent lighting, material detail,
and depth of field hold across close-up food and wide architectural scenes.}
\label{fig:gallery_cuisine}
\end{figure*}

\begin{figure*}[p]
\centering
\includegraphics[width=\textwidth,height=0.9\textheight,keepaspectratio]{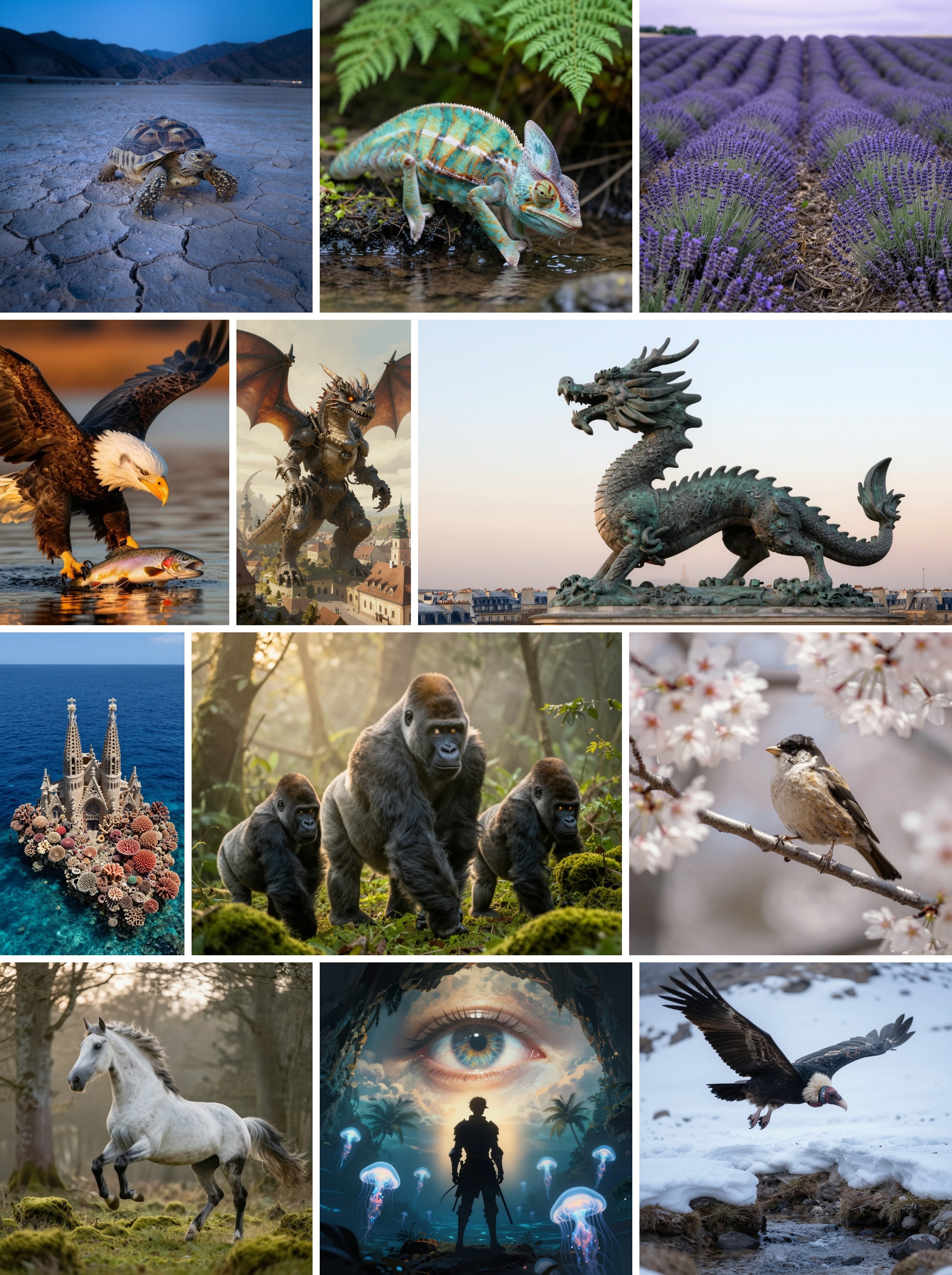}
\caption{\textbf{\mageflow{} gallery: general \& imaginative concepts.} A mix of wildlife, natural
scenery, and fantastical subjects (mechanical and sculptural creatures, surreal composites),
demonstrating broad subject coverage and coherent structure under widely varying styles.}
\label{fig:gallery_general}
\end{figure*}

\begin{figure*}[p]
\centering
\includegraphics[width=\textwidth,height=0.9\textheight,keepaspectratio]{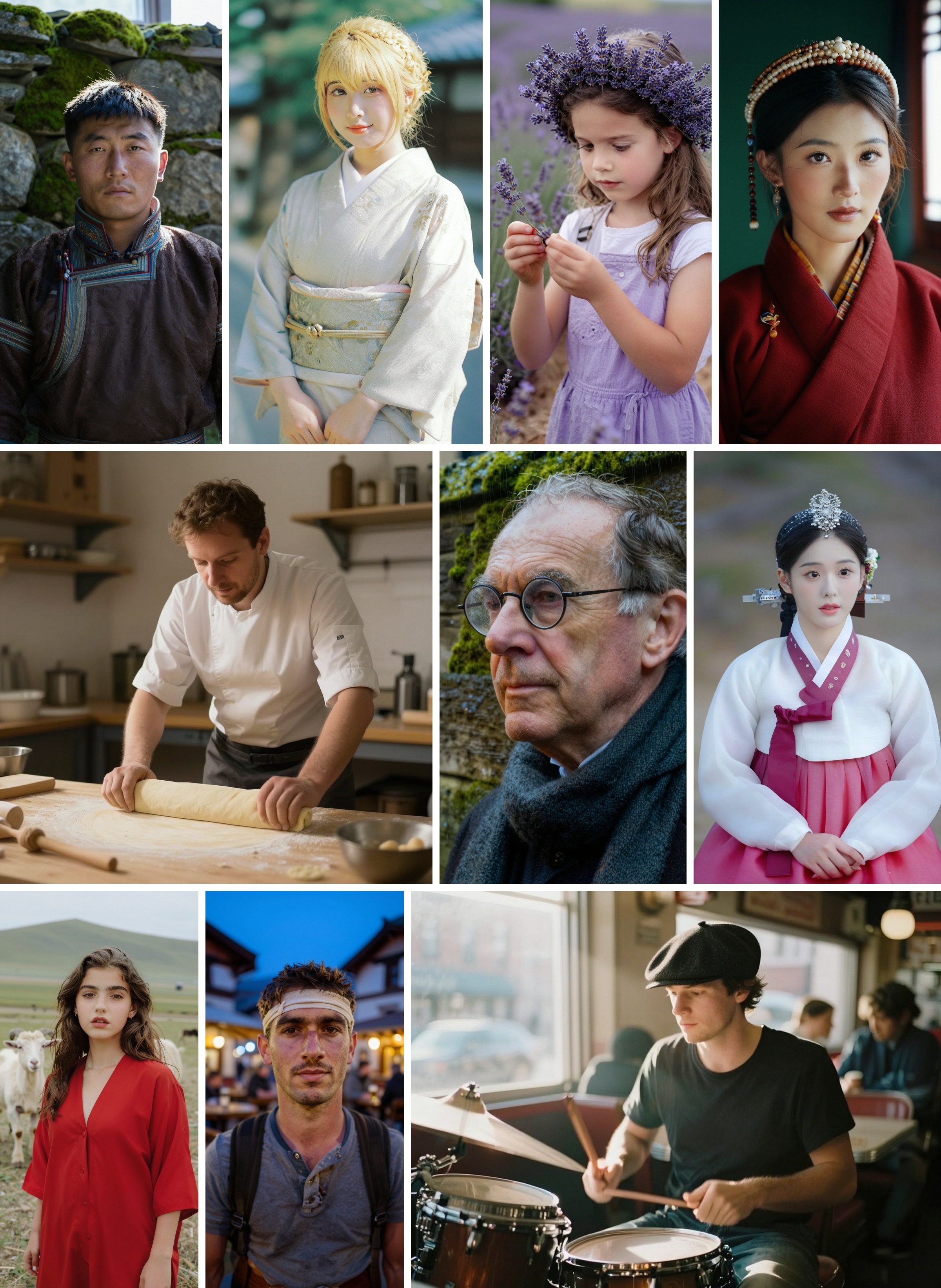}
\caption{\textbf{\mageflow{} gallery: portraits \& people.} Human subjects spanning ages, genders,
ethnicities, and cultural dress, in both studio and environmental settings, with plausible anatomy,
natural skin and hair detail, and controlled lighting.}
\label{fig:gallery_portrait}
\end{figure*}

\begin{figure*}[p]
\centering
\includegraphics[width=\textwidth,height=0.9\textheight,keepaspectratio]{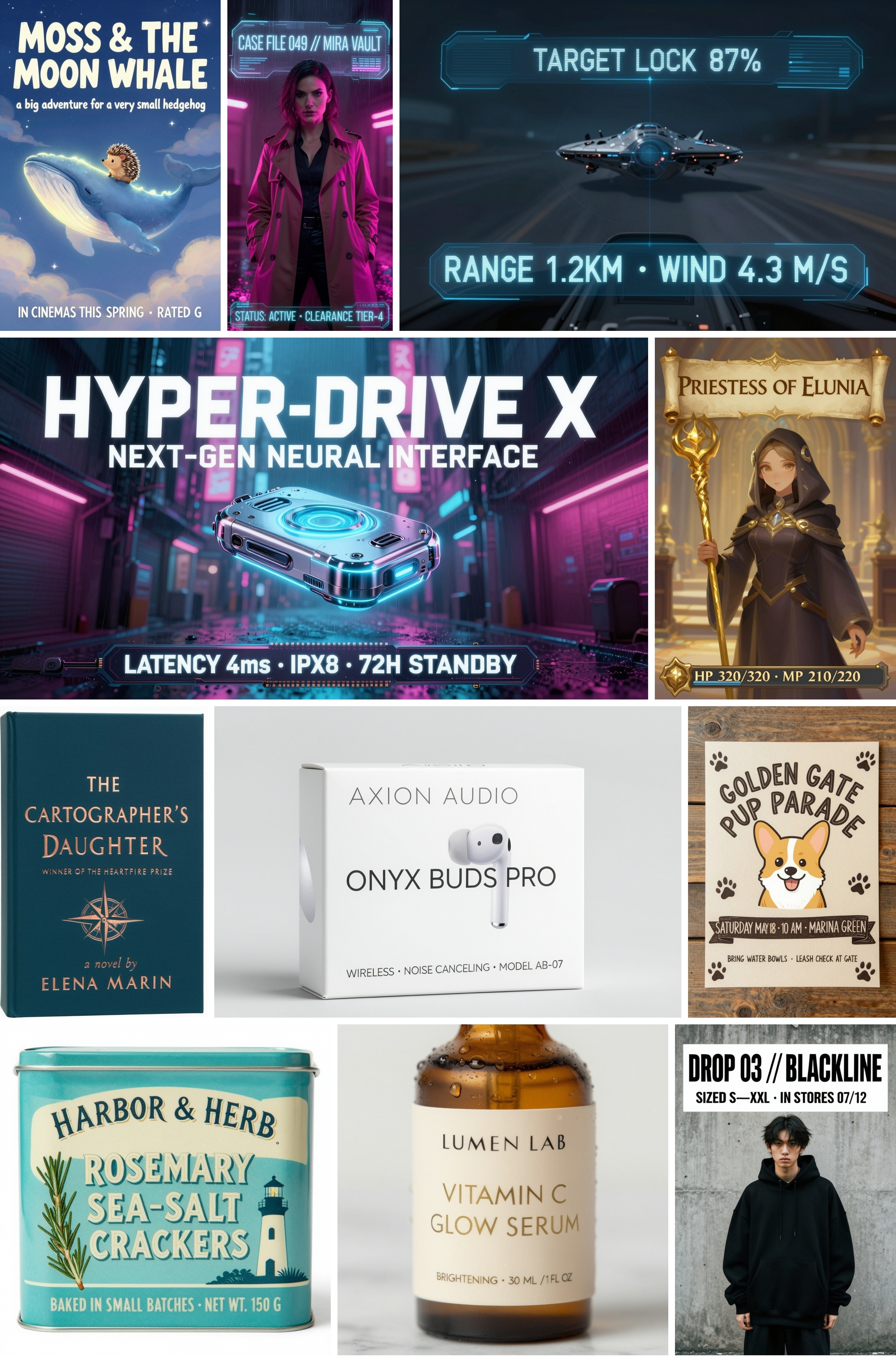}
\caption{\textbf{\mageflow{} gallery: English text rendering.} Product packaging, book and magazine
covers, posters, and signage. Multi-line English copy---brand names, taglines, and fine print---is
reproduced \emph{verbatim} from the prompt and stays legible down to the smallest lines at full
resolution.}
\label{fig:gallery_text_en}
\end{figure*}

\begin{figure*}[p]
\centering
\includegraphics[width=\textwidth,height=0.9\textheight,keepaspectratio]{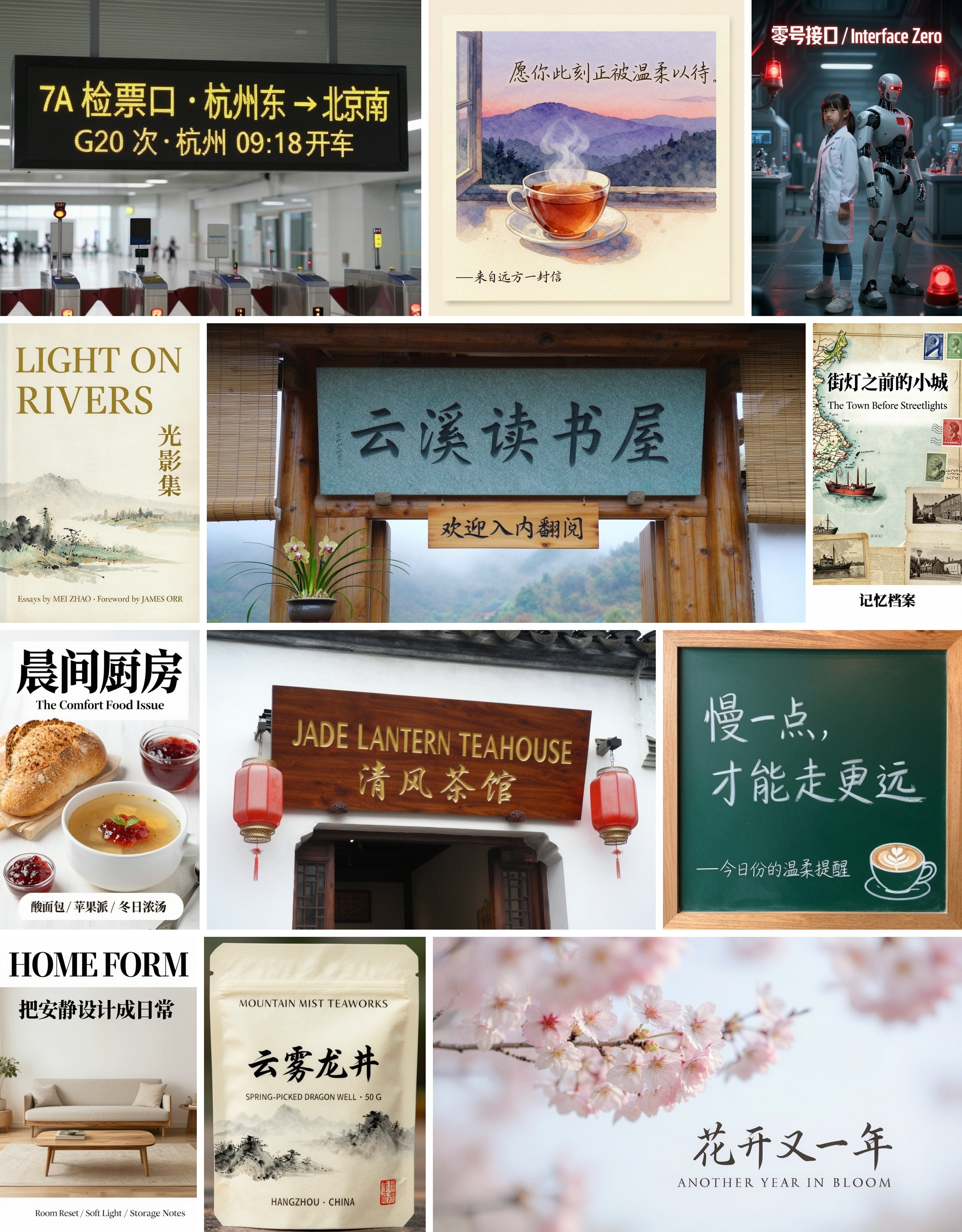}
\caption{\textbf{\mageflow{} gallery: Chinese \& multilingual text rendering.} Signage, packaging,
and covers carrying Chinese---and several bilingual Chinese/English---layouts. The model reproduces
the prompted characters, including denser glyphs, without the stroke-level artifacts typical of
diffusion models.}
\label{fig:gallery_text_zh}
\end{figure*}

%,fig:gallery_general,fig:gallery_portrait,fig:gallery_text_en,fig:gallery_text_zh} 
Fig.~\ref{fig:gallery_cuisine} to \ref{fig:gallery_text_zh}
show representative \mageflow generations at native resolutions and aspect ratios. The first three galleries cover broad photorealistic and imaginative generation, including cuisine, landmarks, general concepts, and portraits. The last two galleries focus on English and Chinese text rendering, where \mageflow produces legible text on packaging, posters, magazine covers, and signage rather than text-like textures.

\subsubsection{Instruction-based editing visualizations}
\label{sec:edit_qualitative}

\begin{figure*}[p]
\centering
\includegraphics[width=\textwidth,height=0.88\textheight,keepaspectratio]{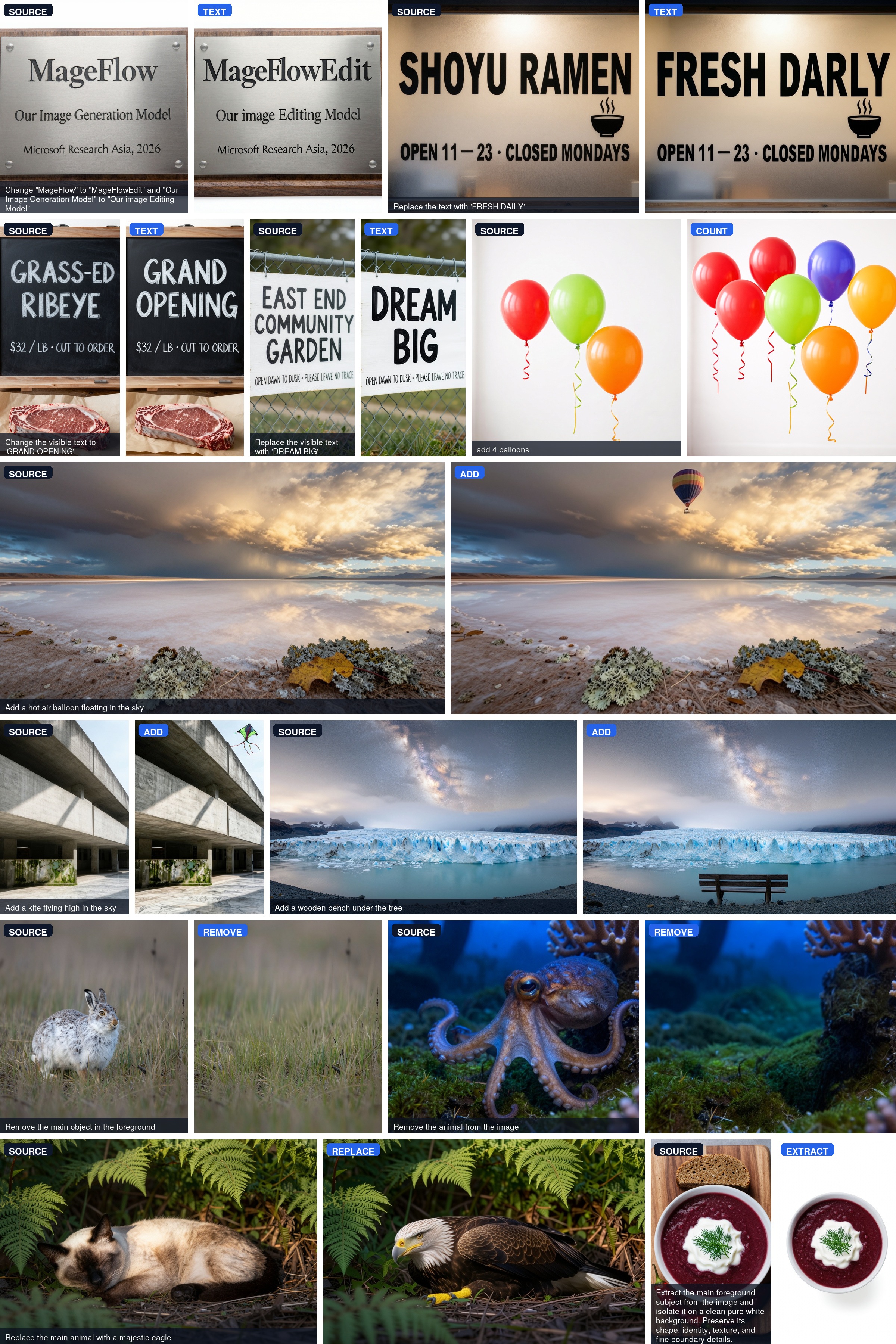}
\caption{\textbf{Qualitative gallery of \mageedit{}: localized content and object editing.}}
\label{fig:edit_gallery_content}
\end{figure*}

\begin{figure*}[p]
\centering
\includegraphics[width=\textwidth,height=0.88\textheight,keepaspectratio]{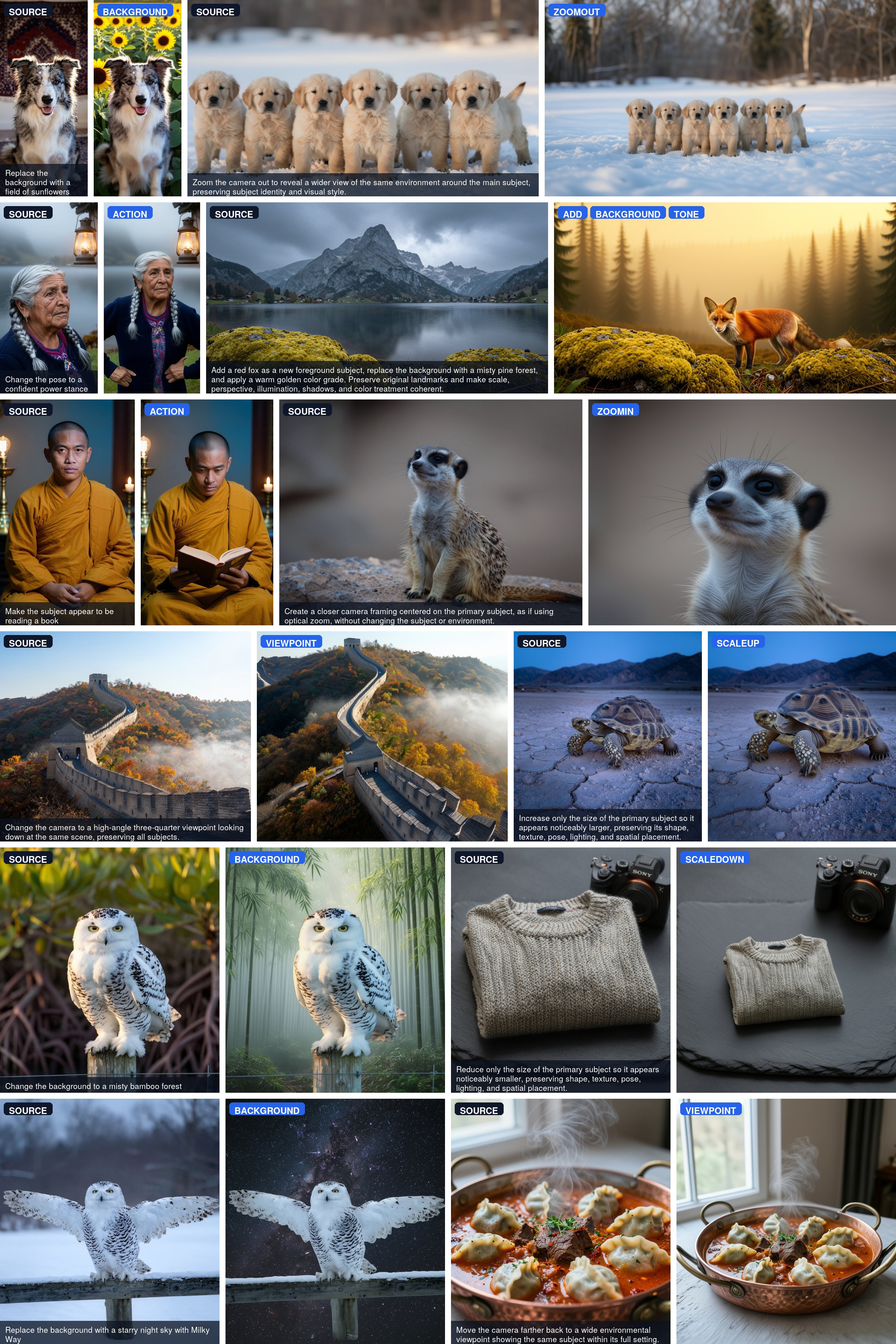}
\caption{\textbf{Qualitative gallery of \mageedit{}: scene, subject, and camera transformations.}}
\label{fig:edit_gallery_scene_subject}
\end{figure*}

\begin{figure*}[p]
\centering
\includegraphics[width=\textwidth,height=0.88\textheight,keepaspectratio]{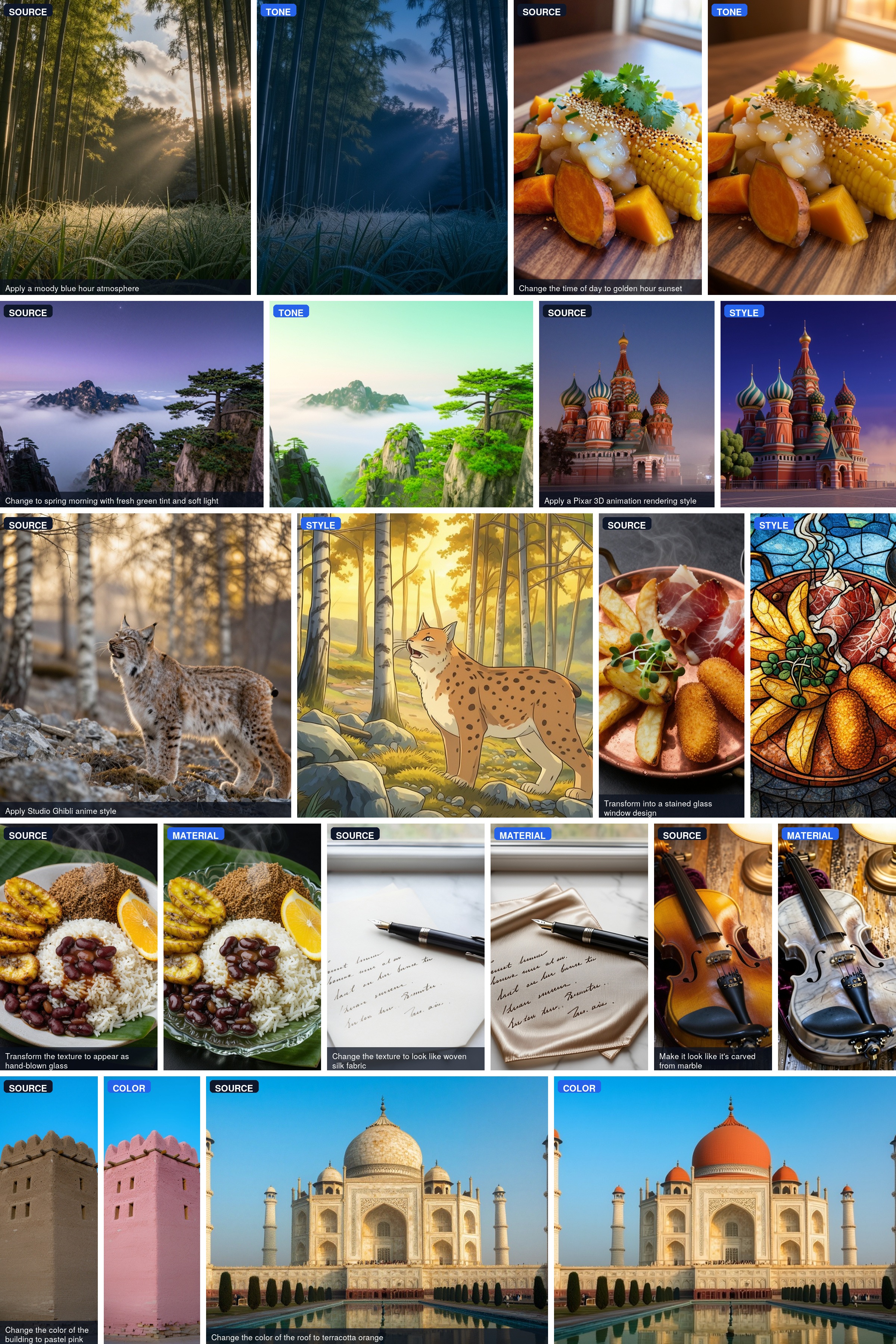}
\caption{\textbf{Qualitative gallery of \mageedit{}: appearance and artistic rendering.}}
\label{fig:edit_gallery_appearance}
\end{figure*}

\begin{figure*}[p]
\centering
\includegraphics[width=\textwidth,height=0.88\textheight,keepaspectratio]{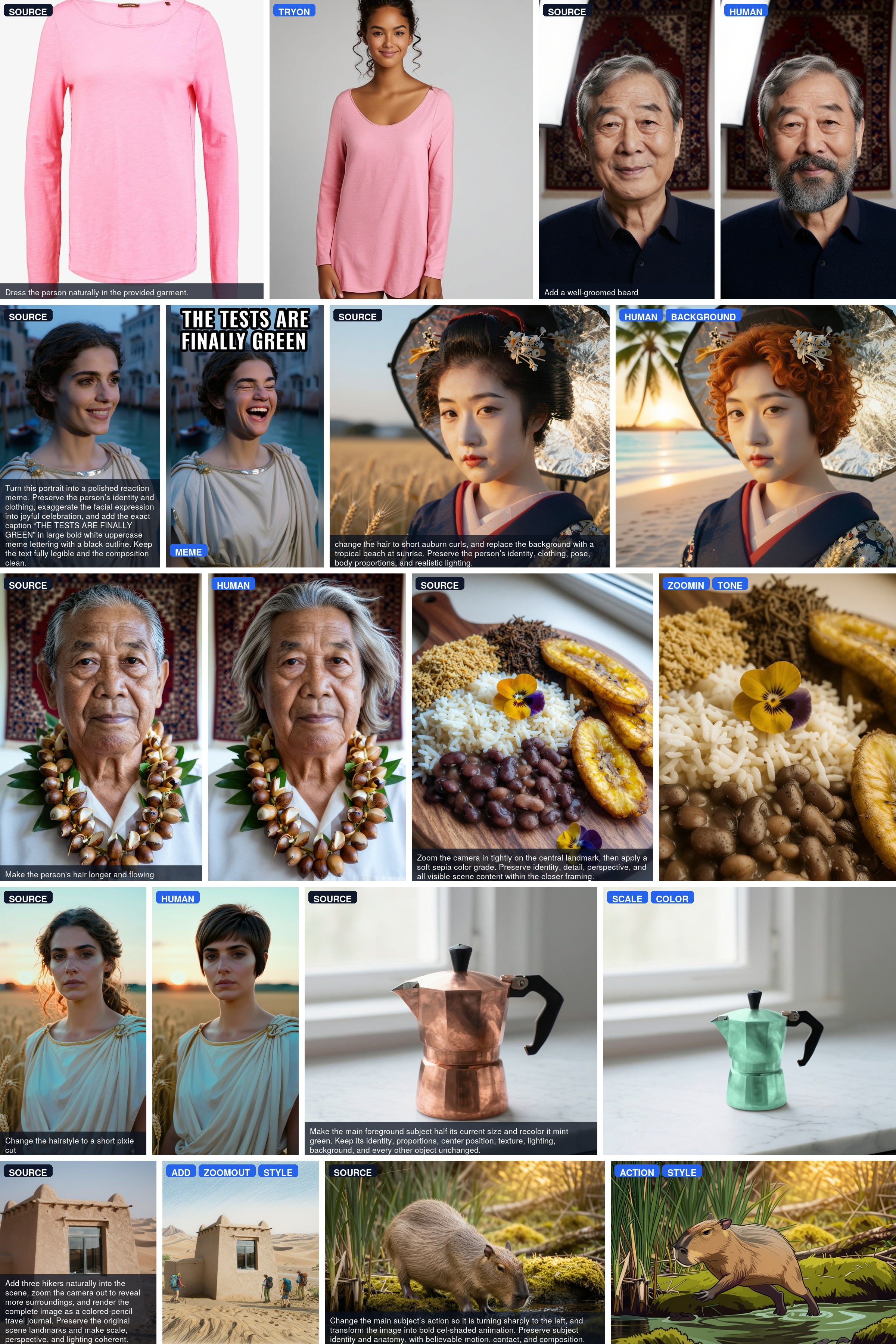}
\caption{\textbf{Qualitative gallery of \mageedit{}: human-centered and creative editing.}}
\label{fig:edit_gallery_human_creative}
\end{figure*}

\begin{figure*}[p]
\centering
\includegraphics[width=\textwidth,height=0.88\textheight,keepaspectratio]{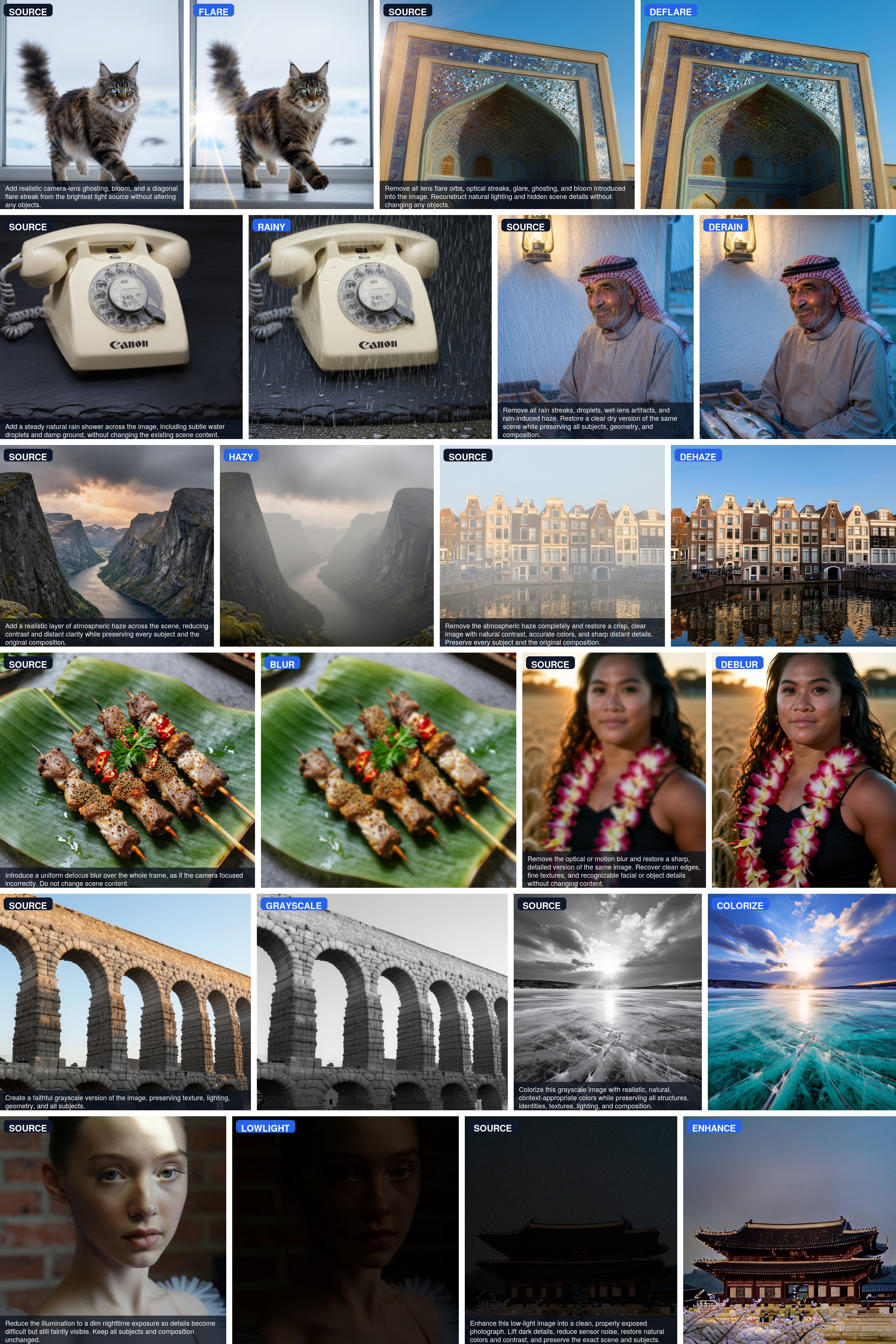}
\caption{\textbf{Qualitative gallery of \mageedit{}: bidirectional degradation and restoration.}}
\label{fig:edit_gallery_restoration}
\end{figure*}

Fig.~\ref{fig:edit_diversity} provides a high-level overview of \mageedit across semantic content editing, appearance transformation, restoration, and structure-aware output tasks. Fig.~\ref{fig:edit_gallery_showcase_1} and \ref{fig:edit_gallery_showcase_2} then present two compact source-centric overviews in which each source image is paired with multiple edited outputs. All showcased edit types are supported bidirectionally. As shown in Fig.~\ref{fig:edit_gallery_content} to \ref{fig:edit_gallery_restoration}, 
the six larger multi-source galleries provide broader coverage of content and object editing, scene and camera transformations, appearance, human-centered and creative editing, composite multi-task edits, low-level control, and paired degradation--restoration tasks. Throughout these galleries, dark labels denote source images and blue labels indicate the edit type applied to each output.
%,fig:edit_gallery_scene_subject,fig:edit_gallery_appearance,fig:edit_gallery_human_creative,fig:edit_gallery_lowlevel,fig:edit_gallery_restoration})

% \input{sections/4_text_to_image}
% \input{sections/5_image_edit}
%
\section{Conclusion}
\label{sec:conclusion}

We presented the \mageflow generative stack, a compact and efficient 4B-scale framework for text-to-image generation and instruction-based image editing. Instead of relying on continued backbone scaling, \mageflow is built through the joint design of a lightweight latent tokenizer, a native-resolution diffusion Transformer, and system-level training infrastructure. \magevae substantially reduces the cost of latent encoding and decoding while preserving reconstruction fidelity. NR-MMDiT supports flexible training and inference across resolutions and aspect ratios through native-resolution packing. Fused CUDA kernels further remove memory-bound overhead from the repeated blocks of the tokenizer, text encoder, and diffusion backbone. Together, these designs make large-scale native-resolution training practical under a fixed compute budget.

On this foundation, we build a complete generation-and-editing model family, including Base, RL-aligned, and Turbo variants for both \mageflow and \mageedit. Diffusion-NFT improves alignment with prompt-following, text-rendering, aesthetic, and editing preferences, while Decoupled-DMD distillation with adversarial perceptual guidance produces 4-step Turbo models for low-latency inference. Across standard generation and editing benchmarks, the resulting models achieve competitive performance against substantially larger open-source systems, while maintaining low memory usage and fast inference in both generation and editing settings.

These results show that strong visual generation does not necessarily require tens-of-billions-parameter backbones. With an efficient tokenizer, native-resolution modeling, and carefully optimized training infrastructure, a 4B-scale stack can serve as a practical and research-friendly foundation for image generation, controllable editing, post-training alignment, and vertical-domain applications. Future work will explore more robust multi-image editing, stronger multilingual long-text rendering, and tighter integration with agentic visual creation workflows.

\section*{Contributor List}
\label{sec:contributor}

\noindent\textbf{Contributors:} Xinjie Zhang$^{*\dagger}$, Peng Zhang$^{*}$, Shicheng Zheng$^{*}$, Jinghao Guo$^{*}$, Zhaoyang Jia$^{*}$, Yifei Shen$^{*}$, Xun Guo, Yuxuan Luo, Jiahao Li, Wenxuan Xie, Fanyi Pu, Xiaoyi Zhang, Kaichen Zhang, Zongyu Guo, Tianci Bi, Dongnan Gui, Zhening Liu, Zimo Wen, Zihan Zheng, Senqiao Yang, Xiao Li, Jinglu Wang, Bin Li, Yan Lu

\vspace{0.5em}
\noindent{\footnotesize $^{*}$Equal Contribution. \quad $^{\dagger}$Project Lead (xinjiezhang@microsoft.com).}

\begin{figure*}[p]
\centering
\includegraphics[width=\textwidth,height=0.88\textheight,keepaspectratio]{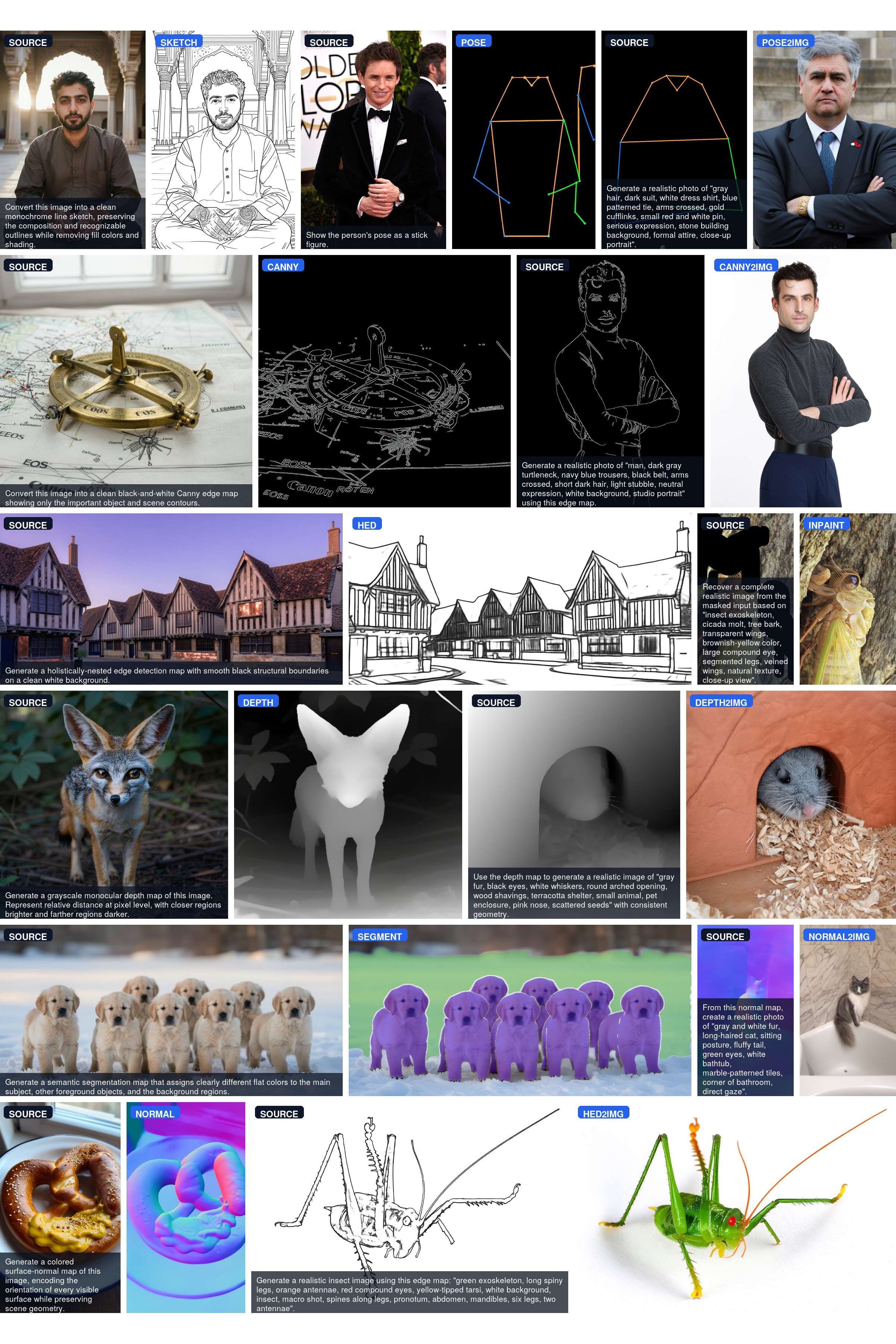}
\caption{\textbf{Qualitative gallery of \mageedit{}: low-level vision and conditional reconstruction.}}
\label{fig:edit_gallery_lowlevel}
\end{figure*}

\clearpage
\appendix

\section{Detailed \magevae Training Process}
\label{app:vae_training}

As described in Section~\ref{sec:model_vae_train}, the training of \magevae proceeds in three stages. Let $x$ denote an input image and
\begin{equation}
    z^{a} = \mathcal{P}\!\left(\mathcal{E}_{\mathrm{FLUX}}(x)\right)
\end{equation}
denote its anchor latent, where $\mathcal{E}_{\mathrm{FLUX}}$ is the frozen FLUX.2-VAE encoder and $\mathcal{P}$ denotes latent patchification.

\paragraph{Stage I: Multi-step flow-matching pre-training.}
We first pre-train both the encoder and decoder as multi-step diffusion models using flow matching in the $\mathcal{X}$-prediction parameterization. For a target $y_0$, we sample $\epsilon\sim\mathcal{N}(0,I)$ and $t\sim\mathcal{U}(0,1)$, and construct the linear interpolation
\begin{equation}
    y_t = (1-t)y_0 + t\epsilon.
\end{equation}
Instead of directly predicting the velocity field, the network predicts the clean endpoint $y_0$:
\begin{equation}
    \mathcal{L}_{\mathrm{FM}}^{\mathcal{X}}(\theta)
    =
    \mathbb{E}_{y_0,\epsilon,t}
    \left[
        \left\|
        \mathcal{X}_{\theta}(y_t,t\mid c)-y_0
        \right\|_2^2
    \right].
    \label{eq:magevae_xpred}
\end{equation}
The corresponding velocity field used for ODE sampling can be recovered as
\begin{equation}
    v_{\theta}(y_t,t\mid c)
    =
    \frac{y_t-\mathcal{X}_{\theta}(y_t,t\mid c)}{t}.
\end{equation}
For the decoder, the prediction target and condition are $(y_0,c)=(x,z^a)$, so that the model learns to reconstruct image pixels from anchor latents. For the encoder, they are $(y_0,c)=(z^a,x)$, such that the encoder learns to generate patchified FLUX.2-VAE anchor latents from image pixels. The Stage-I objective is therefore
\begin{equation}
    \mathcal{L}_{\mathrm{I}}
    =
    \mathcal{L}_{\mathrm{FM,dec}}^{\mathcal{X}}
    +
    \mathcal{L}_{\mathrm{FM,enc}}^{\mathcal{X}}.
\end{equation}

\paragraph{Stage II: One-step decoder distillation.}
We next distill the multi-step decoder into a one-step decoder $\mathcal{D}_{\psi}$. Given an anchor latent $z^a$, the reconstruction is obtained with a single network evaluation:
\begin{equation}
    \hat{x}=\mathcal{D}_{\psi}(z^a).
\end{equation}
We combine pixel-level and perceptual reconstruction objectives,
\begin{equation}
    \mathcal{L}_{\mathrm{rec}}
    =
    \|x-\hat{x}\|_1
    +
    \mathcal{L}_{\mathrm{LPIPS}}(x,\hat{x}),
\end{equation}
with a DINOv2-projected GAN loss~\citep{sauer2021projected} and a DMD loss~\citep{yin2024dmd}. The projected discriminator operates on features extracted by a frozen DINOv2 network, encouraging the reconstruction to match the real-image distribution in a perceptually meaningful feature space. Meanwhile, DMD transfers the distribution-level prior of the compression-oriented diffusion teacher~\citep{jia2026cod} to the one-step decoder. The complete Stage-II decoder objective is
\begin{equation}
    \mathcal{L}_{\mathrm{II}}
    =
    \underbrace{
        \|x-\hat{x}\|_1
        +
        \mathcal{L}_{\mathrm{LPIPS}}(x,\hat{x})
    }_{\mathcal{L}_{\mathrm{rec}}}
    +
    0.01\,\mathcal{L}_{\mathrm{GAN}}^{\mathrm{DINO}}
    +
    0.1\,\mathcal{L}_{\mathrm{DMD}}.
    \label{eq:magevae_stage2}
\end{equation}
The diffusion teacher and DINOv2 feature extractor remain frozen during distillation, while the projected discriminator and the fake-score model used by DMD are optimized with their respective objectives.

\paragraph{Stage III: Joint one-step VAE optimization.}
Finally, we convert the encoder into a one-step model and optimize it together with the distilled decoder. Let
\begin{equation}
    q_{\phi}(z\mid x)
\end{equation}
denote the latent distribution produced by the one-step encoder and let $q_a(z\mid x)$ denote the corresponding frozen FLUX.2-VAE anchor-latent distribution. We regularize the learned latent space using
\begin{equation}
    \mathcal{L}_{\mathrm{KL}}
    =
    \mathbb{E}_{x}
    \left[
        D_{\mathrm{KL}}
        \left(
            q_{\phi}(z\mid x)
            \,\|\,q_a(z\mid x)
        \right)
    \right].
    \label{eq:magevae_anchor_kl}
\end{equation}
This preserves the structure of the pretrained anchor latent space and prevents the encoder distribution from drifting during perceptual fine-tuning.

\subparagraph{Stage III.1: Encoder warm-up with a fixed decoder.}
We first freeze the one-step decoder and optimize only the encoder. For
\begin{equation}
    z\sim q_{\phi}(z\mid x),
    \qquad
    \hat{x}=\mathcal{D}_{\psi}(z),
\end{equation}
the encoder is trained using the complete Stage-II reconstruction and perceptual objective together with the anchor-latent KL regularizer:
\begin{equation}
    \mathcal{L}_{\mathrm{III.1}}
    =
    \mathcal{L}_{\mathrm{II}}
    +
    0.1\,\mathcal{L}_{\mathrm{KL}},
    \qquad
    \psi\ \text{fixed}.
    \label{eq:magevae_stage31}
\end{equation}
Although the decoder parameters are frozen, gradients from all components of $\mathcal{L}_{\mathrm{II}}$ are back-propagated through the decoder to train the encoder. This warm-up stage aligns the one-step encoder with the already distilled decoder without perturbing the decoder's reconstruction ability.

\subparagraph{Stage III.2: End-to-end fine-tuning.}
After encoder warm-up, we unfreeze the decoder and jointly optimize the complete one-step encoder--decoder pipeline:
\begin{equation}
    \mathcal{L}_{\mathrm{III.2}}
    =
    \mathcal{L}_{\mathrm{II}}
    +
    0.1\,\mathcal{L}_{\mathrm{KL}},
    \qquad
    \phi,\psi\ \text{jointly optimized}.
    \label{eq:magevae_stage32}
\end{equation}
End-to-end fine-tuning allows the encoder and decoder to co-adapt while the KL term maintains compatibility with the anchor-latent space. The resulting \magevae provides one-step encoding and decoding, high-fidelity reconstruction, and a compact, regularized latent space suitable for subsequent \mageflow training.

\section{VAE Latency Across Resolutions}
\label{app:vae_latency}

Reconstruction quality alone does not determine whether a latent tokenizer is practical: native-resolution training and high-resolution inference require the VAE to encode and decode large images repeatedly, so its latency directly bounds end-to-end throughput. \cref{tab:vae_latency} compares the encoding and decoding latency of \magevae against representative open-source image and video VAEs across resolutions from $512^2$ to $4096^2$, all measured on an 80GB NVIDIA A100 GPU under bf16 precision, complementing the plots in \cref{fig:vae_speed} with exact numbers.

\begin{table}[t]
\centering
%\captionsetup{justification=centering}
\caption{\textbf{VAE encoding\,/\,decoding latency across resolutions.} Each cell reports encoding\,/\,decoding latency in milliseconds (lower is better), measured on an 80GB NVIDIA A100 GPU under bf16 precision. \magevae is the fastest at every resolution and scales gracefully to $4096^2$, whereas other VAEs slow down by orders of magnitude or run out of memory (OOM).}
\label{tab:vae_latency}
\small
\setlength{\tabcolsep}{6pt}
\renewcommand{\arraystretch}{1.1}
\begin{tabular}{l cccc}
\toprule
\multirow{2}{*}{\textbf{Method}} & \multicolumn{4}{c}{\textbf{Encoding\,/\,Decoding Latency (ms)}} \\
\cmidrule(lr){2-5}
 & $\mathbf{512^2}$ & $\mathbf{1024^2}$ & $\mathbf{2048^2}$ & $\mathbf{4096^2}$ \\
\midrule
\rowcolor{red!10} \textbf{\magevae} & \textbf{18.0\,/\,23.4} & \textbf{19.1\,/\,33.8} & \textbf{36.5\,/\,92.4} & \textbf{149.6\,/\,375.0} \\
FLUX.2-VAE~\cite{flux-2-2025} & 19.9\,/\,37.8 & 83.8\,/\,153.3 & 491\,/\,782 & 47\,323\,/\,183\,920 \\
Qwen-Image-VAE~\cite{qwenimage} & 22.9\,/\,36.0 & 88.3\,/\,137.9 & 458\,/\,654 & 106\,695\,/\,192\,042 \\
HunyuanVideo-VAE~\cite{kong2024hunyuanvideo} & 46.4\,/\,91.7 & 185.7\,/\,364.3 & 948\,/\,40\,705 & OOM\,/\,OOM \\
HunyuanImage-3.0-VAE~\cite{hunyuanimage3} & 126.6\,/\,217.8 & 508.2\,/\,877.1 & 129\,803\,/\,220\,498 & OOM\,/\,OOM \\
\bottomrule
\end{tabular}
\end{table}

\magevae is the fastest at every resolution, and its advantage widens sharply as resolution grows. At $512^2$ all methods are broadly comparable, but by $1024^2$ \magevae already encodes\,/\,decodes in $19.1/33.8$\,ms, versus $83.8/153.3$\,ms for FLUX.2-VAE and $88.3/137.9$\,ms for Qwen-Image-VAE. At $2048^2$ the gap reaches roughly an order of magnitude, and at $4096^2$ \magevae still runs in $149.6/375.0$\,ms while FLUX.2-VAE and Qwen-Image-VAE require tens to hundreds of seconds ($47.3/183.9$\,s and $106.7/192.0$\,s, respectively), and the video-based HunyuanVideo-VAE and HunyuanImage-3.0-VAE run out of memory. This near-flat scaling is a direct consequence of the single-step, memory-efficient design of \magevae, and it is what makes native-resolution packing and high-resolution generation practical under a fixed compute budget.

\section{Recaptioning System Prompt}
\label{app:caption_prompt}

As described in \cref{sec:data_t2i}, we recaption each retained image at multiple granularities with
Qwen3-VL-32B-Instruct~\citep{qwen3vl}. The four caption layers introduced there, namely the entity,
phrase, composition, and photographic descriptions, are produced by the single system prompt below in
one pass and returned as a JSON object. The prompt fixes the objectivity and consistency principles, a
step-by-step captioning procedure, the JSON output schema for the four layers, and a style guide, so
that the four granularities stay mutually consistent.
\begin{tcblisting}{enhanced,colback=promptbg,colframe=promptborder,arc=3pt,boxrule=0.8pt,left=8pt,right=8pt,top=4pt,bottom=4pt,fonttitle=\bfseries\sffamily,title={Multi-granularity recaptioning system prompt},listing only,listing engine=listings,listing options={basicstyle=\ttfamily\fontsize{6pt}{7pt}\selectfont,breaklines=true,columns=fullflexible,breakindent=0pt,keepspaces=true}}
You are a world-class multi-granularity image captioning expert.
Your task is to analyze the given image and produce a structured, detailed, and objective description following these principles:

### Core Principles
1. Absolute Objectivity: Describe only what is visually present. Avoid subjective words like "beautiful" or "sad." Express aesthetic qualities through concrete descriptions of color, light, shadow, and composition.
2. Physical and Logical Consistency: Ensure all elements adhere to real-world physics and spatial logic (e.g., shadows align with light sources, objects rest naturally).
3. Structured Description: Start from the overall scene, then move to background, mid-ground, and foreground. Use directional terms like "left side," "center," "background" to clarify spatial layout.
4. Present Tense: Describe as if observing the image now (e.g., "A man stands," "Light falls on...").
5. Rich and Specific Language: Use precise adjectives for quantity, size, shape, color, and texture. Avoid vague terms.

### Steps
1. Identify the most important part of the image: Specify the subject(s), their color, quantity, and position.
2. Describe the background: Indicate environment type (e.g., mountains, sea, urban, indoor cafe), lighting, and atmosphere.
3. Detail composition and style: Include artistic medium, perspective, and any stylistic elements (e.g., oil painting, anime, photorealistic).
4. Extract subject information as phrases: Provide a concise set of keywords describing the subject.
5. Summarize the image information: Combine all details into a coherent summary.

### Output Format (JSON with Four Layers and Guidelines)
{
  "Entity": "Describe the most important subject(s) in the image with clear identification. This should be a short phrase naming the main object or scene (e.g., 'Hollywood Sign in Los Angeles', 'A vibrant mosaic artwork').",
  "Phrase": "Provide a set of concise keywords or short phrases that capture the subject's attributes, colors, shapes, and related objects. These should be comma-separated and reflect what is visually present (e.g., 'iconic landmark, Hollywood Sign, cloudy sky, modern house, natural surroundings'). Very similar phrase only appear once.",
  "Composition": "Write one single sentence that objectively summarizes the main subject and its context, mentioning only what is visually present in the image, including the setting or environment, without adding lighting, perspective, texture, or any subjective interpretation.",
  "Photographic": "Provide a detailed artistic and technical description of the image in several coherent sentences, explicitly mentioning the image category, inferred style (reinforced), any recognized text (OCR), main entity, named entities, lighting and color conditions, camera composition, and notable photography techniques, ensuring the description is objective and visually grounded."
}

### Style Guide
- Infer the style from the image content (e.g., photorealistic, cinematic, anime, oil painting, minimalist, cyberpunk).
- If the style is unclear, default to photorealistic photography.
- Apply the inferred style consistently in the Composition and Photographic layers.
- Mention the style explicitly in the Photographic description and reinforce it through lighting, perspective, and atmosphere details.
\end{tcblisting}

\section{Reward Models}
\label{app:reward_models}

This appendix details the four reward evaluators used in the Diffusion-NFT post-training stage
(\cref{sec:training_nft}). Three of them score text-to-image generation samples: text rendering,
aesthetic quality, and semantic understanding; these are used both for \mageflow post-training and for
the generation stream of \mageedit. The fourth, RationalRewards, scores instruction-based editing
samples in the editing stream of \mageedit. Each RL prompt carries a capability tag that routes it to
exactly one evaluator, which turns a generated or edited image into a scalar reward in $[0,1]$; rewards
from different evaluators are never mixed within a prompt. Below we describe each evaluator together
with a concrete example of how it scores.

\paragraph{Text rendering (PaddleOCR-VL-1.5).}
Every text-rendering prompt is annotated with a set of target strings
$\mathcal{T}=\{t_1,\dots,t_n\}$ that must appear, legibly and correctly spelled, in the image. We
render the sample, run PaddleOCR-VL-1.5~\citep{cui2026paddleocr} on it to obtain a recognized string,
and split that string on line and punctuation separators into segments, normalizing each segment by
removing spaces and lower-casing it to form a set $\mathcal{S}$. Each target $t\in\mathcal{T}$ is
normalized in the same way, with spaces removed and lower-cased, and we write $\tilde{t}$ for the
resulting normalized target. A target that occurs verbatim as a substring of some segment scores $1$;
otherwise we take its smallest character-level edit distance to any segment, capped at the target
length, and convert it to a similarity. Writing
$d(t)=\min_{s\in\mathcal{S}}\min\!\big(\mathrm{Lev}(s,\tilde{t}),\,|\tilde{t}|\big)$ for that best
capped distance, where $\mathrm{Lev}(\cdot,\cdot)$ is the character-level Levenshtein (edit) distance
and $|\tilde{t}|$ the target length, the per-target score is
\begin{equation}
p(t)=
\begin{cases}
1, & \text{if}\ \tilde{t}\subseteq s\ \text{for some}\ s\in\mathcal{S},\\[4pt]
\max\!\big(1-d(t)/|\tilde{t}|,\;0\big), & \text{otherwise,}
\end{cases}
\end{equation}
and the OCR reward is the mean over the prompt's targets,
$r_{\mathrm{ocr}}=\tfrac{1}{n}\sum_{i=1}^{n}p(t_i)\in[0,1]$. Capping the distance at $|\tilde{t}|$
bounds the penalty of a missing or badly garbled string to one full target length, so an absent
target scores $0$ and a perfect rendering scores $1$. For example, for the target ``First Place
Winner'' (normalized to \texttt{firstplacewinner}, length $|\tilde{t}|=16$) an exact reading scores
$1$, a single dropped character (``First Place Winer'', $d=1$) scores $1-1/16\approx0.94$, and absent
or unrecognizable text scores $0$. The recognizer takes no system prompt: it is queried with a single
user message containing the rendered image together with the literal instruction \texttt{OCR:}, decoded
greedily at temperature $0$.

\paragraph{Aesthetic quality (Qwen3.5-27B).}
Aesthetic-quality prompts are scored by a Qwen3.5-27B~\citep{qwen3.5} judge that grades the generated
image against a bank of detailed quality criteria. Each criterion is posed as an independent binary
(Yes${=}1$/No${=}0$) question, and the reward is the mean of the answers, so it takes evenly spaced
values in $[0,1]$; a criterion that does not apply, for example a hand-anatomy check on an image with
no hands, returns $1$ by convention. Grading quality as many focused yes/no items, rather than
eliciting a single opaque score, yields a stable and less reward-hackable signal. The judge is driven
by the system prompt and per-criterion user template below.
\begin{tcolorbox}[enhanced,colback=promptbg,colframe=promptborder,fonttitle=\bfseries\sffamily,arc=3pt,boxrule=0.8pt,left=10pt,right=10pt,top=8pt,bottom=8pt,title={Aesthetic-quality judge: system prompt and user template}]
\footnotesize\raggedright
\textbf{System.}\\
{\ttfamily You are evaluating whether an AI-generated image satisfies the following visual quality criterion.\\ You will receive a user prompt that was used to generate the image and a visual quality evaluation criterion.\\ Your job is to return 1 if the image fully satisfies the criterion, or 0 if it clearly fails. Do not explain or elaborate.\\ Only output: 1 or 0.}\\[5pt]
\textbf{User} (one call per criterion).\\
{\ttfamily The prompt of the generated image that needs to be evaluated is: \{user\_prompt\}\\ Evaluation Criterion (\{key\}): \{criterion\}\\ Scoring Instructions:\\ - Return 1 if the image fully satisfies the criterion.\\ - Return 0 if the image clearly fails.\\ Output: 1 or 0 only}
\end{tcolorbox}
The quality criteria plugged into \texttt{\{criterion\}} span general photographic quality and
detailed human-anatomy checks:
\begin{tcblisting}{enhanced,colback=promptbg,colframe=promptborder,arc=3pt,boxrule=0.8pt,left=8pt,right=8pt,top=4pt,bottom=4pt,fonttitle=\bfseries\sffamily,title={Aesthetic-quality judge: quality criteria},listing only,listing engine=listings,listing options={basicstyle=\ttfamily\scriptsize,breaklines=true,columns=fullflexible,breakindent=0pt,keepspaces=true}}
Naturalness:
Image looks like a realistic photograph with correct perspective, natural shadows, and physically plausible lighting.

Detail and Clarity:
Textures, edges, and fine details are sharp and coherent; no smearing, noise, or blur in the main subjects.

No Artifacts:
Image is free from visible distortions, watermarks, digital-rendering glitches, or impossible geometry.

Face and Eye Anatomy:
If a person's face is visible, both eyes are anatomically correct: two matched pupils with iris, symmetric placement, no warped, duplicate, or missing eyes, no asymmetric distortion. If no face is visible, return 1.

Hand and Finger Correctness:
If hands are visible, each hand has exactly five distinct fingers with no fusion, no extra or missing digits, and plausible knuckle structure and finger lengths. If no hand is visible, return 1.

Body Proportions and Limbs:
If a body is shown, limbs (arms, legs) and torso are anatomically plausible: correct count (2 arms + 2 legs), plausible joint articulation, no extra or missing limbs, reasonable head-to-body ratio. If no body is shown, return 1.

Skin Texture Realism:
If human skin is visible, texture shows natural variation (pores, subtle tone differences); not plastic, smeared, over-smoothed, or blurred to a wax-like appearance. If no skin is visible, return 1.

Pose Coherence:
If a person is posed, the pose is physically plausible: limbs do not intersect impossibly, weight distribution makes sense, no floating limbs or contorted joints. If no person is shown, return 1.
\end{tcblisting}
% For short, subject-driven prompts the judge additionally applies two prompt-aware criteria: a
% Subject-Match check (is the specific person, character, or described subject actually rendered?) and a
% Style-Match check (is a requested style such as photorealistic, anime, or oil painting visibly
% applied?).

\paragraph{Semantic understanding (Qwen3.5-27B).}
Semantic-understanding prompts are scored by the same Qwen3.5-27B judge under the same binary system
prompt and per-criterion user template as the aesthetic reward above; the only difference is that each
criterion is a yes/no alignment check derived from the prompt rather than a quality criterion. The
judge answers each check independently, and the reward is the fraction answered ``yes''. Decomposing
alignment into fine-grained faithfulness checks gives a dense, interpretable signal for multi-object
scenes. For each prompt, the checks cover:
\begin{promptbox}[Semantic-understanding judge: evaluation criteria]
\footnotesize\raggedright
{\ttfamily - Objects: is every object the prompt names present, with none of the required objects missing?\\ - Attributes: does each object have the specified color, material, shape, and state?\\ - Counts: is the exact number of each object present, neither too few nor too many?\\ - Spatial relations: are the objects positioned and arranged relative to one another as the prompt specifies?\\ - Actions and interactions: are the described actions, poses, and interactions between subjects correctly depicted?\\ - Scene and setting: do the background, environment, and overall composition match the prompt?\\ - Faithfulness: is the image free of hallucinated objects, attributes, or changes the prompt did not request?}
\end{promptbox}

\paragraph{Instruction-based editing.}
Editing rollouts are scored by RationalRewards~\citep{wang2026rationalrewards}, a reasoning reward
model that is shown the source image, the instruction, and the edited result, and rates four aspects
on a $1$--$4$ scale: (i)~text faithfulness (adherence to the edit instruction); (ii)~image
faithfulness (preservation of source content outside the edited region); (iii)~physical and visual
quality (plausibility and freedom from artifacts); and (iv)~text rendering (legibility of any edited
text, marked not-applicable when the edit involves no text). Writing $a_j\in[1,4]$ for the applicable
aspect scores, the reward linearly maps their mean from $[1,4]$ onto $[0,1]$:
\begin{equation}
r_{\mathrm{edit}}=\mathrm{clip}\!\left(\frac{\bar{a}-1}{3},\,0,\,1\right),
\qquad
\bar{a}=\frac{1}{|\mathcal{A}|}\sum_{j\in\mathcal{A}}a_j,
\end{equation}
where $\mathcal{A}$ is the set of applicable aspects, so an all-$4$ edit scores $1$ and an all-$1$
edit scores $0$.

\section{Application: Scientific Diagram Generation}
\label{sec:sciforma}

The \mageflow model features exceptional fine-tuning efficiency, enabling rapid adaptation to highly demanding, structure-critical visual domains. To demonstrate this transferability, we evaluate \magebase on scientific diagram generation, a representative task introduced by~\cite{sciforma2026}. Producing these figures serves as a rigorous stress test for generative models, as academic schematics require synchronous correctness across dense text rendering, directional arrow routing, and global layout topology.

\paragraph{Joint Training Strategy.}
Unlike the original two-stage fine-tuning in SciForma~\citep{sciforma2026}, we implement a single-stage data mixture. As shown in Table~\ref{tab:sciforma_data}, we construct a balanced data mixture to preserve general generative fidelity while building architectural competence. This joint recipe incorporates the \textbf{SciForma-700K} dataset alongside natural scenes, targeted text-rendering subsets, and poster designs.

\begin{table}[t]
\centering
\small
\setlength{\tabcolsep}{8pt}
\caption{\textbf{Training data mixture for \mageflow.} The configuration balances downstream domain performance with generic generative capabilities.}
\label{tab:sciforma_data}
\begin{tabular}{l r r r}
\toprule
\textbf{Source Dataset} & \textbf{Images} & \textbf{Packs} & \textbf{Weight} \\
\midrule
SciFormaData-700K (1024px) & 646,132 & 161,536 & 50\% \\
SciFormaData-700K (768px)  & 663,353 &  82,944 & 20\% \\
Natural scenes         & 357,009 &  74,269 & 15\% \\
Text rendering         &  74,866 &  15,360 & 10\% \\
Poster designs         &  38,290 &  10,084 &  5\% \\
\midrule
Total                  & 1,779,650 & 344,193 & 100\% \\
\bottomrule
\end{tabular}
\end{table}

\paragraph{Optimization Setup.}
We fine-tune the 4B \magebase checkpoint for 130k steps on an $8\times$ NVIDIA B200 GPU machine. The optimization is conducted via Supervised Fine-Tuning (SFT) within the native \magevae latent space at a $1024$-pixel native-aspect-ratio regime. We apply sequence packing with a fixed length of 20{,}480 tokens, utilize the AdamW optimizer with a constant learning rate of 1e-5 and a 500-step warmup. An Exponential Moving Average (EMA) with a decay rate of 0.99 is maintained throughout every 100 steps. We denote the resulting checkpoint as \mageflowsci.

\paragraph{Quantitative Results.}
We evaluate the fine-tuning efficiency on the \textbf{SciFormaBench-2K} evaluation suite~\cite{sciforma2026}, which assesses diagram quality across three structural dimensions: component, arrow, and text accuracy.

As outlined in Table~\ref{tab:sciforma_quant}, our 4B \magebase already demonstrates a stronger structural prior than the zero-shot FLUX.2-klein-Base baselines. This advantage is further amplified via fine-tuning; \mageflowsci achieves an Overall score of \textbf{61.61}, outperforming its zero-shot counterpart (40.80) by 20.81 points, with component and arrow accuracies increased by 13.70 and 24.70 points, respectively. These results validate that the 4B \magebase can be efficiently fine-tuned for specialized downstream applications. While this single-stage strategy establishes a viable baseline, the performance can be further enhanced by expanding to multi-stage fine-tuning or incorporating the proposed M-DPO algorithm~\cite{sciforma2026}.

\begin{table}[t]
\centering
\small
\setlength{\tabcolsep}{6pt}
\caption{\textbf{Quantitative results on the SciFormaBench-2K benchmark ($\uparrow$).} Metrics are reported on a 0--100 scale. The baseline models are evaluated in a zero-shot manner. \mageflowsci represents our fine-tuned Model on the data mixture from Table~\ref{tab:sciforma_data}.}
\label{tab:sciforma_quant}
\begin{tabular}{l r c c c c}
\toprule
\textbf{Model} & \textbf{\# Params} & \textbf{Component} & \textbf{Arrow} & \textbf{Text} & \textbf{Overall} \\
\midrule
FLUX.2-klein-Base (Zero-shot) & 4B & 50.04 & 18.45 & 10.95 & 27.05 \\
FLUX.2-klein-Base (Zero-shot) & 9B & 51.50 & 25.20 & 23.60 & 33.87 \\
\magebase (Zero-shot) &  4B & 56.79 & 35.82 &  28.17 & 40.80  \\
\midrule
\rowcolor{red!8}
\textbf{\mageflowsci (Ours)}  & \textbf{4B} & \textbf{70.49} & \textbf{60.52} & \textbf{52.60} & \textbf{61.61} \\
\bottomrule
\end{tabular}
\end{table}

\begin{figure*}[t]
\centering
\includegraphics[width=\linewidth]{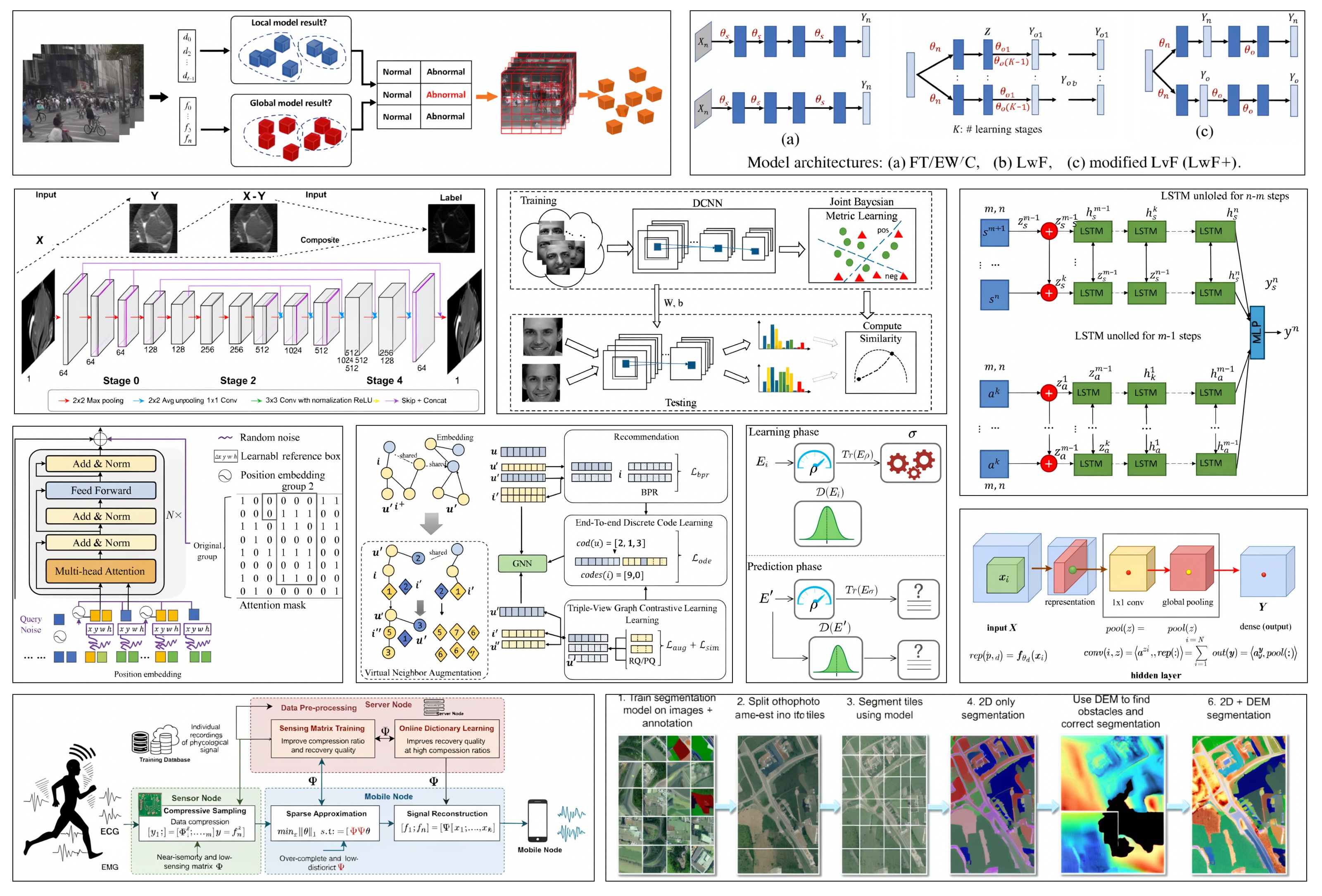}
\caption{\textbf{Scientific diagram generation results of \mageflowsci.} Examples include neural network layouts, unrolled temporal modules, graph contrastive frameworks, and text-annotated processing pipelines with legible text and correct directional arrows.}
\label{fig:sciforma_gallery}
\end{figure*}

\paragraph{Qualitative Analysis.}
Figure~\ref{fig:sciforma_gallery} presents the generation results of \mageflowsci across various scientific diagram archetypes. The results show precise global layouts and granular annotations. Specifically, the model generates neural network architectures with skip-connections, renders unrolled transformer blocks, and routes processing pipelines with directional vectors. Notably, it renders legible mathematical indices and step-by-step text descriptors inside nodes and dense workflows.

\begin{figure*}[t]
\centering
\includegraphics[width=\linewidth]{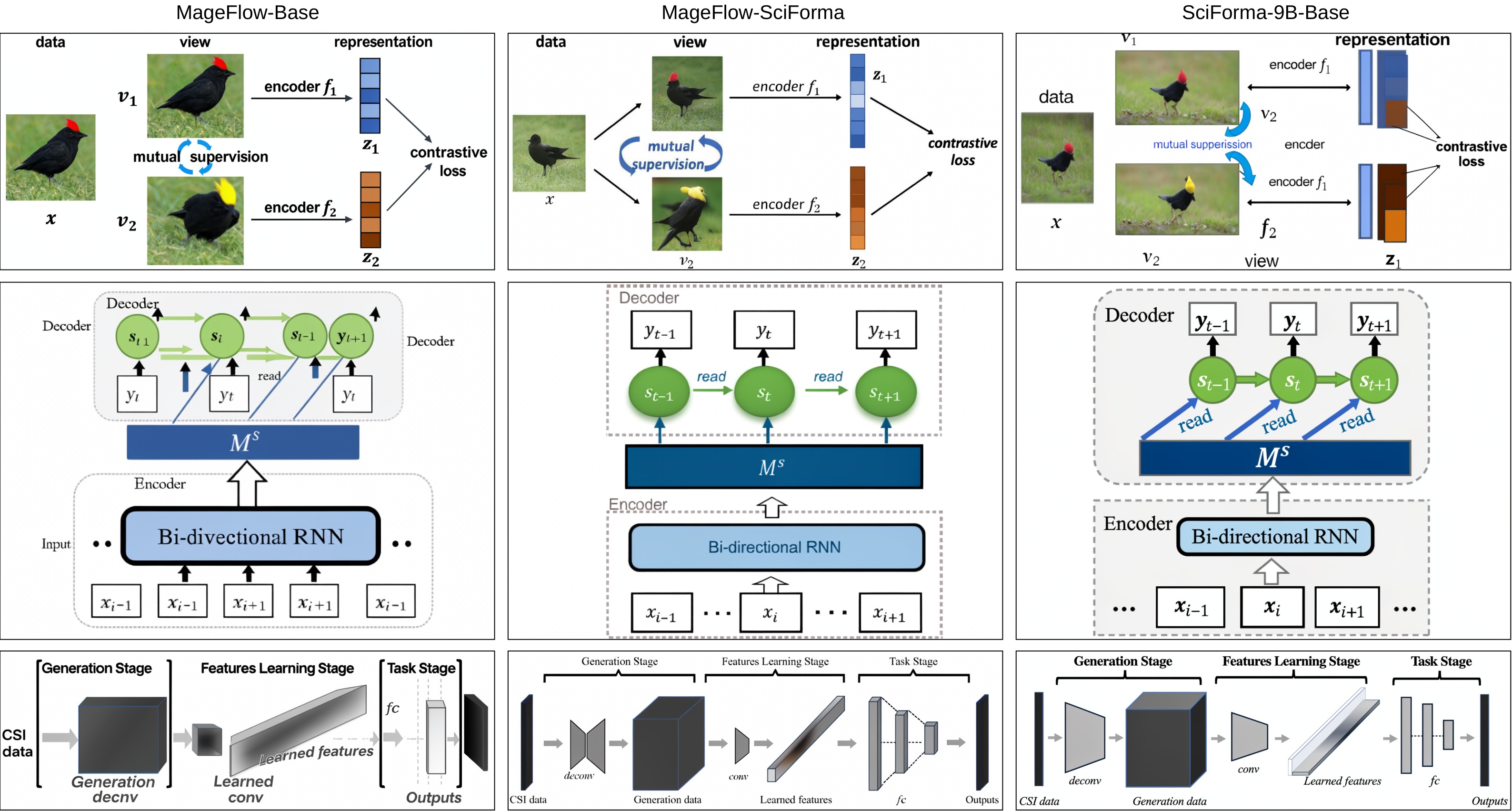}
\caption{\textbf{Qualitative comparison on scientific diagram generation.} Comparison among zero-shot \magebase, fine-tuned \mageflowsci, and SciForma-9B-Base. Fine-tuning resolves layout and text errors, allowing our 4B model to match the 9B vertical baseline.}
\label{fig:sciforma_comparison}
\end{figure*}

Figure~\ref{fig:sciforma_comparison} compares \magebase, \mageflowsci, and SciForma-9B-Base to evaluate the impact of fine-tuning. While \magebase outlines basic layout components, it exhibits limitations in text legibility and arrow semantics. After joint fine-tuning, \mageflowsci corrects these layout issues, rendering precise text and correct directional arrows. Furthermore, our 4B model produces schematic renderings comparable to SciForma-9B-Base, indicating that lightweight models can close the performance gap against larger domain-specific counterparts.

\newpage
\FloatBarrier
\bibliography{references}
\bibliographystyle{unsrtnat}

\end{document}